\documentclass[11pt]{article}
\pdfoutput=1

\usepackage{yx}

\title{Bayesian Design Principles for Frequentist Sequential Learning}

\author{Yunbei Xu\footnote{Decision, Risk, and Operations Division, Graduate School of Business; Email: \texttt{yunbei.xu@gsb.columbia.edu}.} \\ Columbia University
\and
Assaf Zeevi\footnote{Decision, Risk, and Operations Division, Graduate School of Business; Email: \texttt{assaf@gsb.columbia.edu}.} \\ Columbia University
}

\date{}

\begin{document}

\maketitle

\begin{abstract}

We develop a general theory to optimize the frequentist
regret for sequential learning problems,
where efficient bandit and reinforcement learning
algorithms can be derived from unified Bayesian
principles. We propose a novel optimization approach to generate ``algorithmic beliefs'' at each round, and use Bayesian posteriors to make decisions. The optimization objective to create ``algorithmic beliefs,'' which we term ``Algorithmic Information Ratio,'' represents an intrinsic complexity measure that effectively characterizes the frequentist regret of any algorithm. To the best of our knowledge, this is the first systematical approach to make Bayesian-type algorithms prior-free and applicable to adversarial settings, in a generic and optimal manner. Moreover,
the algorithms are simple and often efficient to implement. As a major application, we present a novel algorithm for  multi-armed bandits that achieves the ``best-of-all-worlds'' empirical performance in the stochastic, adversarial, and non-stationary environments. And we illustrate how these principles can be used in linear bandits, bandit convex optimization, and reinforcement learning.
\end{abstract}

\tableofcontents

\section{Introduction}
\subsection{Background}
  We address a broad class of sequential learning problems in the presence of {\it partial feedback}, which arise in numerous application areas including personalized recommendation  \citep{li2010contextual}, game playing \citep{silver2016mastering} and control \citep{mnih2015human}. An agent sequentially chooses among a set of possible decisions to maximize the cumulative reward. By ``partial feedback'' we mean the agent is only able to observe the feedback of her chosen decision, but does not generally observe what the feedback would be if she had chosen a different decision.  For example,  in multi-armed bandits (MAB),  the agent can only  observe the reward of her chosen action, but does not  observe the rewards of other actions.  In reinforcement learning (RL), the agent is only able to observe her state insofar as the chosen action is concerned,  while  other possible outcomes are not observed and the underlying state transition dynamics are unknown. In this paper, we present a unified approach that applies to bandit problems, reinforcement learning, and beyond.

The central challenge for sequential learning with partial feedback is to determine the optimal trade-off between {\it exploration} and {\it exploitation}. That is, the agent needs to try different decisions to learn the environment; at the same time, she wants to focus on ``good'' decisions that maximize her payoff. There are two basic approaches to study such exploration-exploitation trade-off: frequentist and Bayesian. One of the most celebrated examples of the frequentist approach is the family of Upper Confidence Bound (UCB) algorithms \citep{lai1985asymptotically, auer2002finite}. Here, the agent typically uses sample average or regression to estimate the mean rewards; and she optimizes the upper confidence
bounds of the mean rewards to make decisions.
Another widely used frequentist algorithm is EXP3  \citep{auer2002nonstochastic} which was designed for adversarial bandits; it  uses inverse probability weighting (IPW) to estimate the rewards, and then applies exponential weighting to construct decisions. 
One of the most celebrated examples of the Bayesian approach is Thompson Sampling (TS) with a pre-specifed, fixed prior \citep{thompson1933likelihood}. Here, the agent updates the Bayesian posterior at each round to learn the environment, and she uses draws from that posterior to optimize decisions. 

The advantage of the frequentist approach is that it does not require a priori knowledge of the environment. However,  it  heavily depends on a case-by-case analysis exploiting special structure of a   particular problem. For example, regression-based approaches can not be easily extended to adversarial problems; and IPW-type estimators are only known for simple rewards/losses such as discrete and linear. The advantage of the Bayesian approach is that Bayesian posterior is a generic and often optimal estimator if the prior is known. However, the Bayesian approach requires knowing the prior at the inception, which may not be accessible in complex or adversarial environments.
Moreover, maintaining posteriors is
computationally expensive for most priors.

In essence, the frequentist approach requires less information, but is less principled, or more bottom-up. 
On the other hand, the Bayesian approach is more principled, or top-down,  but requires stronger assumptions. In this paper we focus on the following research question:

{\it Can we design principled Bayesian-type algorithms, that are prior-free, computationally efficient, and work well in both stochastic and  adversarial/non-stationary environments? }

\subsection{Contributions} 

 In this paper, we synergize frequentist and Bayesian approaches to successfully answer the above question, through a novel idea that creates  ``algorithmic beliefs'' that are generated sequentially in each round, and uses Bayesian posteriors to make decisions. Our contributions encompass comprehensive theory, novel methodology, and applications thereby. We summarize the main contributions as follows.

\paragraph {Making Bayesian-type algorithms prior-free and applicable to adversarial settings.} To the best of our knowledge, we provide the first approach that allows Bayesian-type algorithms to operate without prior assumptions and hence also be applicable in adversarial settings, in a generic, optimal, and often computationally efficient manner. The regret bounds of our algorithms are no worse than the best theoretical guarantees known in the literature. In addition to its applicability in adversarial/non-stationary environments, our approach offers the advantages of being prior-free and often computationally manageable, which are typically not achievable by traditional Bayesian algorithms, except for simple model classes like discrete \citep{agrawal2012analysis} and linear \citep{agrawal2013thompson} rewards/losses.  It is worth noting that the main ideas underlying our methodology and proofs are quite insightful and  can be explained in a succinct manner.  

 \paragraph {General theory of ``Algorithmic Information Ratio'' (AIR).} We introduce intrinsic complexity measures that serve as objective functions in order to create ``algorithmic beliefs'' through round-dependent information. We refer to these measures as ``Algorithmic Information Ratio'' (AIR) types, including the adaption of ``Model-index AIR'' (MAIR) in the stochastic setting. Our approach always selects algorithmic beliefs by approximately maximizing AIR, and we show that AIR can bound the frequentist regret of any algorithm. We then show that AIR can be upper bounded by previously known complexity measures such as information ratio (IR) \citep{russo2016information} and   decision-estimation coefficient (DEC) \citep{foster2021statistical}. As an immediate consequence, our machinery converts existing regret upper bounds using information ratio and DEC, into simple frequentist algorithms with tight guarantees. And we provide methods and guarantees to approximately maximize AIR.

\paragraph {Novel algorithm for MAB with ``best of all worlds'' empirical performance.}
   As a major illustration, we propose a novel algorithm for Bernoulli multi-armed bandits (MAB) that
   achieves the ``best-of-all-worlds'' empirical performance in the stochastic, adversarial, and non-stationary environments. This algorithm is quite different from, and performs much better than, the traditional EXP3 algorithm, which has been the default choice for adversarial MAB for decades. At the same time, the algorithm outperforms UCB and is comparable to Thompson Sampling in the stochastic environment. Moreover, it outperforms  Thompson Sampling and ``clairvoyant'' restarted algorithms in non-stationary environments.

\paragraph{Applications to linear bandits, bandit convex optimization, and reinforcement learning.} Our theory can be applied to various settings, including linear and convex bandits, and reinforcement learning, by the principle of approximately maximizing AIR. Specifically, for linear bandits, we derive a modified version of EXP2 based on our framework, which establishes a novel connection between inverse propensity weighting (IPW) and Bayesian posteriors. For bandit convex optimization, we propose the first algorithm that attains the best-known $\tilde{O}(d^{2.5}\sqrt{T})$ regret \citep{lattimore2020improved} with a finite $\text{poly}(e^d\cdot T)$ running time. Lastly, in reinforcement learning, we provide simple algorithms that match the regret bounds proven in the early work \cite{foster2021statistical}.

\paragraph{Combining estimation and decision-making.} Our approach jointly optimizes the belief of an environment and probability of decision. Most existing algorithms including UCB, EXP3, Estimation-to-Decision (E2D) \citep{foster2021statistical}, TS, and Information-Directed Sampling (IDS) \citep{russo2014learning} maintain a different viewpoint that separates algorithm design into a black-box estimation method (sample average, linear regression, IPW, Bayesian posterior...) and a decision-making rule that takes the estimate as  an input to an optimization problem. In contrast, by optimizing AIR to generate new beliefs, our algorithm {\it simultaneously} deals with estimation and optimization. This viewpoint is quite powerful  and broadens the general scope of  bandit algorithms.

\subsection{Related literature}
The literature in the broad area of our research is vast. To maintain a streamlined and concise presentation, we focus solely on the most relevant works.  \cite{russo2016information, russo2014learning} propose the concept of ``information ratio'' (IR) to analyze and design Bayesian bandit algorithms. Their work studies Bayesian regret with a known prior rather than the frequentist regret. \cite{lattimore2021mirror} proposes an algorithm called ``Exploration by Optimization (EBO),'' which is the first general frequentist algorithm that optimally bounds the frequentist regret of bandit problems and partial monitoring using information ratio. However, the EBO algorithm is more of a conceptual construct as it requires intractable optimization over the complete class of ``functional estimators,'' and hence may not be implementable in most settings of interest. Our algorithms are inspired by EBO, but are simpler in structure and run in decision and model spaces (rather than intractable functional spaces). In particular, our approach advances EBO by employing  explicit construction of estimators, offering flexibility in selecting updating rules, and providing thorough design and computation guidelines that come with provable guarantees.  Our work also builds upon the line of research initiated by  \cite{foster2021statistical}, which introduces the concept of the ``decision-estimation coefficient'' (DEC) as a general complexity measure for bandit and RL problems. Remarkably, DEC not only provides sufficient conditions but also offers the first necessary characterization (lower bounds) for interactive decision making, akin to the VC dimension and the Rademacher complexity in statistical learning. It is important to note that while DEC is tight for one part of the regret, as demonstrated in \cite{foster2021statistical} and \cite{foster2023tight}, the remaining ``estimation complexity'' term in the regret bound can be sub-optimal when the model class is large. The proposed E2D algorithm in \cite{foster2021statistical} separates the black-box estimation method from the decision-making rule. For this reason, E2D is often unable to achieve optimal regret, even for multi-armed bandits, when the size of the model class is redundant in terms of estimation complexity. Moreover, E2D only works for stochastic environments. \cite{foster2021statistical} also provides an adaptation of EBO from \cite{lattimore2021mirror} to improve estimation complexity, and the subsequent work \cite{foster2022complexity}  extends the theory of DEC to adversarial environments by adapting EBO. However, as adaptations of EBO, these algorithms present computational challenges, as discussed earlier.

\section{Preliminaries and definition of AIR}
\subsection{Problem formulation}\label{subsec problem formulation}
To state our results in the broadest manner, we adopt the general formulation of Adversarial Decision Making with Structured Observation (Adversarial DMSO). This setting extends the original Adversarial DMSO setup introduced in \cite{foster2022complexity} by incorporating adversarial partial monitoring, where the reward may not be directly observable. The setting covers broad problems including bandit problems, reinforcement learning, partial monitoring, and beyond. For a locally compact metric space we denote by  $\Delta(\cdot)$ the set of Borel probability measures on that space. Let $\Pi$ be a compact decision space. Let $\M$ be a compact model class where each model $M: \Pi\rightarrow \OC$ is a mapping from the decision space to a locally compact observation space $\OC$.  A problem instance in this protocol can be described by the decision space $\Pi$ and the model class $\M$. We define the mean reward function associated with model $M$ by $f_M$.

Consider a $T-$round game played by a randomized player in an adversarial environment. At each round $t=1,\ldots,T$, the agent determines a probability $p_t$ over the decisions, and the environment selects a model  $M_t\in \M$. Then the decision $\pi_t\sim p_t$ is sampled and an observation $o_t\sim M_t(\pi_t)$ is revealed to the agent.  An {\it admissible algorithm} $\alg$ can be described by a sequence of mappings where the $t-$th mapping maps the past decision and observation sequence $\{\pi_i, o_i\}_{i=1}^{t-1}$ to a probability $p_t$ over decisions.  Throughout the paper we denote $\{\mathfrak{F}_t\}_{t=1}^T$ as the filtration where $\mathfrak{F}_t$ is the $\sigma-$algebra generated by random variables $\{\pi_t,o_t\}_{s=1}^t$, and use the notations $\E_{t}\lb\cdot\rb=\E[\cdot|\mathfrak{F}_t]$ for conditional expectation starting round $t+1$.

The {\it frequentist regret}  of the algorithm $\alg$ against the usual target of single best decision in hindsight  is defined as
\begin{align*}
    \reg_T= \sup_{\pi^*\in\Pi}\E\left[\sum_{t=1}^T f_{M_t}(\pi^*)-\sum_{t=1}^T f_{M_t}(\pi_t)\right],
\end{align*} where the expectation is taken with respect to the randomness in decisions and observations. There is a large literature that focuses on the so-called  stochastic environment, where $M_t=M^*\in \M$ for all rounds, and the single best decision $\pi^*\in \arg\min f_{M^*}(\pi)$ is the natural oracle. Regret bounds for adversarial sequential learning problems naturally apply to stochastic problems. 
We illustrate how the general formulation covers bandit problems, and leave the discussion of reinforcement learning to Section \ref{sec mair}.

\begin{example}[Bernoulli multi-armed bandits (MAB)]\label{example mab}
We illustrate how the general formulation reduces to the basic MAB problem with Bernoulli reward. Let  $\Pi=[K]=\{1,\cdots, K\}$ be a finite set of $K$ actions, and $\F$ be the set of all possible mappings from $[K]$ to $[0,1]$. Take
$\M=\{M_{f}:f\in \F\}$ as the induced model class, where each $M_f$ maps $\pi$ into the Bernoulli distribution  $\text{Bern}(f(\pi))$. The mean reward function for model $M_f$ is $f$ itself. At each round $t$, the environment selects a mean reward function $f_t$, and the observation $o_t$ is the incurred reward $r_t(\pi_t)\sim \text{Bern}(f_t(\pi_t))$. 
\end{example}

\begin{example}[Structured bandits]\label{example structured bandits}
We consider bandit problems with general structure of the mean reward function. Let  $\Pi$ be a $d-$dimensional action set, and $\F\subseteq\{f: \Pi\rightarrow[0,1]\}$ be a function class that encodes the structure of the mean reward function. Take $\M=\{M_f: f\in\F\}$ as the induced model class, where each  $M_f$ maps $\pi$ to the Bernoulli distribution $\text{Bern}(f(\pi))$. The mean reward function for model $M_f$ is $f$ itself. 
For example, in $d-$dimensional linear bandits, the mean reward function $f$ is parametrized by some $\theta\in\Theta\subseteq\R^d$ such that $f(\pi)=\theta^T\pi, \forall\pi\in\Pi$. And in bandit convex optimization, the mean reward (or loss) function class $\F$ is the set of all concave (or convex) mappings from $\Pi$ to $[0,1]$.
\end{example}

\subsection{Algorithmic Information Ratio}\label{subsec def air}
 
 Let $\nu$ be a probability measure of the joint random variable  $(M, \pi^*)\in\M\times \Pi$, and $p$ be a distribution of another independent random variable $\pi\in\Pi$. Given a probability measure $\nu$, let 
\begin{align*}
    \nu_{\pi^*}(\cdot)= \int_\M \nu(M, \cdot)dM
\end{align*}
be the marginal distribution of $\pi^*\in\Pi$. Viewing $\nu$ as a prior belief over $(M, \pi^*)$, we define $\nu(\cdot|\pi,o)$ as the Bayesian posterior belief conditioned on the decision being $\pi$ and the observation (generated from the distribution $M(\pi)$)  being $o$; and we define the marginal posterior belief of $\pi^*$ conditioned on  $\pi$ and $o$ as 
\begin{align*}
    \nu_{\pi^*|\pi,o}(\cdot)=\int_{\M}\nu(M,\cdot|\pi,o) dM,
\end{align*} where the subscript $\pi^*$ in $\nu_{\pi^*|\pi,o}$ is an index notation, whereas $\pi$ and $o$ in $\nu_{\pi^*|\pi,o}$ are random variables. For two probability measures $\P$ and $\mathcal{Q}$ on a measurable space, the Kullback-Leibler (KL) divergence between $\P$ and $\mathcal{Q}$ is defined by $\KL(\P,\mathbb{Q})=\int \log \frac{d\P}{d\mathbb{Q}}d\P$  if the measure $\P$ is absolutely continuous with respect to the measure
 $\mathcal{Q}$, and $+\infty$ otherwise.

Now we introduce a central definition in this paper---Algorithmic Information Ratio (AIR). 
 \begin{definition}[Algorithmic Information Ratio]\label{def air}
 Given a reference probability $q\in\inte(\Delta(\Pi))$  and learning rate $\eta>0$, we define the ``Algorithmic Information Ratio'' (AIR) for probability $p$ of $\pi$ and belief $\nu$ of $(M,\pi^*)$ as
 \begin{align*}
     \air_{q,\eta}(p,\nu)= \E_{ p,\nu}\lb f_M( \pi^*)-f_M(\pi)-\frac{1}{\eta}\KL(\nu_{\pi^*|\pi,o}, q)\rb,
 \end{align*}
 where the expectation is taken with $\pi\sim p,(M,\pi^*)\sim \nu, o\sim M(\pi)$.
 \end{definition}

The term ``Algorithmic Information Ratio'' was used to highlight the key difference between AIR and classical information ratio (IR) measures (to be presented shortly in \eqref{eq: information ratio}). Firstly, AIR incorporate a reference probability $q$ in its definition, while classical IR does not. This additional flexibility makes AIR useful for algorithm design and analysis. Secondly, AIR is defined in an offset form, whereas IR is defined in a ratio form. We choose the word ``algorithmic'' because AIR is particularly suited to designing constructive and efficient frequentist algorithms; we remain the term ``ratio'' as it is consistent with previous literature on the topic. The formulation of AIR  is inspired by the optimization objectives in recent works \cite{lattimore2021mirror} on EBO  and \cite{foster2021statistical} on DEC. There are various equivalences between different variants of AIR, DEC, EBO, and IR when considering minimax algorithms, worst-case environments, and choosing the appropriate index, divergence, and formulation (see the next two subsections for details). However, AIR crucially focuses on ``algorithmic belief'' $\nu$ rather than maximizing with respect to the worst-case deterministic model, providing algorithmic unity, interpretability, and flexibility.

 Note that AIR is linear with respect to $p$ and  concave with respect to $\nu$, as conditional entropy is always concave with respect to the joint probability measure (see Lemma \ref{lemma conditional entropy}). By the generalized Pythagorean theorem (Lemma \ref{lemma pythegorean}) for the KL divergence and the fact about posterior $\E_{p,\nu}[\nu_{\pi^*|\pi,o}]=\nu_{\pi^*}$, we have the equality $\E_{p,\nu}[\KL(\nu_{\pi^*|\pi,o},q)]=\E_{p,\nu}[\KL(\nu_{\pi^*|\pi,o},\nu_{\pi^*})]+\KL(\nu_{\pi^*},q)$. Thus it will be illustrative to write AIR as the sum of three items:
 \begin{align*}
     \air_{q,\eta}(p,\nu)
     = &\underbrace{\E_{\pi\sim p,(M,\pi^*)\sim\nu}\lb f_M(\pi^*)-f_M(\pi)\rb}_{\text{expected regret}}\\
     &-\frac{1}{\eta}\underbrace{\E_{\pi\sim p,(M,\pi^*)\sim\nu,o\sim M(\pi)}\lb\KL(\nu_{\pi^*|\pi,o},\nu_{\pi^*})\rb}_{\text{information gain}}-\frac{1}{\eta}\underbrace{\KL(\nu_{\pi^*},q)}_{\textup{regularization by $q$}},
 \end{align*}
 where: the ``expected regret'' measures the difficulty of exploitation; the ``information gain'' is the amount of information gained about the marginal distribution of $\pi^*$ by observing the random variables $\pi$ and $o$, and this in fact measures the degree of exploration; and the last ``regularization'' term forces the marginal distribution of  $\pi^*$ to be ``close'' to the reference probability distribution $q$. By maximizing AIR, we generate an ``algorithmic belief'' that simulates the worst-case environment. This algorithmic belief will automatically balance exploration and exploitation, as well as being close to the chosen reference belief (e.g., a standard reference is the posterior from previous round, as used in traditional Thompson Sampling).

\subsection{Bounding AIR by IR and DEC} \label{subsec air ir dec} 
Notably, our framework allows for the utilization of essential all existing upper bounds for information ratio and DEC in practical applications, enabling the derivation of the sharpest regret bounds known, along with the development of constructive algorithms. In this subsection we demonstrate that AIR can be upper bounded  by IR and DEC.

We present here the traditional definition of  Bayesian information ratio \citep{russo2016information}. See \cite{russo2016information, russo2014learning, lattimore2019information, lattimore2020bandit, hao2022regret} for upper bounds of IR in bandit problems and structured RL problems.

 \begin{definition}[Information ratio]\label{def ir}
 Given belief $\nu$ of $(M,\pi^*)$ and decision probability $p$ of $\pi$, the information ratio  is defined as
 \begin{align}\label{eq: information ratio}
    \ir(\nu,p)= \frac{(\E_{\nu, p}\lb f_M(\pi^*)-f_M(\pi) \rb)^2}{\E_{\nu,p}\lb \KL({\nu}_{\pi^*|\pi,o}, {\nu}_{\pi^*})\rb}.
\end{align}
 \end{definition}
Note that the traditional information ratio \eqref{eq: information ratio} does not involve any reference probability distribution $q$ (unlike AIR).  By completing the square, it is easy to show that AIR can always be bounded by IR as follows. 
\begin{lemma}[Bounding AIR by IR]\label{lemma air ir}
    For any $q\in \text{int}(\Delta(\Pi))$, $p\in\Delta(\Pi)$, belief $\nu\in\Delta(\M\times \Pi)$, and $\eta>0$, we have
\begin{align*}
    \air_{q,\eta}(p,\nu)\leq \frac{\eta}{4}\cdot\ir(\nu,p).
\end{align*}
\end{lemma}

The recent paper \cite{foster2021statistical} introduced the complexity measure DEC, with the aim of unifying bandits and many reinforcement learning problems of interest.
 \begin{definition}[Decision-estimation coefficient]\label{def dec}Given a model class $\M$, a nominal model $\bar{M}$ and $\eta>0$, we define the decision-estimation coefficient by 
\begin{align}
    \dec_{\eta}\lp\M,\bar{M}\rp= \inf_{p\in\Delta(\Pi)}\sup_{M\in\M}\E_{\nu, p}\Big[ f_M(\pi_M)-f_M(\pi)-\frac{1}{\eta}D_\textup{H}^2\lp M(\pi), \bar{M}(\pi)\rp\Big],\nonumber
\end{align}
where $D_{\textup{H}}^2(\P,\mathbb{Q})=\int(\sqrt{d\P}-\sqrt{d\mathbb{Q}})^2$ is the squared Hellinger distance between two probability measures. And we define
\begin{align}
    \dec_{\eta}\lp\M\rp= \sup_{\bar{M}\in\textup{conv}(\M)}\dec_\eta(\M,\bar{M}).\label{eq: dec class}
\end{align}
\end{definition} 
DEC offers a unifying perspective on various existing structural conditions in the literature concerning RL. For comprehensive explanations on how DEC relates to and subsumes these structural conditions, such as Bellman rank \citep{jiang2017contextual}, Witness rank \citep{sun2019model}, (Bellman-) Eluder dimension \citep{russo2013eluder, wang2020reinforcement, jin2021bellman}, bilinear classes \citep{du2021bilinear}, and linear function approximation \citep{dean2020sample, yang2019sample, jin2020provably}, we encourage readers to consult Section 1 and Section 7 in \cite{foster2021statistical} and the references therein.  Moreover, a slightly strengthened version of DEC, defined through the KL divergence instead of the squared Hellinger divergence, 
\begin{align}
    \dec^\KL_{\eta}\lp\M\rp= \sup_{\bar{M}\in\textup{conv}}\inf_{p\in\Delta(\Pi)}\sup_{M\in\M}\E_{\nu, p}\Big[ f_M(\pi_M)-f_M(\pi)-\frac{1}{\eta}\KL\lp M(\pi), \bar{M}(\pi)\rp\Big],\nonumber
\end{align}
can be upper bounded by the traditional information ratio. This result follows from Proposition 9.1 in \cite{foster2021statistical}. Note that the convex hull feasible region for the reference model $\bar{M}$ in the notation \eqref{eq: dec class} is fundamental because the  convex hull also appears in the lower bound, see the discussions in \cite{foster2023tight} and    Section 3.5.1 in \cite{foster2021statistical} for details.

We demonstrate in the following lemma that the worst-case value of AIR, when employing a ``maximin'' strategy for selecting $p$, is equivalent to the KL version of DEC applied to the convex hull of the model class (see below). Consequently, it is bounded by the standard version of DEC using the squared Hellinger distance.

\begin{lemma}[Bounding AIR by DEC]\label{lemma air dec}Given model class $\M$ and $\eta>0$, we have 
\begin{align}\label{eq: air dec}
\sup_{q\in  \inte(\Delta(\Pi))}\sup_{\nu}\inf_{p}\ \air_{q,\eta}(p,\nu)
= \texttt{\textup{DEC}}^{\KL}_{\eta}(\textup{conv}(\M))\leq \texttt{\textup{DEC}}_{\eta}(\textup{conv}(\M)).
\end{align}
\end{lemma}
To prove Lemma \ref{lemma air dec}, we can start by noting that the left-hand side of \eqref{eq: air dec} is equivalent to the ``parametric information ratio,'' defined as
\begin{align}\label{eq: pir}
\max_{\nu}\min_p \E_{\nu,p}\lb f_M(\pi^*)-f_M(\pi)-\frac{1}{\eta}\KL(\nu_{\pi^|\pi,o}, \nu_{\pi^*})\rb,
\end{align}
which was introduced in \cite{foster2022complexity}. This equivalence can be shown by using the concavity of AIR to exchange $\sup$ over $q$ and $\min$ over $p$. Furthermore, the equivalence between \eqref{eq: pir} and $\texttt{\textup{DEC}}^{\KL}_{\eta}(\textup{conv}(\M))$ has been established by Theorem 3.1 in \cite{foster2022complexity}. Therefore, we obtain a proof of Lemma \ref{lemma air dec}.

We conclude that AIR is the tightest complexity measure in the adversarial setting, with its maximin value equivalent with the EBO objective in \cite{lattimore2021mirror}, the KL version of DEC, and the offset version of IR \eqref{eq: pir}. Such tightness is established by Lemma \ref{lemma air dec} and the lower bounds in \cite{foster2022complexity}. However, for reinforcement learning problems in the stochastic setting, it is often desirable to remove the convex hull on the right-hand side of \eqref{eq: air dec}. To this end, we introduce a tighter version of AIR, called ``Model-index AIR'' (MAIR), which is upper bounded by the original version of DEC  using model class $\M$ rather than its convex hull (see Lemma \ref{lemma mdir relationship} in the next subsection), and allows us to apply essentially all existing regret upper bounds using DEC to our framework.

\subsection{Model-index AIR (MAIR) and DEC}\label{subsec MAIR}
Denote decision $\pi_M\in {\arg\min}_{\Pi} f_M(\pi)$ be the induced optimal decision of model $M$. In the stochastic environment, where $M_t=M^*\in\M$ for all rounds, the benchmark policy in the definition of regret is the natural oracle $\pi_{M^*}$. Unlike the adversarial setting, where algorithmic beliefs are formed over pairs of models and optimal decisions, in the stochastic setting, we only need to search for algorithmic beliefs regarding the underlying model. This distinction allows us to develop a strengthened version of AIR, which we call ``Model-index AIR'' (MAIR), particularly suited for studying reinforcement learning problems.

\begin{definition}[Model-index AIR]\label{def mair}Denote $\rho\in \inte(\Delta(\M))$ be a reference distribution of models, and $\mu\in \Delta(\M)$ be a prior belief of models,  we define the ``Model-index Algorithmic Information Ratio'' as
\begin{align*}
 \mair_{\rho,\eta}(p, \mu)&=  \E_{\mu, p}\lb f_M( \pi_M)-f_M(\pi)-\frac{1}{\eta}\KL(\mu(\cdot|\pi,o), \rho)\rb,
\end{align*}
where $\mu(\cdot|\pi,o)$ is the Bayesian posterior belief of models
induced by the prior belief $\mu$.
\end{definition}

By the data processing inequality (Lemma \ref{lemma data processing}),  KL divergence between two model distributions will be no smaller than KL divergence between the two induced decision distributions. Thus we have the following Lemma.
\begin{lemma}[Relationship of MAIR and AIR]\label{lemma mair air}When $q$ is the decision distribution of $\pi_M$ induced by the model distribution $\rho$, and $\nu$ is the  distribution of $(M, \pi_M)$ induced by the model distribution $\mu$, we have
\begin{align*}
    \mair_{\rho, \eta}(p,\mu)&\leq \air_{q,\eta}(p,\nu).
\end{align*}

\end{lemma}

Lemma \ref{lemma air dec} has shown that the worst-case value of AIR under the ``maximin'' strategy is smaller than DEC of the convex hull of $\M$. Now we demonstrate that the worst-case value of MAIR under a ``maximin'' strategy is smaller than DEC of the original model class $\M$, which does not use the convex hull in its argument. In fact, there is an exact equivalence between the worst-case value of MAIR and the KL version of DEC, as is the case in Lemma \ref{lemma air dec} for AIR.

\begin{lemma}[Bounding MAIR by DEC]\label{lemma mdir relationship}Given model class $\M$ and $\eta>0$, we have 
\begin{align*}
\sup_{\rho\in  \inte(\Delta(\M))}\sup_{\mu}\inf_{p} \ {{\texttt{{\textup{MAIR}}}}}_{\rho,\eta}(p,\nu) = \dec_{\eta}^\KL(\M)\leq   \texttt{\textup{DEC}}_{\eta}(\M).
\end{align*}
\end{lemma}
 We conclude that lemma \ref{lemma mdir relationship} enables our approach to match the tightest known regret upper bounds using DEC, and the tightness of MAIR follows from the lower bounds of DEC in \cite{foster2021statistical}. In Section \ref{sec mair}, we discuss how to derive principled algorithms using MAIR and application to reinforcement learning. In Section \ref{subsec generic MAIR}, we also discuss how the pursuit of MAIR comes at the cost of larger estimation complexity, explaining the trade-off between AIR and MAIR.

\section{Algorithms}\label{sec algorithm}
In this section we focus on the adversarial setting and leverage AIR to analyze regret and design algorithms. All the results are extended to leverage MAIR in the stochastic setting in Section \ref{sec mair}. For the comparison between AIR and MAIR, we refer to the ending paragraphs of Section \ref{subsec AMS} and Section \ref{subsec generic MAIR} for detailed discussion.
\subsection{A generic regret bound leveraging AIR}
Given an arbitrary admissible algorithm $\alg$ (defined in Section \ref{subsec problem formulation}), we can  generate a sequence of \textit{algorithmic beliefs} $\{\nu_t\}_{t=1}^T$ and a corresponding sequence of \textit{reference probabilities} $\{q_t\}_{t=1}^T$ in a sequential manner as shown in Algorithm \ref{alg: frequentist sampling}.
\begin{algorithm}[htbp]
\caption{Maximizing AIR to create algorithmic beliefs}
\label{alg: frequentist sampling}
Input algorithm $\alg$ and learning rate $\eta>0$.

Initialize $q_1$ to be the uniform distribution over $\Pi$. 
\begin{algorithmic}[1]
\FOR{round $t=1,2,\cdots, T$}
    \STATE{{ Obtain $p_t$ from $\alg$. Find a distribution $\nu_t$ of $(M, \pi^*)$  that solves }
    \begin{align*}
       \sup_{\nu\in\Delta(\M\times \Pi)} \texttt{AIR}_{q_t,\eta}(p_t, \nu).
    \end{align*}}
    \STATE{ The algorithm $\alg$ samples decision $\pi_t\sim p_t$ and observes the feedback $o_t\sim M_t(\pi_t)$.}
    \STATE{ Update $q_{t+1}={(\nu_t)}_{\pi^*|\pi_t,o_t}$.}
\ENDFOR
\end{algorithmic}
\end{algorithm}
Maximizing AIR to create algorithmic beliefs is an alternative approach to traditional  estimation procedures, as the resulting algorithmic beliefs will simulate the true or worst-case environment. In particular, this approach only stores a single  distribution $q_{t+1}=(\nu_t)_{\pi^*|\pi_t,o_t}$ at round $t$, which is the Bayesian posterior obtained from belief $\nu_t$ and observations $\pi_t,o_t$, and it is made to forget all the rest information from the past. We should note that all values in the sequence $\{q_t\}_{t=1}^T$ will reside within the interior of  $\Delta(\Pi)$. This is due to the fact that the (negative) AIR is a Legendre function concerning the marginal distribution $\nu_{\pi^*}$.\footnote{We refer to Definition \ref{def legendre} and Lemma \ref{lemma minimizer Legendre} for more details regarding the property that ${(\nu_t)}_{\pi^*}\in\inte(\Delta(\Pi))$ for all $t$. Furthermore,  it is worth emphasizing that AIR (and the negative KL divergence term)  can be appropriately defined as negative infinity ($-\infty$)  when the reference probability $q$ resides on the boundary of $\Delta(\Pi)$ and $\nu_{\pi^*}$ is not absolutely continuous with respect to $q$. This is a direct result of the well-defined nature of the KL divergence. Such simple extension can  provide an alternative solution to address any concerns related to boundary issues that may arise in the paper.}

Based on these algorithmic beliefs, we can provide  regret bound for an arbitrary algorithm. Here we assume $\Pi$ to be finite (but potentially large) for simplicity; this assumption can be relaxed using standard discretization and covering arguments.
\begin{theorem}[Generic regret bound for arbitrary learning algorithm]\label{thm regret}Given a finite decision space $\Pi$, a compact model class $\M$, the regret of an arbitrary learning algorithm $\alg$ is bounded as follows, for all $T\in\N$,
\begin{align}
   \reg_T\leq \frac{\log |\Pi|}{\eta}+ \E\lb\sum_{t=1}^T \texttt{\textup{AIR}}_{q_t,\eta}(p_t,\nu_t)\rb.\label{eq: thm regret}
\end{align}
\end{theorem}

Note that Theorem \ref{thm regret} provides a powerful tool to study the regret of an arbitrary algorithm using the concept of AIR. Furthermore, it suggests that the algorithm should choose decision with probability $p_{t+1}$ according to the posterior $({(\nu_t)}_{\pi^*|\pi_t,o_t}$. Building on this principle to generate algorithmic beliefs, we provide two concrete algorithms: ``Adaptive Posterior Sampling'' (APS) and ``Adaptive Minimax Sampling'' (AMS). Surprisingly, their regret bounds are as sharp as the best known regret bounds of existing Bayesian algorithms that {\it require} knowledge of a well-specified prior. An analogy of the theorem leveraging MAIR in the stochastic setting is presented as Theorem \ref{thm regret approx stochastic}.

\subsection{Adaptive Posterior Sampling (APS)} When the agent always selects $p_{t+1}$ to be equal to the posterior $q_{t+1}={(\nu_t)}_{\pi^*|\pi_t,o_t}$, and optimizes for algorithmic beliefs as in Algorithm \ref{alg: frequentist sampling}, we call the resulting algorithm ``Adaptive Posterior Sampling'' (APS).
\begin{algorithm}[htbp]
\caption{ Adaptive Posterior Sampling (APS)}
\label{alg: adaptive posterior sampling}
Input learning rate $\eta>0$. 

Initialize $p_1=\text{Unif}(\Pi)$.  
\begin{algorithmic}[1]
\FOR{round $t=1,2,\cdots, T$}
    \STATE{{ Find a distribution $\nu_t$ of $(M, \pi^*)$  that solves }
    \begin{align*}
       \sup_{\nu\in\Delta(\M\times \Pi)} \texttt{AIR}_{p_t,\eta}(p_t, \nu).
    \end{align*}}
    \STATE{ Sample decision $\pi_t\sim p_t$ and observe  $o_t\sim M_t(\pi_t)$.}
    \STATE{ Update $p_{t+1}={(\nu_t)}_{\pi^*|\pi_t,o_t}$.}
\ENDFOR
\end{algorithmic}
\end{algorithm}

At round $t$,  APS inputs $p_t$ to the objective $\air_{p_t,\eta}(p_t,\nu)$ to optimize for the algorithmic belief $\nu_t$; and it sets $p_{t+1}$ to be the Bayesian posterior obtained from belief $\nu_t$ and observations $\pi_t,o_t$. 
Unlike traditional TS, APS does not require knowing the prior or stochastic environment; instead, APS creates algorithmic beliefs ``on the fly'' to simulate the worst-case environment.  We can prove the following theorem using the regret bound \eqref{eq: thm regret} in Theorem \ref{thm regret}, and the upper bounds of AIR by IR and DEC established in Section \ref{subsec air ir dec}.

\begin{theorem}[Regret of APS]\label{thm adaptive posterior sampling}Assume that  $f_M(\pi)\in[0,1]$ for all $M\in\M$ and $\pi\in\Pi$. The regret of Algorithm \ref{alg: adaptive posterior sampling} with $\eta= \sqrt{2\log|\Pi|/(( \ir_{\textup{H}}(\textup{TS})+4)\cdot T)}$ and all $T\geq 5\log |\Pi|$ is bounded by
\begin{align*}
   \reg_T\leq  \sqrt{2\log |\Pi| \lp \texttt{\textup{IR}}_{\textup{H}}(\textup{TS})+4\rp T},
\end{align*}
where $\ir_{\textup{H}}(\textup{TS}):=\sup_{\nu}\ir_{\textup{H}}(\nu,\nu_{\pi^*})$  is the maximal value of information ratio\footnote{For technical reason we use the squared Hellinger distance to define $\ir_{\textup{H}}$ (instead of KL as in the definition \eqref{eq: information ratio} of $\ir$). There is no essensial difference between the definitions of $\ir$ and $\ir_{\textup{H}}$ in practical applications, since all currently known bounds 
on the information ratio hold for the stronger definition $\ir_{\textup{H}}$ with added absolute constants. See Appendix \ref{appendix information ratio H} for details.} for Thompson Sampling. Moreover, the regret of Algorithm \ref{alg: adaptive posterior sampling} with any $\eta\in(0,1/3]$ is bounded as follows, for all $T\in\N$
\begin{align*}
    \reg_T\leq \frac{\log|\Pi|}{\eta}+T\cdot \lp \dec_{2\eta}^{\textup{TS}}(\textup{conv}(\M))+2\eta\rp,
\end{align*}  
where $\dec_{2\eta}^{\textup{TS}}(\textup{conv}(\M)):=\sup_{\bar{M}\in\textup{conv}(\M)}\sup_{\mu\in\Delta(\textup{conv}(\M))}\E_{\mu, p^\textup{TS}}\lb f_M(\pi_M)-f_M(\pi)-\frac{1}{2\eta}D_{\textup{H}}^2(M(\pi),\bar{M}(\pi))\rb$ is DEC of $\textup{conv}(\M)$ for the Thompson Sampling strategy $p^{\textup{TS}}(\pi)=\mu(\{M:\pi_M=\pi\})$.
\end{theorem}
For $K-$armed bandits, APS achieves the near-optimal regret $O(\sqrt{KT\log K})$ because $\ir_{\textup{H}}(\text{TS})\leq 4K$; for $d-$dimensional linear bandits, APS recovers the optimal regret $O(\sqrt{d^2T})$ because $\ir_{\textup{H}}(\textup{TS})\leq 4d$. See Appendix \ref{appendix information ratio H} for the proof of these information ratio bounds.

The main messages about APS and Theorem \ref{thm adaptive posterior sampling} are: 1) the regret bound of APS is no worse than the standard regret bound of TS \citep{russo2016information}, but in contrast to the latter, does not rely on any
knowledge needed to specify a prior!  2) Because APS only keeps the marginal beliefs of $\pi^*$ but forgets beliefs of the models, it is robust to adversarial and non-stationary environments. And 3) Experimental results in Section \ref{sec Bernoulli MAB} show that APS achieves ``best-of-all-worlds'' empirical performance for Bernoulli MAB in different environments.

 To the best of our knowledge, Theorem \ref{thm adaptive posterior sampling} is the first generic result to make TS prior-free and applicable to adversarial environment. To that end, we note that Corollary 19 in \cite{lattimore2021mirror} only applies to $K-$armed bandits because of their truncation procedure.

\subsection{Adaptive Minimax Sampling (AMS)} \label{subsec AMS}
When the agent selects decision $p_t$ by solving the minimax problem
\begin{align*}
    \inf_{p_t}\sup_{\nu}\texttt{\textup{AIR}}_{q_t, \eta}(p,\nu),
\end{align*}
and optimizes for algorithmic beliefs as in Algorithm \ref{alg: frequentist sampling}, we call the resulting algorithm ``Adaptive Minimax Sampling'' (AMS).
\begin{algorithm}[htbp]
\caption{Adaptive Minimax Sampling (AMS-EBO)}
\label{alg: adaptive minimax sampling}
Input learning rate $\eta>0$. 

Initialize $q_1=\text{Unif}(\Pi)$. 
\begin{algorithmic}[1]
\FOR{round $t=1,2,\cdots, T$}
    \STATE{{ Find a distribution $p$ of $\pi$ and a distribution $\nu_t$ of $(M, \pi^*)$  that solves the saddle point of }
    \begin{align*}
       \inf_{p\in \Delta(\Pi)}\sup_{\nu\in\Delta(\M\times \Pi)} \texttt{AIR}_{q_t,\eta}(p, \nu).
    \end{align*}}
    \STATE{ Sample decision $\pi_t\sim p_t$ and observe  $o_t\sim M_t(\pi_t)$.}
    \STATE{ Update $q_{t+1}={(\nu_t)}_{\pi^*|\pi_t,o_t}$.}
\ENDFOR
\end{algorithmic}
\end{algorithm}
 By the regret bound \eqref{eq: thm regret} in Theorem \ref{thm regret} and  the upper bounds of AIR by DEC and IR established in Section \ref{subsec air ir dec}, it is straightforward to prove the following theorem.
\begin{theorem}[Regret of AMS]\label{thm adaptive minimax sampling} For a finite decision space $\Pi$ and a compact model class $\M$, the regret of Algorithm \ref{alg: adaptive minimax sampling} with any $\eta>0$ is always bounded as follows, for all $T\in\N$, 
\begin{align}\label{eq: AMS DEC}
  \reg_T\leq  \frac{\log |\Pi|}{\eta}+\dec_{\eta}^{\KL}(\textup{conv}(\M))\cdot T.
\end{align} In particular, the regret of Algorithm \ref{alg: adaptive minimax sampling} with  $\eta= 2\sqrt{\log|\Pi|/(\texttt{\textup{IR}}(\textup{IDS})\cdot T)}$ and all $T\in\N$ is bounded by 
\begin{align*}
  \reg_T\leq  \sqrt{\log |\Pi|\cdot \texttt{\textup{IR}}(\textup{IDS})\cdot T},
\end{align*}
where $\texttt{\textup{IR}}(\textup{IDS}):=\sup_{\nu}\inf_p \ir(\nu,p)$ is the  maximal information ratio of Information-Directed Sampling (IDS). 
\end{theorem}
Theorem \ref{thm adaptive minimax sampling} shows that the regret bound of AMS is always no worse than that of IDS \citep{russo2014learning}. Algorithm \ref{alg: adaptive minimax sampling} is implicitly equivalent with a much simplified implementation of the EBO  algorithm from \cite{lattimore2021mirror}, but this implementation runs in computationally tractable spaces (rather than intractable functional spaces) and does not require limit analysis. We propose the term ``AMS'' with the intention of describing the algorithmic idea broadly, encompassing its various variants, including the important one we will discuss shortly, Model-index AMS. Alternatively, the EBO perspective is convenient and essential when one wishes to establish connections with mirror descent and analyze general Bregman divergences, as we do in Appendix \ref{appendix thm approx bregman}.

In Section \ref{sec mair}, we develop a model-index version of AMS in the stochastic setting, which we term ``Model-index AMS'' (MAMS) and introduce as Algorithm \ref{alg: model adaptive minimax sampling}.  Model-index AMS leverage MAIR as the optimization objective in the algorithmic belief generation. The regret of MAMS is always bounded by 
\begin{align}\label{eq: MAMS DEC pre}
  \reg_T\leq  \frac{\log |\M|}{\eta}+\dec_{\eta}^{\KL}(\M)\cdot T.
\end{align}
Compared with the regret bound of AMS in  \eqref{eq: AMS DEC},  the regret bound \eqref{eq: MAMS DEC pre} of MAMS uses DEC of the original model class $\M$ rather than its convex hull $\textup{conv}(\M)$; on the other hand, the estimation complexity term in \eqref{eq: MAMS DEC pre} is $\log|\M|/\eta$, which is larger than the $\log|\Pi|/\eta$ term in \eqref{eq: AMS DEC}. Furthermore, MAMS is only applicable to the stochastic environment, while AMS is applicable to the adversarial environment. We refer to Section \ref{subsec generic MAIR} for a more detailed comparison between AIR and MAIR.

\subsection{Using approximate maximizers}\label{subsec approximate belief}
In Algorithm \ref{alg: frequentist sampling}, we ask for the algorithmic beliefs to maximize AIR. In order to give computationally efficient algorithms in practical applications (MAB, linear bandits,
RL, ...), we will require the algorithmic beliefs  to approximately maximize AIR. This argument is made rigorous in the following theorem, which uses the first-order optimization error of AIR to represent the regret bound. 

\begin{theorem}[Generic regret bound using approximate maximizers]\label{thm regret approx}Given a finite  $\Pi$, a compact $\M$, an arbitrary algorithm $\alg$ that produces decision probability $p_1,\dots,p_T$, and a sequence of beliefs $\nu_1, \dots, \nu_{T}$ where $q_{t}={(\nu_{t-1})}_{\pi^*|\pi,o}\in \inte(\Delta(\Pi))$ for all rounds, we have
\begin{align*}
   \reg_T\leq \frac{\log |\Pi|}{\eta}+ \E\lb \sum_{t=1}^T 
   \lp\texttt{\textup{AIR}}_{q_t,\eta}(p_t,\nu_t)+\sup_{\nu^*}\lla \left. \frac{\partial \air_{q_t,\eta}(p_t,\nu) }{\partial \nu} \right\vert_{\nu=\nu_t}, \nu^*-{\nu_t} \rra\rp\rb.
\end{align*}
\end{theorem}
Thus we give a concrete approach towards systematical algorithms with rigorous guarantees---by minimizing the gradient of AIR, we aim to approximately maximize AIR. This is an important factor upon which we advance the existing literature in EBO and DEC. We not only provide regret guarantees for ``minimax algorithm'' and ``worst-case environment,'' but also offer a comprehensive path for designing algorithms by leveraging the key idea of ``algorithmic beliefs.'' An analogy of this theorem leveraging MAIR in the stochastic setting is presented as Theorem \ref{thm regret approx stochastic} in Section \ref{sec mair}.

\section{Application to Bernoulli MAB}\label{sec Bernoulli MAB}
Our Bayesian design principles  give rise to a novel algorithm for the Bernoulli multi-armed bandits (MAB) problem. It is well-known that every bounded-reward MAB problem can equivalently be reduced to the Bernoulli MAB problem, so our algorithm and experimental results actually apply to all bounded-reward MAB problems. The  reduction is very simple: assuming the rewards are always bounded in $[a,b]$, then after receiving $r_t(\pi_t)$ at each round, the agent re-samples a binary reward $\tilde{r}_t(\pi_t)\sim\text{Bern}((r_t(\pi_t)-a)/b-a)$ so that $\tilde{r}_t(\pi_t)\in\{0,1\}$.

\subsection{Simplified APS for Bernoulli MAB}\label{subsec APS BMAB}

In Example \ref{example mab},  $\Pi=[K]=\left\{1,\cdots, K\right\}$ is a set of $K$ actions, and each model $\M$ is a mapping from actions to Bernoulli distributions. 
Given belief $\nu\in\Delta(\M\times[K])$, 
we introduce the following parameterization: $\forall i,j\in [K]$,
\begin{align*}
    \theta_{i}(j)&:=\E\lb r(j)|\pi^*=i \rb,  &\text{(conditional mean reward)}
    \\\alpha(i)&:={\nu_{\pi^*}}(i),  &\text{(marginal belief)}\\
    \beta_{i}(j)&:= \alpha(i)\cdot \theta_{i}(j). & \text{(guarantees concavity)}
\end{align*} 
Then we have a concave parameterization of AIR by the $K(K+1)-$dimensional vector $(\alpha, {{\beta}})=(\alpha, \beta_1,\cdots, \beta_K)$: 
\begin{align*}
      \air_{q,\eta}(p,\nu)= \sum_{i\in [K]}\beta_{i}(i)-\sum_{i,j\in[K]}p(j)\beta_i(j)\nonumber\\-\frac{1}{\eta}\sum_{i,j\in [K]}p(j)\alpha(i)\kl\lp\frac{\beta_i(j)}{\alpha(i)}, \sum_{i\in[K]}{\beta_i}(j)\rp-\frac{1}{\eta}\KL(\alpha,q),
\end{align*}
where  $\kl(x,y):= x\log \frac{x}{y}+(1-x)\log\frac{1-x}{1-y}$ for all $x,y\in(0,1)$. By setting the gradients of AIR with respect to all $K^2$ coordinates in $\beta$ to be exactly zero, and choosing $\alpha=p$ (which results in the gradient of AIR with respect to $\alpha$ being suitably bounded), we are able to write down a simplified APS algorithm in closed form (see Algorithm \ref{alg: BMAB}). We apply Theorem \ref{thm regret approx} to show that the algorithm achieves optimal $O(\sqrt{KT\log K})$ regret in the general adversarial setting. 
\begin{theorem}[Regret of Simplified APS for Bernoulli MAB]\label{thm BMAB}The regret of Algorithm \ref{alg: BMAB} with $\eta= \sqrt{\log K/(2KT+4T)}$ and $T\geq 3$ is bounded by
\begin{align*}
    \reg_T\leq 2\sqrt{2(K+2)T\log K}.
\end{align*}
\end{theorem}
We leave the detailed derivation and analysis of the Algorithm \ref{alg: BMAB} to Appendix \ref{appendix APS BMAB}.

\begin{algorithm}[htbp]
\caption{Simplified APS for Bernoulli MAB}
\label{alg: BMAB}
Input learning rate $\eta>0$. 

Initialize $p_1=\text{Unif}(\Pi)$.  
\begin{algorithmic}[1]
\FOR{round $t=1,2,\cdots, T$}
    \STATE{ Sample action $\pi_t\sim p_t$ and receives $r_t(\pi_t)$.}
    \STATE{ Update $p_{t+1}$ by
    \begin{align*}
    p_{t+1}(\pi_t)&=\begin{cases}\frac{1-\exp(-\eta)}{1-\exp(-\eta/p_t(\pi_t))}, & \text{if $r_t(\pi_t)=1$}\\
    \frac{1-\exp(\eta)}{1-\exp(\eta/p_t(\pi_t))}, & \text{if $r_t(\pi_t)=0$}\end{cases}, \text{ and}\\
    {p_{t+1}(\pi)}&= p_t(\pi)\cdot\frac{1-{p}_{t+1}(\pi_t)}{1-p_{t}(\pi_t)}, \quad \forall\pi\neq \pi_t. 
\end{align*}}
\ENDFOR
\end{algorithmic}
\end{algorithm}
At each round, Algorithm \ref{alg: BMAB} increases the weight of the selected action $\pi_t$ if $r_t(\pi_t) = 1$, and decreases the weight if
$r_t(\pi_t) = 0$. The algorithm also maintains the ``relative weight'' between all unchosen actions $\pi \neq \pi_t$, allocating probabilities to these actions proportionally to $p_t$. Algorithm \ref{alg: BMAB} is clearly very different from the well-known EXP3 algorithm, which instead updates $p_{t+1}$ by the formula
\begin{align*}
    p_{t+1}(\pi)=p_t(\pi)\exp\left(\eta\cdot\frac{r_t(\pi_t)\mathds{1}\{\pi=\pi_t\}}{p_t(\pi_t)}\right), \quad\forall \pi\in\Pi.
\end{align*}
In Section \ref{subsec linear bandits} we recover a modified version of EXP3 by Bayesian principle assuming the reward distribution is exponential or Gaussian. 
We conclude that Algorithm \ref{alg: adaptive posterior sampling} uses a precise posterior for Bernoulli
reward, while EXP3 estimates worst-case exponential or Gaussian reward. This may explain why Algorithm \ref{alg: BMAB} performs much better in all of our experiments.

 \paragraph{Reflection symmetry and mirror map:}
We would like to highlight a notable property of our new algorithm: Algorithm \ref{alg: BMAB} exhibits {\it reflection symmetry} (or {\it reflection invariance}), signifying that when we interchange the roles of reward and loss, the algorithm remains unchanged. The theoretical convergence of our algorithm is not contingent on any truncation. On the contrary, the traditional IPW estimator does not exhibit this property: the reward-based IPW and the loss-based IPW yield distinct algorithms. While the expected regret convergence of EXP3 with the loss-based estimator does not necessitate truncation, EXP3 with the reward-based estimator mandates truncation for convergence. (The original EXP3 algorithm from \cite{auer2002nonstochastic} is reward-based; we refer to \S 11.4 in \cite{lattimore2020bandit} for loss-based EXP3.) 

We would also like to comment on the relationship between Algorithm \ref{alg: BMAB} and mirror descent. Algorithm \ref{alg: BMAB} is equivalent to running mirror descent (specifically exponential weight here) using the reward estimator:
\begin{align}\label{eq: estimator bernoulli}
    \hat{r}_t(\pi)&= \frac{1}{\eta}\log\frac{{(\nu_t)}_{\pi^*|\pi_t,r_t(\pi_t)}(\pi)}{p_t(\pi)}\nonumber\\&=
    \frac{1}{\eta}\log\frac{{(\nu_t)}_{\pi^*|\pi_t,0}(\pi)}{p_t(\pi)}+\frac{\mathds{1}(\pi=\pi_t)r_t(\pi_t)}{p_t(\pi)}-r_t(\pi_t).
\end{align}
From this perspective, Algorithm \ref{alg: BMAB} can be understood as addressing a long-standing question: how to fix the issue of exploding variance in the IPW estimator and automatically find the optimal bias-variance trade-off. It is worth noting that the derivation of our algorithms reveals a deep connection between frequentist estimators and the natural parameters of exponential family distributions (see Appendix \ref{subsec exponential family} for details).
We hope this framework can be applied to challenging problems where a more robust estimator than IPW is required, such as bandit convex optimization \citep{bubeck2017kernel} and efficient linear bandits \citep{abernethy2008competing}.

\subsection{Numerical experiments}\label{subsec experiments}
We implement Algorithm \ref{alg: BMAB} (with the legend ``APS'' in the figures) and compare it with other algorithms across the stochastic, adversarial and non-stationary environments (code available {\hypersetup{urlcolor=magenta}\href{https://github.com/xuyunbei/MAB-code/tree/main}{here}}). We plot expected regret (average of 100 runs) for different choices of $\eta$, and set a forced exploration rate $\gamma=0.001$ in all experiments. We find APS 1) outperforms UCB and matches TS in the stochastic environment; 2) outperforms EXP3 in the adversarial environment; and 3) outperforms EXP3 and is comparable to the ``clairvoyant'' benchmarks (that have prior knowledge of the changes) in the non-stationary environment. For this reason we say Algorithm \ref{alg: BMAB} (APS) achieves the ``best-of-all-worlds'' performance. We note that the optimized choice of $\eta$ in APS differ instance by instance, but by an initial tuning we typically see good results, whether we tune $\eta$ optimally or not optimally.

\subsubsection{Stochastic Bernoulli MAB}
\begin{figure}

    \begin{center}
  \centerline{\includegraphics[width=0.6\linewidth]{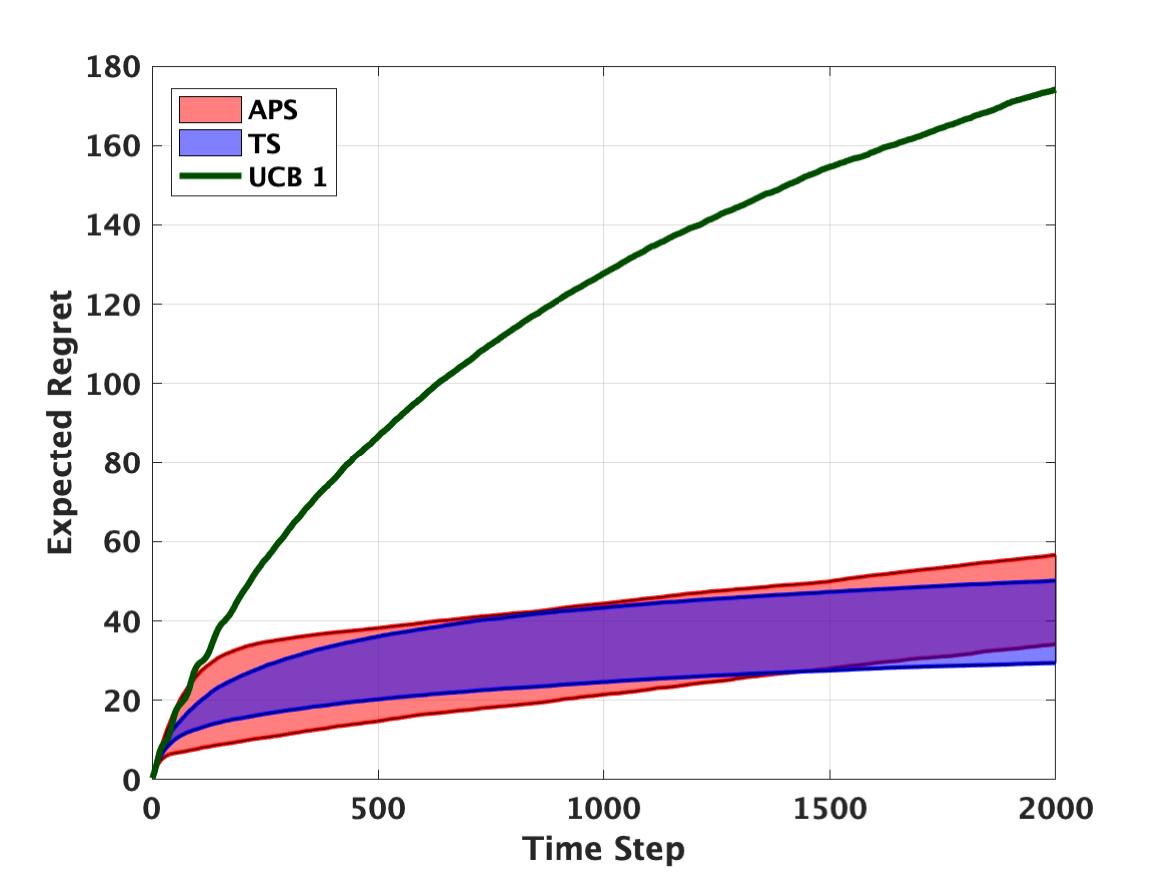}}
  
  \caption{Sensitivity analysis in a stochastic bandit problem.}
  \label{fig:SMAB}
\end{center}

\end{figure}
In Figure \ref{fig:SMAB} we report the expected regret for APS with different choices of $\eta$, TS with different Beta priors, and the UCB 1 algorithm, in a stochastic 16-armed Bernoulli bandit problem. We refer to this as ``sensitivity analysis'' because the red, semi-transparent, area reports the regret of APS when learning rates $\eta$ are chosen across a range of values drawn from the interval $[0.05,0.5]$ (the interval is specified by an initial tuning); and the priors of TS are chosen from Beta$(c,1)$ where $c\in[0.5,5]$. In particular, the bottom curve of the red (or blue) area is the regret curve of APS (or TS) using optimally tuned $\eta$ (respectively,  prior). The conclusion is that APS outperforms UCB 1, and is comparable to TS in this stochastic environment.

\subsubsection{Adversarial Bernoulli MAB}
\begin{figure}

\end{figure}

\begin{figure}
\centering
\begin{minipage}{.4\textwidth}
  \centering
  \includegraphics[width=1\linewidth]{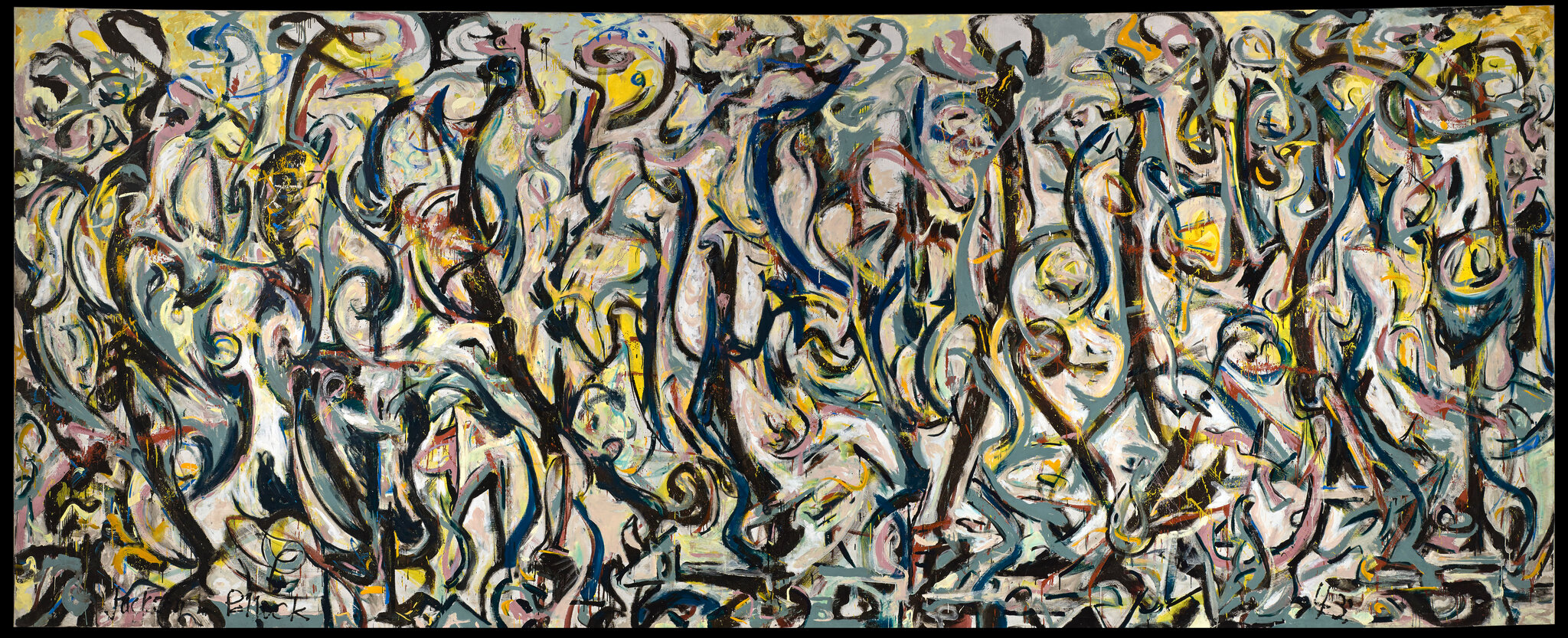}
  \caption{Jackson Pollock, {\it Mural} (1943) [Oil on canvas]. University of Iowa Museum of Art.}
  \label{fig:Pollock}
\end{minipage}%
\begin{minipage}{.6\textwidth}
  \centering
  \includegraphics[width=1\linewidth]{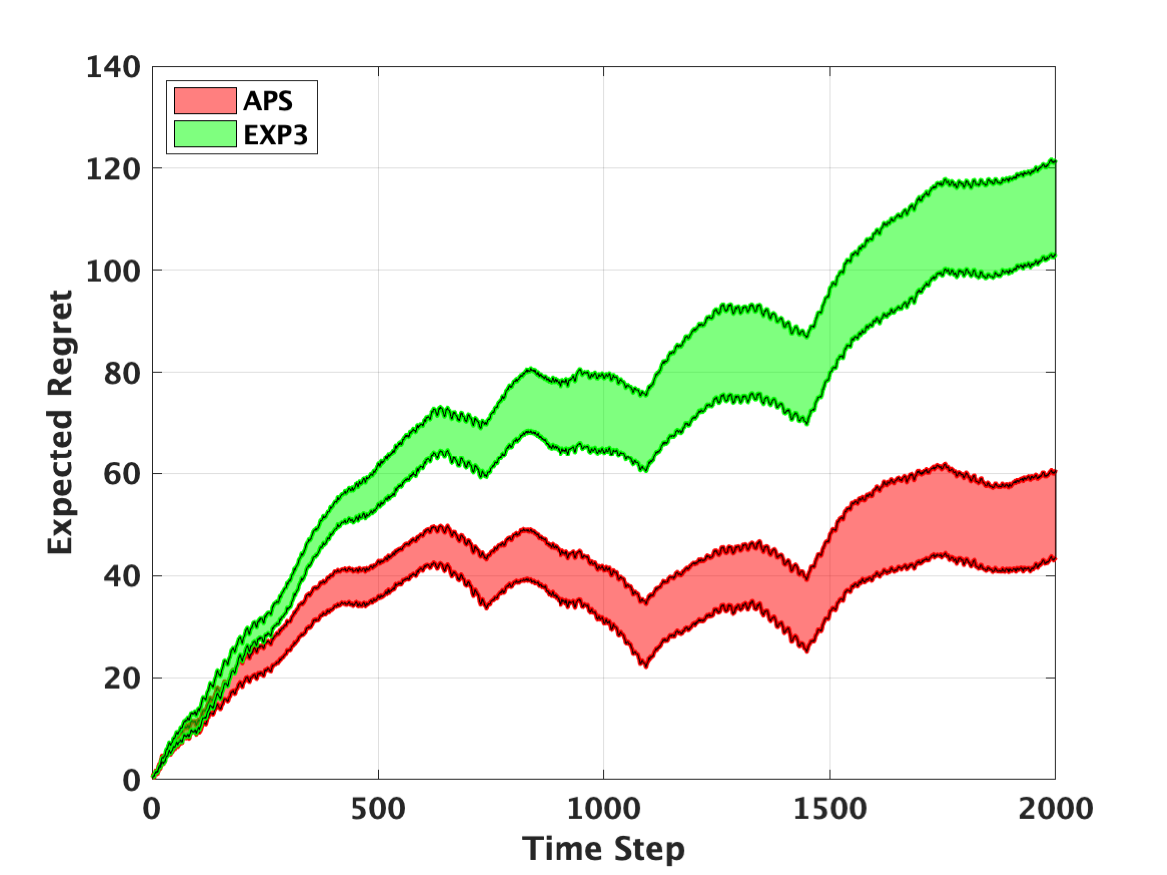}
  \caption{Sensitivity analysis in an adversarial bandit problem.}
  \label{fig:AMAB}
\end{minipage}
\end{figure}

 We equidistantly take 16 horizontal lines from an abstract art piece by Jackson Pollock to simulate the rewards (pre-specified) in an  adversarial environment, and study this via a 16-armed bandit problem. To be more specific, we transform these 16 horizontal lines into grayscale values, resulting in each line becoming a sequence of 2000 values in the range [0, 1]. As depicted in Figure \ref{fig:Pollock}, it's evident that the environment is inherently adversarial and lacks any statistical regularity. This is further confirmed by empirical findings that the best-in-hindsight arm changes over time. Figure \ref{fig:AMAB} shows the sensitivity analysis for  APS and EXP3 when both the learning rates are chosen from $[0.1,5]$ (the interval is specified by an initial tuning). In particular, the red and green lower curves compare the optimally tuned versions of APS and EXP3. The conclusion is that APS outperforms EXP3 whether $\eta$ is tuned optimally or not.

\subsubsection{Non-stationary Bernoulli MAB (with change points)}

\begin{figure}
\centering
\begin{minipage}{.5\textwidth}
  \centering
  \includegraphics[width=1\linewidth]{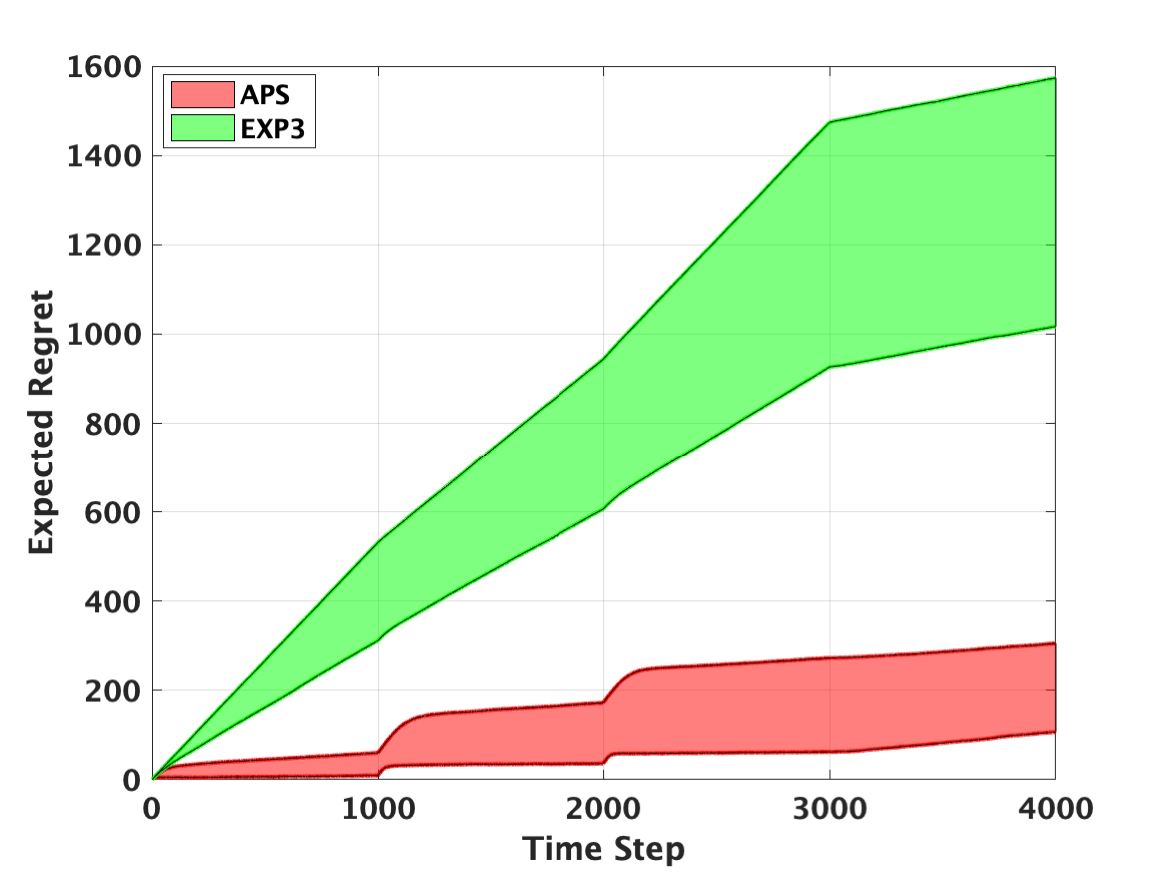}
  \caption{Sensitivity analysis in a\\ ``change points'' environment.}
  \label{fig:NS1}
\end{minipage}%
\begin{minipage}{.5\textwidth}
  \centering
  \includegraphics[width=1\linewidth]{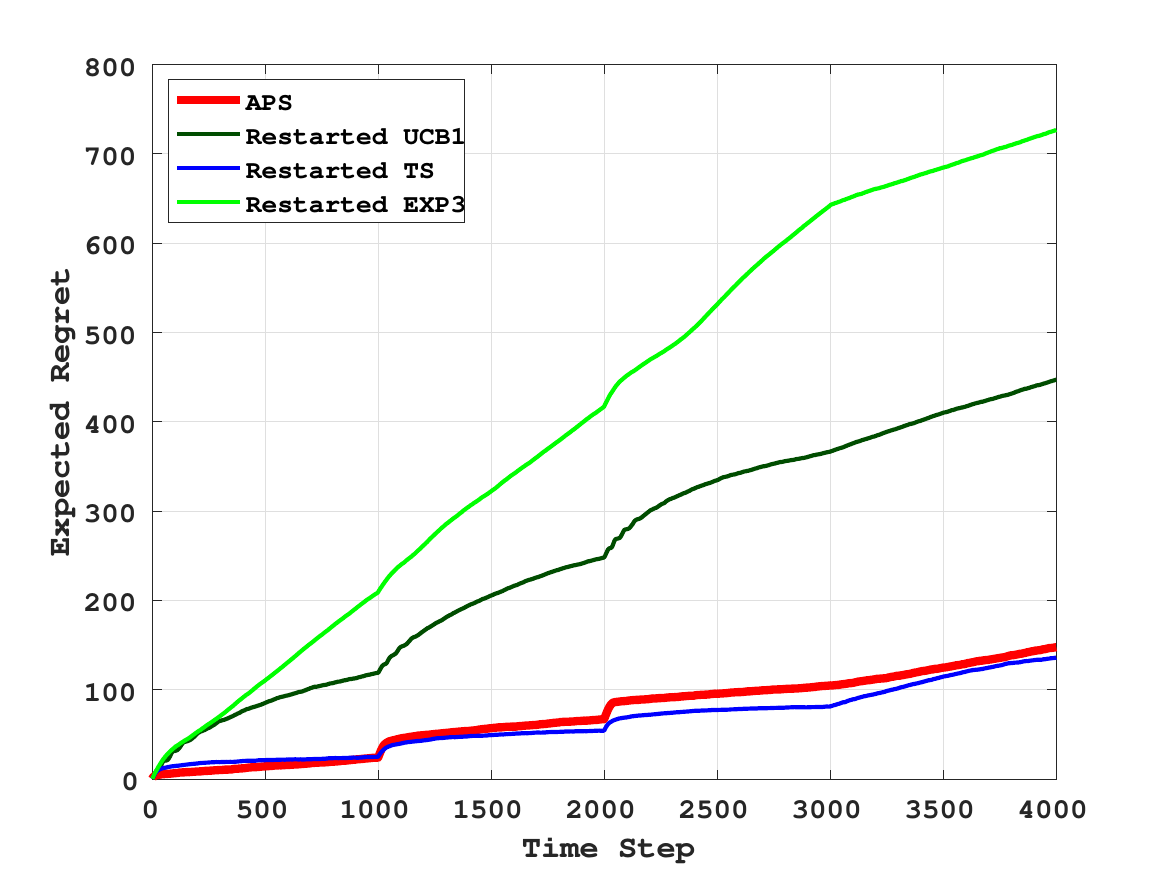}
  \caption{Comparing APS to ``clairvoyant''\\ restarted algorithms in a ``change points''\\ environment.}
  \label{fig:NS2}
\end{minipage}
\end{figure}

We study a 16-armed Bernoulli bandit problem in a non-stationary environment. We generate 4 batches of i.i.d. sequences, where the changes in the environment occur  after round 1000, round 2000, and round 3000. We consider a stronger notion of regret known as the dynamic regret \citep{besbes2014stochastic}, which compares the cumulative reward of an algorithm to the cumulative reward of the  best non-stationary policy (rather than a single arm) in hindsight. In this particular setting, the benchmark is to select the best arm in all the 4 batches. In Figure \ref{fig:NS1} we perform sensitivity analysis for APS and EXP3, where the learning rates are chosen across $[0,05,5]$. 
Since the agent will not know when and how the adversarial environment changes in general, it is most reasonable to compare APS with EXP3 without any knowledge of the environment as in Figure \ref{fig:NS1}. We observe that APS  dramatically improves the dynamic regret by several times.

In Figure \ref{fig:NS2}, we compare APS to three ``clairvoyant'' restarted algorithms, which require knowing that the environment consists of 4 batches of i.i.d. sequences, as well as knowing the exact change points. We tune the parameters in these algorithms optimally. Without knowledge of the environment, APS performs better than restarted EXP3 and restarted UCB 1, and is comparable to restarted TS. (It is important to emphasize again that the latter algorithms are restarted based on foreknowledge of the change points.)

\subsubsection{Non-stationary Bernoulli MAB (with ``sine curve'' reward sequences)}
\begin{figure}
\centering
\begin{minipage}{.5\textwidth}
  \centering
  \includegraphics[width=1\linewidth]{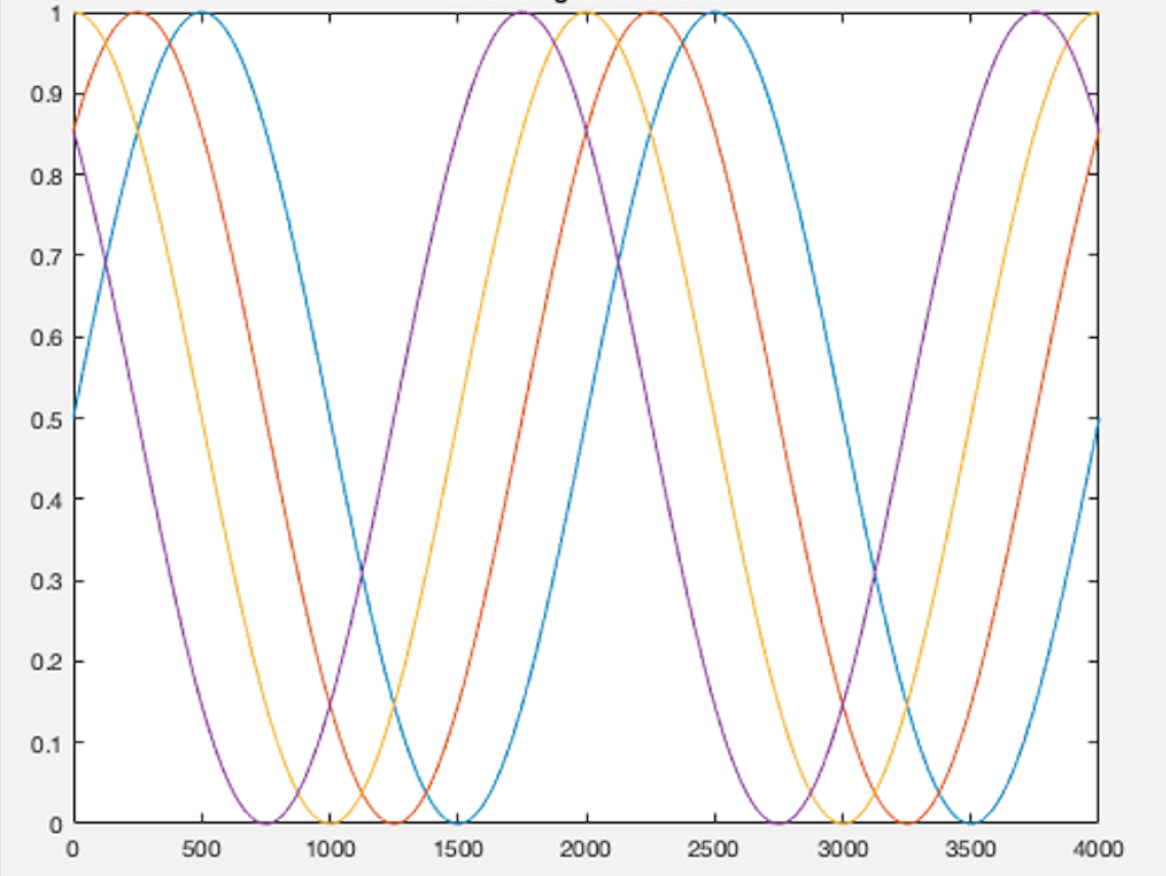}
  \caption{``Sine curve'' reward sequences \\for 4 arms.}
  \label{fig:sine1}
\end{minipage}%
\begin{minipage}{.5\textwidth}
  \centering
  \includegraphics[width=1\linewidth]{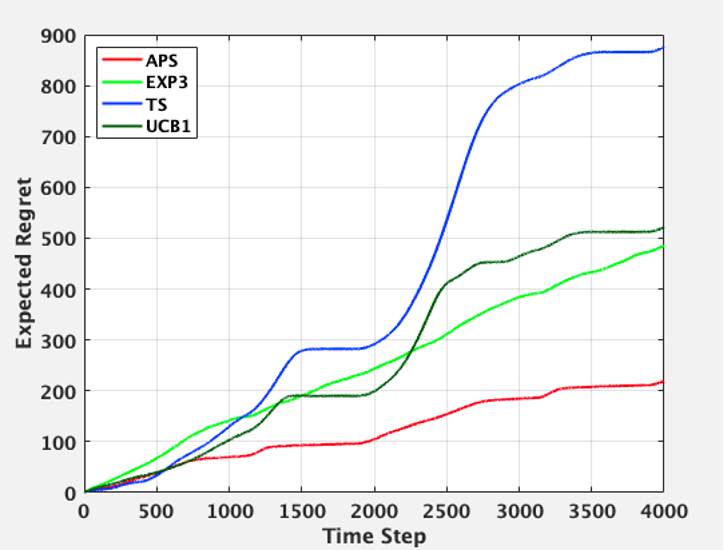}
  \caption{Regret curves in a ``sine curve'' \\environment.}
  \label{fig:sine2}
\end{minipage}
\end{figure}
\begin{figure}
\centering
\begin{minipage}{.5\textwidth}
  \centering
  \includegraphics[width=1\linewidth]{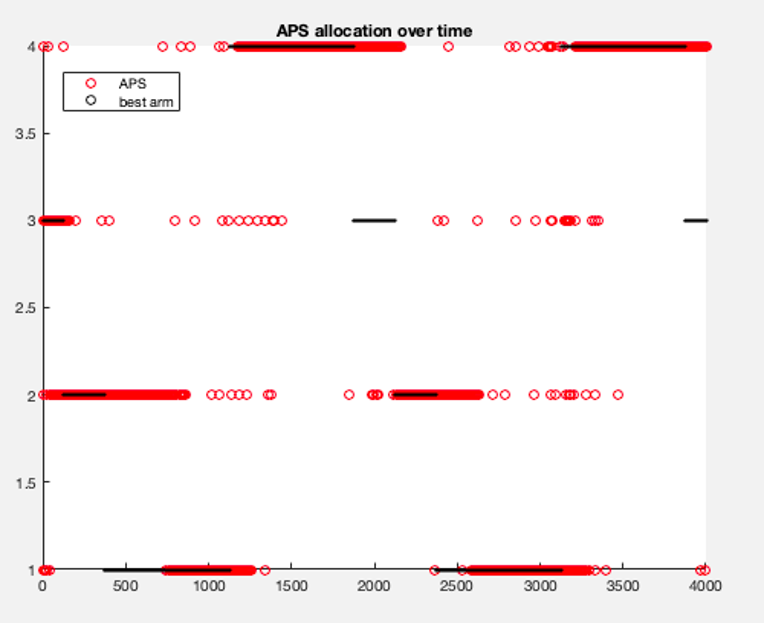}
  \caption{Tracking selected arms of APS\\ in a ``sine curve'' environment.}
  \label{fig:sine3}
\end{minipage}%
\begin{minipage}{.5\textwidth}
  \centering
  \includegraphics[width=1\linewidth]{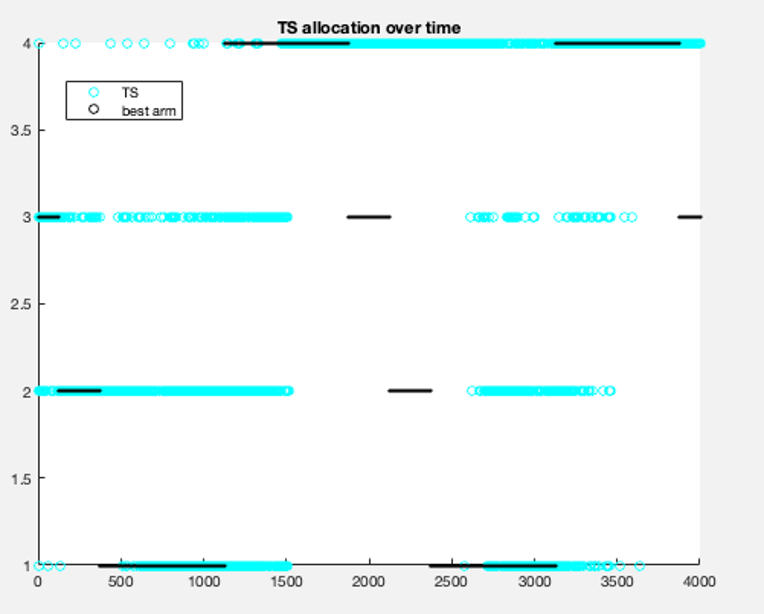}
  \caption{Tracking selected arms of TS\\ in a ``sine curve'' environment.}
  \label{fig:sine4}
\end{minipage}
\end{figure}

We generate a 4-armed bandit problem with the mean-reward structure shown in Figure \ref{fig:sine1}. The four sine curves (with different colors) in Figure \ref{fig:sine1} represent the mean reward sequences of the 4 arms. We tune the parameters in all the algorithms to optimal and report their regret curves in Figure \ref{fig:sine2}. As shown in Figure \ref{fig:sine2}, APS achieves the best performance, while TS fails in this non-stationary environment. This experiment shows the vulnerability of TS if the environment is not stationary, such as the sine curve structure shown here.

To better illustrate the smartness of APS compared with TS in the non-stationary environment, we track the selected arms and the best arms throughout the process. In Figure \ref{fig:sine3} and Figure \ref{fig:sine4}, the horizontal line represents the 4000 rounds, and the vertical lines represent the 4 arms (indexed as 1, 2, 3, and 4). In Figure \ref{fig:sine3}, the red points show the selected arms of APS, and the black points represent the best arms at each round in this ``sine curve'' non-stationary environment. In Figure \ref{fig:sine4}, the blue points show the selected arms of TS. The more consistent the selected arms are with the best arms (black points), the better choices an algorithm makes. Comparing Figure \ref{fig:sine3} and Figure \ref{fig:sine4}, we can see that APS is highly responsive to changes in the best arm, whereas TS is relatively sluggish in this regard. The implication of this experiment is that creating a new algorithmic belief at each round has the potential to significantly improve performance and be a game changer in many problem settings.

We conclude that these experiments provide some numerical evidence indicating that  APS achieves the ``best-of-all-worlds'' across stochastic, adversarial, and non-stationary environments.

\section{Key intuition and underlying theory}
It is worth noting that the proofs of Theorem \ref{thm regret} and its generalizations are quite insightful and parsimonious. The two major steps in the proofs may be interesting on its own. The first step is a succinct analysis to bound the cumulative regret by sum of AIR (see Section \ref{subsec regret AIR}); and the second step is to extend the classical minimax theory of ``exchanging values'' into a constructive approach to design minimax decisions (estimators, algorithms, etc.), which will be presented in Section \ref{subsec minimax theory}. At the end, we also illustrate how such analysis generalizes to approximate algorithmic beliefs, as presented in Theorem \ref{thm regret approx} and Theorem \ref{thm regret approx stochastic}. 

\subsection{Bounding regret by sum of AIR}\label{subsec regret AIR}
For every $\bar{\pi}\in\Pi$, we have 
\begin{align}\label{eq: telescope}
    \sum_{t=1}^T \left[\log\frac{q_{t+1}(\bar{\pi})}{q_t(\bar{\pi})}\right]= \log\frac{q_T(\bar{\pi})}{q_1(\bar{\pi})}\leq \log |\Pi|.
\end{align}
 Taking $q_{t+1}={(\nu_t)}_{\pi^*|\pi_t,o_t}$ as in Algorithm \ref{alg: frequentist sampling}, and taking conditional expectation on the left hand side of \eqref{eq: telescope}, we have
\begin{align}\label{eq: expectation telescope}
     \E\lb\sum_{t=1}^T \E_{t-1}\left[\log\frac{(\nu_t)_{\pi^*}(\bar{\pi}|\pi_t,o_t)}{q_t(\bar{\pi})}\right]\rb\leq \log |\Pi|,
\end{align}
where the conditional expectation notation is introduced in Section \ref{subsec problem formulation}.
By subtracting the additive elements on the left-hand side of \eqref{eq: expectation telescope} (divided by $\eta$) from the per-round regrets against $\bar{\pi}$, we obtain
\begin{align}
&\E\lb \sum_{t=1}^T f_{M_t}(\bar{\pi})-\sum_{t=1}^T f_{M_t}(\pi_t)- \frac{1}{\eta}\cdot \sum_{t=1}^T \E_{t-1}\left[\log\frac{(\nu_t)_{\pi^*}(\bar{\pi}|\pi_t,o_t)}{q_t(\bar{\pi})}\right]\rb\nonumber\\=&\E\lb\sum_{t=1}^T \E_{t-1}\left[ f_{M_t}(\bar{\pi})-f_{M_t}(\pi_t)-\frac{1}{\eta}\log\frac{{(\nu_t)}_{\pi^*|\pi_t,o_t}(\bar{\pi})}{q_t(\bar{\pi})}\right]\rb\nonumber\\
    \overset{(\diamond)}{\leq} & \E\lb\sum_{t=1}^T \sup_{M, \bar{\pi}}\E_{t-1}\left[f_{M}(\bar{\pi})-f_{M}(\pi_t)-\frac{1}{\eta}\log\frac{{(\nu_t)}_{\pi^*|\pi_t,o_t}(\bar{\pi})}{q_t(\pi^*)}\right]\rb\nonumber\\
    \overset{(*)}{=} &  \E\lb\sum_{t=1}^T \texttt{\textup{AIR}}_{q_t,\eta}(p_t,\nu_t)\rb,\label{eq: identity Nash}
\end{align}
where the inequality is by taking supremum at each rounds; and the last equality $(*)$ in \eqref{eq: identity Nash} is by Lemma \ref{lemma identity Nash}, an important identity to be explained in Section \ref{subsec minimax theory}, which is derived from the fact that the pair of maximizer $\nu_t$ and posterior functional is a Nash equilibrium of a convex-concave function. We note that the second-to-last formula $(\diamond)$ is true without imposing any restrictions on $\nu_t$; only the last formula $(*)$ relies on the fact that $\nu_t$ maximizes AIR.

Combining \eqref{eq: identity Nash} and  \eqref{eq: expectation telescope}, we obtain the following inequality for every $\bar{\pi}\in\Pi$:
\begin{align*}
\E\left[\sum_{t=1}^T f_{M_t}(\bar{\pi})-\sum_{t=1}^T f_{M_t}(\pi_t)\right]- \frac{\log|\Pi|}{\eta} \leq \E\lb\sum_{t=1}^T \texttt{\textup{AIR}}_{q_t,\eta}(p_t,\nu_t)\rb.
\end{align*}
By taking the supremum over $\bar{\pi}\in \Pi$ on the left-hand side of the inequality, we are able to prove Theorem \ref{thm regret}:
\begin{align*}
\reg_T\leq \frac{\log|\Pi|}{\eta}+\E\lb\sum_{t=1}^T \texttt{\textup{AIR}}_{q_t,\eta}(p_t,\nu_t)\rb.
\end{align*}

\subsection{Duality theory: from value to construction}\label{subsec minimax theory}
Consider a decision space $\X$, a space $\Y$ of the adversary's outcome, and a convex-concave function $\psi(x,y)$ defined in $\X\times \Y$. The classical minimax theorem \citep{sion1958general} says that, under regularity conditions, the minimax and maximin values of $\psi(x,y)$ are equal:
\begin{align*}
    \min_{\X}\max_{\Y} \psi(x,y)= \max_{\Y}\min_{\X} \psi(x,y).
\end{align*}
We refer to  $\arg\min_{\X}\max_{\Y}\psi(x,y)$ as the set of ``minimax decisions,'' as they are optimal in the worst-case scenario. And we say $\tilde{x}\in\arg\min_{\X}\psi(x,\bar{y})$ is ``maximin decision'' if  $\bar{y}\in\arg\max_{\Y}\min_{\X}\psi(x,y)$ is ``maximin adversary's outcome.''  One natural and important question is, {\it when will the ``maximin decision'' $\tilde{x}$ also be a ``minimax decision?''} Making use of strong convexity, we extends the classical minimax theorem for values into the following theorem for constructive minimax solutions: 

\begin{lemma}[A constructive minimax theorem for minimax solutions]\label{lemma minimax decision}
Let $\X$ and $\Y$ be convex and compact sets, and $\psi: \X\times \Y\rightarrow \R$ a function which for all $y$ is strongly convex and continuous in $x$ and for all $x$ is concave and continuous in $y$. For each $y\in\Y$, let $x_y=\min_{x\in\X} \psi(x,y)$ be the corresponding unique minimizer.
Then, by maximizing the concave objective 
\begin{align*}\bar{y} \in \arg\max_{y\in\Y}\psi(x_y,y),\end{align*} we can establish that $x_{\bar{y}}$ is a minimax solution of $\min_{x\in\X}\max_{y\in\Y}\psi(x,y)$ and $(x_{\bar{y}},\bar{y})$ is a Nash equilibrium.
\end{lemma}
The above lemma, although straightforward to prove through the classical minimax theorem, is conceptually interesting because it emphasizes the significance of strong convexity (often achieved through regularization). This strong convexity allows us to obtain ``minimax solutions'' by considering a ``worst-case adversary'' scenario, such as the least favorable Bayesian prior. This approach simplifies the construction of minimax estimators in classical decision theory, moving away from the need to consider ``least favorable prior sequence'' and specific losses by imposing regularization.

Applying  Lemma \ref{lemma minimax decision} to our framework, we can show:  1) Bayesian posterior $\nu_{\pi^*|\pi,o}$ is the optimal functional to make decision under belief $\nu$; and 2) by choosing worst-case belief $\bar{\nu}$, we construct a Nash equilibrium. As a result, we can establish the following per-round identity, which leads to the key identity denoted as $(*)$ in \eqref{eq: identity Nash}. This identity plays a crucial role in the proof of Theorem \ref{thm regret}.

\begin{lemma}[Identity by Nash equilibrium]\label{lemma identity Nash}
Given $q\in \inte(\Delta(\Pi))$, $\eta>0$ and $p\in \Delta(\Pi)$, denote $\bar{\nu}\in \Delta(\M\times\Pi)$. Then we have 
\begin{align*}
    \sup_{M, \bar{\pi}}\E_{\pi\sim p,o\sim M(\pi)}\left[f_M(\bar{\pi})-f_M(\pi)-\frac{1}{\eta}\log\frac{{\bar{\nu}}_{\pi^*|\pi,o}(\bar{\pi})}{q(\bar{\pi})}\right]=\texttt{\textup{AIR}}_{q, \eta}(p, \bar{\nu}).
\end{align*}
\end{lemma}

Furthermore, we establish the following lemma, demonstrating that the role of AIR and its derivatives is generic for any belief selected by the agent and any environment specified by nature. Recall that the second-to-last formula $(\diamond)$ in \eqref{eq: identity Nash} holds without imposing any restrictions on $\nu_t$. Thus, Lemma \ref{lemma identity approx} leads us to Theorem \ref{thm regret approx} and Theorem \ref{thm regret approx stochastic}. This fact is proven through Danskin's theorem \citep{bernhard1995theorem} (Lemma \ref{lemma duality gradients}), which establishes equivalence between the gradient of the optimized objective and the partial derivative at the optimizer. The ``identity'' nature of Lemma \ref{lemma identity approx} indicates that regret analysis using derivatives of AIR can be principled, precise, and reliable for arbitrary algorithmic beliefs (in addition to maximizers) and any type of environment.
\begin{lemma}[Identity for arbitrary algorithmic belief and any environment]\label{lemma identity approx}
Given $q\in \inte(\Delta(\Pi))$, $\eta>0$  $p\in \Delta(\Pi)$, and an arbitrary algorithmic belief $\hat{\nu}\in \Delta(\M\times\Pi)$ selected by the agent and an arbitrary environment $(M, \bar{\pi})$ specified by the nature. Then we have 
\begin{align*}
    \E_{\pi\sim p,o\sim M(\pi) }\left[f_M(\bar{\pi})-f_M(\pi)-\frac{1}{\eta}\log\frac{{\hat{\nu}}_{\pi^*|\pi,o}(\bar{\pi})}{q(\bar{\pi})}\right]=\texttt{\textup{AIR}}_{q, \eta}(p, \hat{\nu})+\lla\left.\frac{\partial \air_{q,\eta}(p,\nu)}{\partial \nu}\right\vert_{\nu=\hat{\nu}}, \mathds{1}(M, \bar{\pi})-\hat{\nu}\rra,
\end{align*}
where $\mathds{1}(M,\bar{\pi})$ is a vector in simplex $\Delta(\M\times \Pi)$ where all the weights are given to $(M,\bar{\pi})$.
\end{lemma}

In addition to their utility in upper bound analyses, we hope that the duality methodology presented in this section may also offer values for lower bound analyses. Furthermore, the derivation of our bandit and RL algorithms unveils a rich mathematical structure, which we will demonstrate in Appendix \ref{appendix MAB} and \ref{appendix mair} for more details.
\section{Applications to infinite-armed bandits}\label{sec linear and convex bandits}
Our design principles can be applied in many sequential learning and
decision making environments. In order to maximize AIR in practical applications,  we parameterize the belief $\nu$, and make the gradient of AIR with respect to such parameter small.  We will present our results for linear bandits and bandit convex optimization in this section and present our results for reinforcement learning in Section \ref{sec mair}. We give a high-level overview of the applications to linear bandits and bandit convex optimization here. 

\paragraph{Application to linear bandits.}  A classical algorithm for adversarial linear bandits (described in Example \ref{example structured bandits}) is the EXP2 algorithm \citep{dani2007price}, which uses IPW for linear reward as a black-box estimation method, and combines it with continuous exponential weight. We derive a modified version of EXP2 from our framework, establishing interesting connection between IPW and Bayesian posteriors.

\paragraph{Application to bandit convex optimization.}
Bandit convex optimization (described in Example \ref{example structured bandits}) is a notoriously challenging problem, and much effort has been put to understanding its minimax regret and algorithm design \citep{bubeck2016multi, bubeck2017kernel, lattimore2020improved}. The best known result, which is of order $\tilde{O}(d^{2.5}\sqrt{T})$, is derived through the non-constructive information-ratio analysis in \cite{lattimore2020improved}. As a  corollary of Theorem \ref{thm adaptive minimax sampling},  Adaptive Minimax Sampling (AMS) recovers the best known regret bound with a simple constructive algorithm, which can be computed in $\text{poly}(e^d\cdot T)$ time. To the best of our knowledge, this is the first finite-running-time  algorithm that attains the best known  $\tilde{O}(d^{2.5}\sqrt{T})$ regret.

\subsection{Maximization of AIR for structured bandits}\label{subsec structure of optimization}
Consider the structured bandit problems described in Example \ref{example structured bandits}. We consider the computation complexity of the optimization problem 
\begin{align}\label{eq: computation problem}
    \sup_{\nu\in\Delta(\M\times \Pi)} \texttt{AIR}_{q,\eta}(p, \nu).
\end{align}

The computational complexity of \eqref{eq: computation problem} may be $O(\text{poly}(\exp(\exp(d))))$ in the worst case as the size of $\M\times \Pi$. However, when the mean reward function class $\F$ is a convex function class, the computational complexity will be $O(\text{poly}(|\Pi|))$ which is efficient for $K-$armed bandits and is no more than $O(\text{poly}(e^d))$ in general (by standard discretization and covering arguments, we may assume $\Pi\subset\R^d$ to have finite cardinality $O(e^d)$ for the simplicity of theoretical analysis). Moreover, we also give efficient algorithm for linear bandits with exponential-many actions. We refer to Appendix \ref{subsec concave parameterization} for the detailed discussion on the parameterization method and computational complexity.

\subsection{Application to Gaussian linear bandits}\label{subsec linear bandits}

We consider the adversarial linear bandit problem with Gaussian reward. In such a  problem, $\Pi=\A\subseteq \R^d$ is a convex action set with dimension $d$. The model class $\M$  can be parameterized by a $d-$dimensional  vector $\theta\in \R^d$ that satisfies $\theta^\top a\in[-1,1]$ for all $a\in\A$.  Here we use the notations $\A$ (as action set), $a$ (as action) and $a^*$ (as optimal action) to follow the tradition of literature about linear bandits. The reward $r(a)$ for each action $a\in\A$ is drawn from a Gaussian distribution that has mean $\theta^\top a$ and variance $1$. To facilitate the handling of mixture distributions, we introduce the ``homogeneous noise'' assumption as follows: for all actions $a\in\A$, the reward of $a$ is denoted as $r(a) = \theta^\top a + \epsilon$, where $\epsilon \sim N(0,1)$ is Gaussian noise that is identical across all actions (meaning all actions share the same randomness within the same rounds). This ``homogeneous noise'' simplifies our expression of AIR, and the resulting algorithms remain applicable to independent noise models and the broader sub-Gaussian setting.

As discussed in Section \ref{subsec structure of optimization}, we restrict our attention to sparse $\nu$ where for each $a^*\in\A$ there is only one  model $M$, which corresponds to the Gaussian  distribution $r(a)\sim N( \theta_{a^*}(a),1)$.  We
parameterize the prior $\nu$ by  vectors $\{\beta_a^*\}_{a^*\in\A}$ and $\alpha\in\Delta(\A)$, where $\alpha=\P_{\nu}(a^*)$ and $\beta_{a^*}=\alpha(a^*)\cdot \theta_{a^*}$.
Note that AIR in this ``homogeneous noise'' setting can be expressed as
 \begin{align}\label{eq: air gaussian bandits}
      \oair_{q,\eta}(p,\nu)= &\int_{\A}\beta_{a^*}^\top a^*da^*-\int_{\A}\int_{\A}p(a)\beta_{a^*}^\top a da^*da\nonumber\\&-\frac{1}{2\eta}\int_{\A}\int_{\A}p(a)\alpha(a^*)\lp\frac{\beta_{a^*}^\top a}{\alpha(a^*)}-\int_{\A}\beta_{a^*}^\top a da^*\rp^2da-\frac{1}{\eta}\KL(\alpha,q).
 \end{align}

By setting the gradients of \eqref{eq: air gaussian bandits} with respect to $\alpha$ to zero and the gradients with respect to $\{\beta_{a^*}\}_{a^*\in\A}$ to nearly zero, we obtain an approximate maximizer of AIR in \eqref{eq: air gaussian bandits}. We calculate the Bayesian posterior, and find that the resulting algorithm is  an exponential weight algorithm with a modified IPW estimator: at each round $t$, the agent update $p_{t+1}$ by 
\begin{align*}
    \tilde{p}_{t+1}(a)\propto p_t(a)\exp\lp \eta \hat{r}_t(a)\rp,
\end{align*}
where $\hat{r}_t$ is the modified IPW estimator for linear reward,
\begin{align}\label{eq: IPS GMAB}
    \hat{r_t}(a)=\underbrace{ a^\top(\E_{a\sim p_t}[aa^\top])^{-1}a_t r_t(a_t) }_{\text{IPW estimator}}+\underbrace{\frac{\eta}{2} \lp a( \E_{a\sim p_t} [aa^\top])^{-1} a- (a^\top (\E_{a\sim p_t}[aa^\top])^{-1}a_t)^2\rp}_{\text{mean zero regularizer}} .
\end{align}
Note that in order to avoid boundary conditions in our derivation,  we require forced exploration to ensure $\lambda_{\min}(\E_{a\sim p}[aa^\top])\geq \gamma$. This can be done with the help of the volumetric spanners constructed in \cite{hazan2016volumetric}. The use of volumetric spanner makes our final proposed algorithm (Algorithm \ref{alg: GLB}) to be slightly more involved, but we only use the volumetric spanner in a ``black-box'' manner. 
\begin{algorithm}[htbp]
\caption{Simplified APS for Gaussian linear bandits}
\label{alg: GLB}
Input learning rate $\eta>0$, forced exploration rate $\gamma$, and action set $\A$.

Initialize $p_1=\text{Unif}(\A)$.  
\begin{algorithmic}[1]
\FOR{round $t=1,2,\cdots, T$}
\STATE{Let $S_t'$ be a $(p_t, \exp(-(4\sqrt{d}+\log(2T))))-$exp-volumetic spanner of $\A$, 

Let $S_t''$ be a $2\sqrt{d}-$ratio-volumetric spanner of $\A$. 

Set $S_t$ as the union of $S_t'$ and $S_t''$.}
    \STATE{ Sample action $a_t\sim p_t$ and receives $r_t(a_t)$.}
    \STATE{ Calculate $\tilde{p}_{t+1}$ by
    \begin{align*}
    \tilde{p}_{t+1}(a)\propto p_t(a)\exp\lp \eta \hat{r}_t(a)\rp, 
\end{align*}
where $\hat{r}_t$ is the modified IPW estimator for linear loss, 
\begin{align*}
    \hat{r_t}(a)= a^\top(\E_{a\sim p_t}[aa^\top])^{-1}a_t r_t(a_t) +\frac{\eta}{2} \lp a( \E_{a\sim p_t} [aa^\top])^{-1} a- (a^\top (\E_{a\sim p_t}[aa^\top])^{-1}a_t)^2\rp.
\end{align*}}
\STATE{Update $p_{t+1}$ by $p_{t+1}(a)=(1-\gamma)\tilde{p}_t(a)+\frac{\gamma}{|S_t|}\mathds{1}\{a\in S_t\}$}
\ENDFOR
\end{algorithmic}
\end{algorithm}
We highlight that the algorithm is computationally efficient (even more efficient than traditional EXP2), because the reward estimator \eqref{eq: IPS GMAB} is strongly concave so that one can apply log-concave sampling when executing exponential weighting.
The additional ``mean zero regularizer'' term in \eqref{eq: IPS GMAB} is ignorable from a regret analysis perspective, so the standard analysis for exponential weight algorithms applies to Algorithm \ref{alg: GLB} to establish the optimal $\tilde{O}(d\sqrt{ T})$ regret bound. One may also analyze Algorithm \ref{alg: GLB} within our algorithmic belief framework through Theorem \ref{thm regret approx}, as we did for Algorithm \ref{alg: BMAB} in Section \ref{appendix APS BMAB}; we omit the analysis here. Finally, we note that the algorithm reduces to a modified version of EXP3 for finite armed bandits, a connection we mentioned at the end of Section \ref{subsec APS BMAB}.

\subsection{Application to bandit convex optimization}
We consider the bandit convex optimization problem described  in Example \ref{example structured bandits}. In bandit convex optimization, $\Pi\subseteq\R^d$ is a $d-$dimensional action set whose diameter is bounded by $\text{diam}(\Pi)$, and the mean reward (or loss) function is required to be concave (respectively, convex) with respect to actions:
\begin{align*}
    \F=\{f: \Pi\rightarrow [0,1]: f \textup{ is concave w.r.t. } \pi\in \Pi\}.
\end{align*}
The problem is often formed with finite (but exponentially large) action set by standard discretization arguments \citep{lattimore2020improved}. Bandit convex optimization is a notoriously challenging problem, and much effort has been put to understanding its minimax regret and algorithm design. The best known result, which is of order $\tilde{O}(d^{2.5}\sqrt{T})$, is derived through the non-constructive information-ratio analysis in \cite{lattimore2020improved}. 
By the information ratio upper bound for the non-constructive Bayesian IDS algorithm in \cite{lattimore2020improved}, Lemma \ref{lemma air ir} that bounds AIR by IR, and Theorem \ref{thm adaptive minimax sampling} (regret of AMS), we immediately have that Algorithm \ref{alg: adaptive minimax sampling} (AMS) with optimally tuned $\eta$ achieves
     \begin{align*}
  \reg_T\leq O\left(d^{2.5}\sqrt{T}\cdot \textup{polylog}(d, \textup{diam}(\A), T)\right)
\end{align*}

As a result, AMS recovers the best known $\tilde{O}(d^{2.5}\sqrt{T})$ regret with a constructive algorithm. By our discussion on the computational complexity in Appendix \ref{subsec concave parameterization}, AMS solves convex optimization in a $\text{poly}(|\Pi|)$-dimensional space, so it can be computed in $\text{poly}(e^d\cdot T)$ time for bandit convex optimization. To the best of our knowledge, this is the first algorithm with a finite running time that attains the best known $\tilde{O}(d^{2.5}\sqrt{T})$ regret. We note that the EBO algorithm in \cite{lattimore2021mirror} has given a constructive algorithm that achieves the same $\tilde{O}(d^{2.5}\sqrt{T})$ regret derived by Bayesian non-constructive analysis. However, EBO operates in an abstract functional space, so it is less clear how to execute the computation.

\section{MAIR and application to RL}\label{sec mair}
In the stochastic environment, where $M_t=M^*\in\M$ for all rounds, we only need to search for algorithmic beliefs regarding the underlying model to determine the best decision $\pi_{M^*}$. This distinction allows us to introduce a strengthened version of AIR in Section \ref{subsec MAIR}, which we term ``Model-index AIR'' (MAIR), particularly suited for studying reinforcement learning problems in the stochastic setting.

Crucially, we can construct a generic and closed-form sequence of algorithmic beliefs that approximate the maximization of MAIR at each round. By leveraging these beliefs, we develop Model-index AMS and Model-index APS algorithms that achieves the sharpest known bounds for RL problems within the bilinear class \citep{du2021bilinear, foster2021statistical}. Some of our algorithms feature a generic and closed-form updating rule, making them potentially well-suited for efficient implementation through efficient sampling oracles.

\subsection{Generic analysis leveraging MAIR}\label{subsec generic MAIR}

In the stochastic setting, by leveraging MAIR, we can prove the following generic regret bound for arbitrary learning algorithm, and with arbitrarily chosen algorithmic belief sequence $\{\mu_t\}_{t=1}^T$.
This generic regret bound is  an analogy of Theorem \ref{thm regret approx}, which is not only applicable to any algorithm but also allows the algorithmic beliefs to be flexibly selected.
 \begin{theorem}[Generic regret bound for arbitrary learning algorithm leveraging MAIR]\label{thm regret approx stochastic}In the stochastic setting, given a finite model class $\M$, the ground truth model $M^*\in\M$, an arbitrary algorithm $\alg$ that produces decision probability $p_1,\dots,p_T$, and a sequence of beliefs $\mu_1, \dots, \mu_{T}$ where $\rho_{t}={(\mu_{t-1})}_{\pi^*|\pi,o}\in \inte(\Delta(\M))$ for all rounds, we have
\begin{align*}
  \reg_T\leq \frac{\log |\M|}{\eta} +\E\lb\sum_{t=1}^T \lp\mair_{\rho_t,\eta}(p_t,\mu_t)+ \lla\left. \frac{\partial \mair_{\rho_t,\eta}(p_t,\mu)}{\partial \mu}\right|_{\mu=\mu_t}, \mathds{1}(M^*)-\mu_t\rra\rp\rb,
\end{align*}
where $\mathds{1}(M^*)$ is the vector whose $M^*-$coordinate is $1$ but all other coordinates are $0$. 
\end{theorem}
An appealing aspect of Theorem \ref{thm regret approx stochastic} is that we only need to bound the gradient-based optimization error with respect to the same $M^*$ for all rounds. This property arises due to the stochastic environment.

Specifically, when the algorithmic beliefs exactly maximizes MAIR at each round (optimization errors equal to $0$), and when applying the minimax strategy in algorithm design, we propose Model-index Adaptive Minimax Sampling (MAMS), see Algorithm \ref{alg: model adaptive minimax sampling} below. 
\begin{algorithm}[htbp]
\caption{Model-index Adaptive Minimax Sampling}
\label{alg: model adaptive minimax sampling}
Input learning rate $\eta>0$. 

Initialize $\rho_1=\text{Unif}(\M)$. 
\begin{algorithmic}[1]
\FOR{round $t=1,2,\cdots, T$}
    \STATE{{ Find a distribution $p$ of $\pi$ and a distribution $\mu_t$ of $M$  that solves the saddle point of }
    \begin{align*}
       \inf_{p\in \Delta(\Pi)}\sup_{\mu\in\Delta(\M)} \texttt{MAIR}_{\rho_t,\eta}(p, \mu).
    \end{align*}}
    \STATE{ Sample decision $\pi_t\sim p_t$ and observe  $o_t\sim M^*(\pi_t)$.}
    \STATE{ Update $\rho_{t+1}={(\mu_t)}_{M|\pi_t,o_t}$.}
\ENDFOR
\end{algorithmic}
\end{algorithm}
Because the minimax value of MAIR is always upper bounded by DEC, as illustrated in Lemma \ref{lemma mdir relationship}, it is straightforward to prove the following theorem.
\begin{theorem}[Regret of MAMS]\label{thm model adaptive minimax sampling} For a finite and compact model class $\M$, the regret of Algorithm \ref{alg: model adaptive minimax sampling} with any $\eta>0$ is always bounded as follows, for all $T\in\N$, 
\begin{align*}
  \reg_T\leq  \frac{\log|\M|}{\eta}+\E\lb\sum_{t=1}^T \mair_{\rho_t,\eta}(p_t,\mu_t)\rb\leq \frac{\log|\M|}{\eta}+\dec_{\eta}^{\KL}(\M)\cdot T.
\end{align*}
\end{theorem}
This shows that the regret bound of MAMS is always no worse than the regret bound of the Estimation-to-Decision (E2D) algorithm in \cite{foster2021statistical}.

\paragraph{Comparing AIR and MAIR.} We have seen from Lemma \ref{lemma air dec}, Lemma \ref{lemma mdir relationship}, and Lemma \ref{lemma mair air} that 1) Maximin AIR can be bounded by DEC of the convex hull $\textup{conv}(\M)$; 2) Maximin MAIR can be bounded by DEC of the original class $\M$; and 3) MAIR is ``smaller'' than AIR. However, as   we have shown in Theorem \ref{thm regret} and Theorem \ref{thm regret approx stochastic}, the regret bound using AIR scales with a $\log |\Pi|$ term (estimation complexity of decision space), while the regret bound using MAIR scales with a bigger $\log |\M|$ term (estimation complexity of model class). We explain their difference as follows.

{\it When to use  \textup{AIR} versus \textup{MAIR}?} First, AIR is applicable to both stochastic and adversarial environments, whereas MAIR may only be applicable to stochastic environments. Second, when using AIR, an estimation complexity term of $\log |\Pi|$ is introduced, whereas MAIR results in a larger estimation complexity term of $\log |\M|$. Therefore, AIR often yields tighter regret bounds for bandit problems. For instance, AIR provides optimal regret bounds for multi-armed bandits and achieves $\sqrt{T}$-type regret bounds for the challenging problem of bandit convex optimization, whereas MAIR may not attain these results. On the other hand, MAIR does achieve optimal regret bounds for stochastic linear bandits.
Notably, MAIR also achieves optimal regret bounds for stochastic contextual bandits with general realizable function classes \citep{foster2020beyond, simchi2022bypassing}, including scenarios with potentially infinite actions  \citep{xu2020upper, foster2020adapting, zhu2022contextual}. Moreover, MAIR is better suited than AIR for reinforcement learning problems in which taking the convex hull to the model class may significantly enhance its expressiveness. For instance, in RL problems, the model class, especially the state transition dynamics, often does not adhere to convexity assumptions.
 In general, AIR is more suitable for ``infinite divisible'' problems where taking the convex hull does not substantially increase the complexity of the model class. Conversely, MAIR is better suited for stochastic model-based bandit and RL problems in which avoiding convex hull operations is preferred.

\subsection{Near-optimal algorithmic beliefs in closed form}

For any fixed decision probability $p$, it is illustrative to write MAIR as
\begin{align}\label{eq: mair mutual information}
    \mair_{\rho,\eta}(p, \mu)&=  \E_{p,\mu}\lb f_M( \pi_M)-f_M(\pi)-\frac{1}{\eta}\KL({\mu}(M|\pi,o), \rho)\rb\nonumber\\
    &=  \E_{p,\mu}\lb f_M( \pi_M)-f_M(\pi)-\frac{1}{\eta}\KL({\mu}(M|\pi,o), \mu)-\frac{1}{\eta}\KL(\mu, \rho)\rb\nonumber\\
    &= \E_{p,\mu}\lb f_M( \pi_M)-f_M(\pi)-\frac{1}{\eta}\KL(M(\pi), {\mu}_{o|\pi})-\frac{1}{\eta}\KL(\mu, \rho)\rb,
\end{align}
where $\mu_{o|\pi}= \E_{M\sim \mu}[M(\pi)]$ is the induced distribution of $o$ conditioned on $\pi$, and the third equality is by property of mutual information. We would like to give a sequence of algorithmic beliefs that approximately maximize MAIR at each rounds, as well as have closed-form expression.

 We consider the following algorithmic priors at each round:
\begin{align}\label{eq: algorithmic belief MAIR}
    \mu_t(M)\propto \rho_t(M)\cdot\exp\lp\underbrace{\eta (f_M(\pi_M)- \E_{\pi\sim p}\lb f_M(\pi)\rb)-\frac{1}{3} \E_{\pi\sim p}\lb D^2_{\textup{H}}(M(\pi), (\rho_t)_{o|\pi})\rb}_{\text{adaptive algorithmic belief}}\rp,
\end{align}
where the adaptive algorithmic belief in \eqref{eq: algorithmic belief MAIR} attempts to optimize the MAIR objective in \eqref{eq: mair mutual information}, with $D^2_{\text{H}}(M(\pi),(\rho_t)_{o|\pi})$  approximates $\KL(M(\pi),(\mu_t)_{o|\pi})$. The factor $1/3$ before the squared Hellinger distance is a technical consequence of the triangle-type inequality of the squared Hellinger distance, which has a factor $2$ rather than $1$ (see Lemma \ref{lemma triangle squared Hellinger}).
And we use their corresponding posteriors to update the sequence of reference probabilities:
\begin{align*}
    \rho_{t+1}(M)=\mu_t(M|\pi_t,o_t)\propto \mu_t(M)M[\pi_t](o_t). 
\end{align*}
This results in the following update of $\rho$:
\begin{align}\label{eq: adaptive posterior sampling model}
    \rho_{t+1}(M)\propto\exp\lp \sum_{s=1}^{t}\lp \underbrace{\log [M(\pi_s)](o_s)}_{\text{log likelihood}} +\underbrace{\eta\lp f_M(\pi_M)-  \E_{\pi\sim p}\lb f_M(\pi)\rb\rp-\frac{1}{3}\E_{\pi\sim p}\lb D^2_{\text{H}}(M(\pi),{(\rho_t)}_{o|\pi})\rb}_{\text{adaptive algorithmic belief}}\rp \rp.
\end{align}
\begin{algorithm}[htbp]
\caption{Closed-form approximate maximizers of MAIR}
\label{alg: model index close form}
Input algorithm $\alg$ and learning rate $\eta>0$.

Initialize $\rho_1$ to be the uniform distribution over $\M$.  
\begin{algorithmic}[1]
\FOR{round $t=1,2,\cdots, T$}
    \STATE{ Obtain $p_t$ from $\alg$. The algorithm $\alg$ samples  $\pi_t\sim p_t$ and observe the feedback $o_t\sim M_t(\pi_{t})$.}
      \STATE{{ Update }
    \begin{align*}
    \mu_t(M)&\propto \rho_t(M)\cdot\exp\lp\eta (f_M(\pi_M)- \E_{\pi\sim p}\lb f_M(\pi)\rb)-\frac{1}{3} \E_{\pi\sim p}\lb D^2_{\text{H}}(M(\pi), {\rho_t}_{o|\pi}\rb\rp,\\
       \rho_{t+1}(M)&=\mu_t(M|\pi_t,o_t).
    \end{align*}}
\ENDFOR
\end{algorithmic}
\end{algorithm}
Using such algorithmic beliefs, the regret of an arbitrary algorithm can be bounded as follows.

\begin{theorem}[Regret for arbitrary algorithm with closed-form beliefs]\label{thm regret model}Given a finite model class $\M$ where the underlying true model is $M^*\in\M$, and $f_M(\pi)\in [0,1]$ for every $M\in\M$ and $\pi\in \Pi$. For an arbitrary algorithm $\alg$ and any $\eta>0$,  the regret of algorithm $\alg$ is bounded as follows, for all $T\in\N$,
\begin{align*}
    \reg_T\leq \frac{\log|\M|}{\eta}+\E\lb\sum_{t=1}^T\E_{\mu_t, p_t}\lb  f_M(\pi_M)-f_M(\pi)- \frac{1}{3\eta} D_{\textup{H}}^2(M(\pi), (\rho_t)_{o|\pi})-\frac{1}{3\eta}\KL(\mu_t,\rho_t)\rb\rb.
\end{align*}
where $\mu_t$ and $\mu_t$ are closed-form beliefs generated according the Algorithm \ref{alg: model index close form}.
\end{theorem}

By combining the belief generation process in Algorithm \ref{alg: model index close form} as an estimation oracle to E2D \citep{foster2021statistical}, we can recover its regret bound using DEC, as we have demonstrated with MAMS in Theorem \ref{thm model adaptive minimax sampling}. Now, we have closed-form expressions for the near-optimal beliefs. Note that our reference probabilities  \eqref{eq: adaptive posterior sampling model} update both the log-likelihood term and an adaptive algorithmic belief term at each iteration. In contrast, most existing algorithms, whether optimistic or not, typically update only the log-likelihood term and rely on a fixed prior term. Furthermore, it's worth highlighting that existing ``optimistic'' or ``model-free'' algorithms often employ a scaled version of the Bayesian posterior formula \citep{agarwal2022model, foster2022note, zhang2022feel, chen2022unified}. In contrast, we approach the problem using the original Bayesian posterior formula by incorporating adaptive beliefs. This approach may also have implications for fundamental problems like density estimation with proper estimators, as earlier methods have typically relied on the scaled version of posterior formulas \citep{zhang2006from, geer2000empirical}, rather than directly utilizing the original posterior formula as we do here.

In our applications, we often use a simple posterior sampling strategy for which we always induce the distribution of optimal decisions from the posterior distribution of models. We refer to the resulting algorithm, Algorithm \ref{alg: model index APS}, as ``Model-index Adaptive Posterior Sampling'' (MAPS).

\begin{algorithm}[htbp]
\caption{ Model-index Adaptive Posterior Sampling}
\label{alg: model index APS}
Input learning rate $\eta$. 

Initialize $\mu_1$ to be the uniform distribution over $\M$.  
\begin{algorithmic}[1]
\FOR{round $t=1,2,\cdots, T$}
    \STATE{ Sample  $\pi_t\sim p_t$ where $p_t(\pi)=\sum_{\pi=\pi_M}\mu_t(M)$, and observe the feedback $o_t\sim M_t(\pi_{t})$.}
    \STATE{Update $\mu_{t}$ and $\rho_{t+1}$ according to Algorithm \ref{alg: model index close form}.}
\ENDFOR
\end{algorithmic}
\end{algorithm}

MAPS draws inspiration from the optimistic posterior sampling algorithm proposed in \cite{agarwal2022model} (also referred to as feel-good Thompson sampling in \cite{zhang2022feel}). However, our approach incorporates adaptive algorithmic beliefs and the original Bayesian posterior formula, rather than using the fixed prior and the scaled posterior update formulas as in \cite{agarwal2022model, zhang2022feel}.

For  model class $\M$, a nominal model $\bar{M}$, and the posterior sampling strategy $p^\textup{TS}(\pi)=\mu(\{M:\pi_M=\pi\})$,  we can define the Bayesian decision-estimation coefficient of Thompson Sampling by 
\begin{align}\label{eq: dec TS}
    \dec_{\eta}^\textup{TS}(\M)=\sup_{\bar{M}\in\text{conv}(\M)}\sup_{\mu\in\Delta(\M)}\E_{\mu, p^\textup{TS}}\Big[ f_M(\pi_M)-f_{M}(\pi)-\frac{1}{\eta}D_{\textup{H}}^2\lp M(\pi), \bar{M}(\pi)\rp\Big].
\end{align} 
This value is bigger than the minimax DEC in Definition \ref{def dec}, but often easier to use in RL problems.

\begin{theorem}[Regret of Model-index Adaptive Posterior Sampling]\label{thm regret model ts}
Given a finite model class $\M$ where $f_M(\pi)\in[0,1]$ for every $M\in\M$ and $\pi\in\Pi$. The regret of Algorithm \ref{alg: model index APS} with $\eta\in(0,1/3]$ is bounded as follows, for all $T\in\N$,
\begin{align*}
   \reg_T\leq \frac{\log|\M|}{\eta}+ \texttt{\textup{DEC}}^{\textup{TS}}_{6\eta}(\M,\bar{M})\cdot T+6\eta T.
\end{align*}

\end{theorem}

\subsection{Application to reinforcement learning}\label{subsec RL}

 By using MAMS (Algorithm \ref{alg: model adaptive minimax sampling}), MAPS (Algorithm \ref{alg: model index APS}), and Algorithm \ref{alg: model index close form} (closed-form algorithmic belief generation), we are able to recover several results in \cite{foster2021statistical} that bound the regret of RL by DEC and the estimation complexity $\log|\M|$ of the model class. Note that MAPS (Algorithm \ref{alg: adaptive posterior sampling}) has the potential to be efficiently implemented through efficient sampling oracles, while the generic E2D algorithm in \cite{foster2021statistical} is not in closed form and requires minimax optimization, and the sharp regret bounds are proved through the non-constructive Bayesian Thompson Sampling. The paper also presents regret bounds for a constructive algorithm using the so-called ``inverse gap weighting'' updating rules \citep{foster2020beyond, simchi2022bypassing}, but that algorithm has worse regret bounds than those proved through the non-constructive approach (by a factor of the bilinear dimension). As a result, Algorithm \ref{alg: model index APS} makes an improvement because its simplicity and achieving the sharpest regret bound proved in \cite{foster2021statistical} for RL problems in the bilinear class.  

We illustrate how the general problem formulation in Section \ref{subsec problem formulation} covers RL problems as follows.

\begin{example}[Reinforcement learning]
An episodic finite-horizon reinforcement learning problems is defined as follows. Let $H$ be the horizon and $\A$ be a finite action set. Each model $M\in\M$ specifies a non-stationary Markov decision process (MDP) $\{\{S^{(h)}\}_{h=1}^H, \A, \{P_{M}^{(h)}\}_{h=1}^H, \{R_{M}^{(h)}\}_{h=1}^H,\mu\}$, where $\mu$ is the initial distribution over  states; and for each layer $h$, $S^{(h)}$ is a finite state space, $P_{M}^{(h)}: S^{(h)}\times \A\rightarrow (S^{(h+1)})$ is the probability transition kernel, and $R^{(h)}_{M}:S^{(h)}\times\A\rightarrow \Delta([0,1])$ is the reward distribution. We allow the transition kernel and loss distribution to be different for different $M\in \M$ but assume $\mu$ to be fixed for simplicity.
Let $\Pi_{\text{NS}}$ be the space of all deterministic non-stationary policies $\pi=(u^{(1)}, \dots, u^{(H)})$, where $u^{(h)}: S^({h})\rightarrow \A$. Given an MDP $M$ and policy $\pi$, the MDP evolves as follows: beginning from $s^{(1)}\sim \mu$, at each layer $h=1,\dots, H$, the action $a^{(h)}$ is sampled from $u^{(h)}(s^{(h)})$, the loss $r^{(h)}(a^{(h)})$ is sampled from $R_{M}(s^{(h)},a^{(h)})$ and the state $s^{(h+1)}$ is sampled from $P_{M}(\cdot|s^{(h)},a^{(h)})$. Define $f_{M}(\pi)=\E[\Sigma_{h=1}^H r^{(h)}(a^{(h)})]$ to be the expected reward under MDP $M$ and policy $\pi$.
The general framework covers episodic reinforcement learning problems by taking the observation $o_t$ to be the trajectory $(s_{t}^{(1)}, a_{t}^{(1)}, r_{t}^{(1)} ), \dots, (s_{t}^{(H)}, a_{t}^{(H)}, r_{t}^{(H)})$ and $\Pi$ be a subspace of $\Pi_{\text{NS}}$. While our framework and complexity measures allow for  agnostic policy classes,  recovering existing results often requires us to make realizability-type assumptions.
\end{example}

We now focus on a broad class of structured reinforcement learning problems called ``bilinear class'' \citep{du2021bilinear}. The following definition of the bilinear class is from \cite{foster2021statistical}.

\begin{definition}[Bilinear class]A model class $\M$ is said to be bilinear relative to reference model $\bar{M}$ if: 

1. There exist functions $W_h(\cdot;\bar{M}): \M\rightarrow \R^d$, $X_h(\cdot;\bar{M}):\M\times \R^d$ such that for all $M\in \M$ and $h\in[H]$, 
\begin{align*}
    |\E^{\bar{M},\pi_M}[Q_h^{M,*}(s_h,a_h)-r_h-V_h^{M,*}(s_{h+1})]|\leq |\langle W_h(M;\bar{M}),X_h(M;\bar{M}) \rangle|.
\end{align*}
We assume that $W_h(M:\bar{M})=0$.

2. Let $z_h=(s_h,a_h,r_h,s_{h+1})$. There exists a collection of estimation policies $\{\pi_M^{\textup{est}}\}_{M\in\M}$ and estimation functions $\{\ell_{M}^{\textup{est}}(\cdot;\cdot)\}_{M\in\M}$ such that for all $M,M'\in\M$ and $h\in[H]$, 
\begin{align*}
    \langle X_h(M;\bar{M}), W_h(M':\bar{M})\rangle =\E^{\bar{M},\pi_M\circ h \pi_M^{\textup{est}}}[\ell_M^{\textup{est}}(M';z_h)].
\end{align*}
If $\pi_M^{\textup{est}}=\pi_M$, we say that estimation is on-policy.

If $M$ is bilinear relative to all $\bar{M}\in\M$, we say that $\M$ is a bilinear class. We let $d_{\textup{bi}}(\M,\bar{M})$ denote the minimal dimension $d$ for which the bilinear class property holds relative to $\bar{M}$, and define $d_{\textup{bi}}(\M)=\sup_{\bar{M}\in\M}d_{\textup{bi}}(\M,\bar{M})$. We let $L_{\textup{bi}}(\M;\bar{M})\geq 1$ denote any almost sure upper bound on $|\ell_M^{\textup{est}}(M';z_h)|$ under $\bar{M}$, and let $L_{\textup{bi}}(\M)=\sup_{\bar{M}\in\M}L_{\textup{bi}}(\M;\bar{M})$.
\end{definition}

For $\gamma\in[0,1]$, let $\pi_M^\gamma$ be the randomized policy that---for each $h$---plays $\pi_{M,h}$ with probability $1-\gamma/H$ and $\pi_{M,h}^{\textup{est}}$ with probability $\gamma/H$. Combining the upper bounds of $\dec_{\eta}^{\text{TS}}$ proved in Theorem 7.1 in \cite{foster2021statistical} with our Theorem \ref{thm model adaptive minimax sampling} (regret of MAMS) and Theorem \ref{thm regret model ts} (regret of MAPS),  we can immediately obtain regret guarantees for RL problems in the bilinear class. 
\begin{theorem}[Regret of MAMS and MAPS for RL problems in the bilinear class]\label{thm rl}
    In the on-policy case, Model-index AMS (Algorithm \ref{alg: adaptive minimax sampling}) and Model-index APS (Algorithm \ref{alg: adaptive posterior sampling}) with optimally tuned $\eta$ achieve regret
\begin{align*}
\reg_T\leq O\big(H^2 L{\text{bi}}^2d_{\text{bi}}(\M)\cdot T\cdot\log|\M|\big).
\end{align*}
In the general case, Model-index AMS (Algorithm \ref{alg: adaptive minimax sampling}) and Model-index APS (Algorithm \ref{alg: adaptive posterior sampling}) with forced exploration rate $\gamma=\big(8\eta H^3L_{\textup{bi}}^2(\M)d_{\textup{bi}}(\M,\bar{M})\big)^{1/2}$ and optimally tuned $\eta$ achieves regret
\begin{align*}
\reg_T\leq O\big(\big(H^3 L_{\text{bi}}^2 d_{\text{bi}}(\M)\log|\M|\big)^{1/3}\cdot T^{2/3}\big).
\end{align*}
\end{theorem}
In particular, as a closed-form algorithm that may be computed through sampling techniques, MAPS matches the sharp results for E2D, MAMS,  and the non-constructive Bayesian Posterior Sampling algorithm used in the proof of Theorem 7.1 in \cite{foster2021statistical}; and MAPS achieves better regret bounds than the closed-form ``inverse gap weighting'' algorithm provided in the same paper. Its regret bound for RL problems in the bilinear class also match the E2D algorithms in \cite{foster2021statistical, foster2022note} that are not in closed-form and require more challenging minimax optimization.

Our results in this section apply to reinforcement learning problems where the DEC is easy to upper bound, but bounding the information ratio may be more challenging, particularly for complex RL problems where the model class $\M$ may not be convex and the average of two MDPs may not belong to the model class. Specifically, we propose MAIR and provide a generic algorithm that uses DEC and the estimation complexity of the model class ($\log|\M|$) to bound the regret. Another promising research direction is to extend our general results for AIR and the tools from Section \ref{subsec linear bandits} to reinforcement learning problems with suitably bounded information ratios, such as tabular MDPs and linear MDPs, as suggested in \cite{hao2022regret}. We hope that our tools can pave the way for developing constructive algorithms that provide regret bounds scaling solely with the estimation complexity of the value function class, which is typically smaller than that of the model class.

\section{Conclusion and future directions}
In this work, we propose a novel approach to solve sequential learning problems by generating ``algorithmic beliefs.'' We optimize the Algorithmic Information Ratio (AIR) to generate these beliefs. Surprisingly, our algorithms achieve regret bounds that are as good as those assuming prior knowledge, even in the absence of such knowledge, which is often the case in adversarial or complex environments. Our approach results in simple and often efficient algorithms for various problems, such as multi-armed bandits, linear and convex bandits, and reinforcement learning. 

Our work provides a new perspective on designing and analyzing bandit and reinforcement learning algorithms. Our theory applies to any algorithm through the notions of AIR and algorithmic beliefs, and it provides a simple and constructive understanding of the duality between frequentist regret and Bayesian regret in sequential learning. Optimizing AIR is a key principle to design effective and efficient bandit and RL algorithms. We demonstrate the effectiveness of our framework empirically via experiments on Bernoulli MAB and show that our derived algorithm achieves ``best-of-all-worlds'' empirical performance. Specifically, our algorithm outperforms UCB and is comparable to TS in stochastic bandits, outperforms EXP3 in adversarial bandits, and outperforms TS as well as clairvoyant restarted algorithms in non-stationary bandits.

Our study suggests several potential research directions, and we hope to see progress made by utilizing the methodology developed in this work.
Firstly, an important task is to provide computational and representational guidelines for optimizing algorithmic beliefs, such as techniques for selecting belief parameterization and index representation (function class approximation).
Secondly, a major goal is to achieve near-optimal regrets with efficient algorithms for challenging problems in infinite-armed bandits, contextual bandits, and reinforcement learning. An initial step involves exploring the Bayesian interpretation of existing frequentist approaches, including gaining a deeper understanding of IPW-type estimators and related computationally-efficient algorithms \citep{abernethy2008competing, agarwal2014taming, bubeck2017kernel}. Moreover, it is worth investigating whether our approach can facilitate the development of more precise regret bounds and principled algorithm design for reinforcement learning problems involving function approximation. 
Thirdly, an important direction is to leverage algorithmic beliefs to study adaptive and dynamic regrets in non-stationary environments and explore instance optimality \citep{wagenmaker2023instance} in stochastic environments. We note that we currently lack theoretical justification for our empirical ``best-of-all-worlds'' performance, and a comparison of our approach to earlier theoretical works on this topic \citep{bubeck2012best, wei2018more}.
 Fourthly, our paper introduces a novel framework for analyzing regret through AIR and offers a rich mathematical structure to explore and uncover, including the geometry of natural parameterization for maximizing AIR and its correspondence to the mirror space (see Appendix \ref{appendix MAB} for details). Fifthly, our aim is to comprehend alternative formulations of AIR, including the constrained  formulation (drawing inspiration from the recent investigation of the constrained formulation of DEC in \cite{foster2023tight}); connection may be made between regularization and the notion of localization \citep{xu2020towards}.
Finally, we hope that the duality identities, which can accommodate arbitrary beliefs, algorithms, and any type of environment (as illustrated in Lemma \ref{lemma identity approx}), may offer values for lower bound analyses and information theory. We encourage exploration in these and all other relevant directions.

\paragraph{Acknowledgement.}We  thank Yunzong Xu for valuable discussions.

\bibliography{references}
\newpage
\appendix
\section{Extensions and proofs for AIR}
\subsection{Extensions of AIR}

\paragraph{Extension to general Bregman divergence}\label{appendix air bregman}
We can generalize AIR from using KL divergence to using general Bregman divergence. And all the results in Section \ref{sec algorithm} can be extended as well. This generalization is inspired by \cite{lattimore2021mirror}, which defines information ratio and studies algorithm design using general Bregman divergence. 

Let $\Psi:\Delta(\Pi)\rightarrow \R\cup\infty$ be a convex Legendre function. Denote $D_{\Psi}$ to be the Bregman divergence of $\Psi$, and $\text{diam}(\Psi)$ to be the diameter of $\Psi$. We refer to Appendix \ref{subsec convex analysis} for the background of these concepts. Given a reference probability $q\in\inte(\Delta(\Pi))$ in the interior of the simplex and learning rate $\eta>0$, we define the generalized Algorithmic Information Ratio with potential function $\Psi$ for decision $p$ and distribution $\nu$ by
 \begin{align}\label{eq: air psi}
     \texttt{AIR}^\Psi_{q,\eta}(p,\nu)= \E\lb f_M( \pi^*)-f_M(\pi)-\frac{1}{\eta}D_{\Psi}({\nu}_{\pi^*|\pi,o}, {\nu}_{\pi^*})-\frac{1}{\eta}D_{\Psi}({\nu}_{\pi^*},q)\rb.
 \end{align}
 We  generalize Theorem \ref{thm regret} and Theorem \ref{thm regret approx} to  general Bregman divergence, with the $\frac{\log|\Pi|}{\eta}$ term in the regret bounds being replaced by a $\frac{\text{diam}(\Psi)}{\eta}$ term. 
Using the extension, we can generalize Theorem \ref{thm adaptive posterior sampling} (regret of APS) and Theorem \ref{thm adaptive minimax sampling} (regret of AMS) to generalized Bregman divergence as well, where the definition of information ratio will also use the corresponding Bregman divergence as in \cite{lattimore2021mirror}. We state the extension of Theorem \ref{thm regret approx} here.
\begin{theorem}[Using general Bregman divergence]\label{thm regret approx Bregman}Assume $\Psi: \Delta(\Pi)\rightarrow \R\cup\infty$ is Legendre and has bounded diameter. Given a compact $\M$, an arbitrary algorithm $\alg$ that produces decision probability $p_1,\dots,p_T$, and a sequence of beliefs $\nu_1, \dots, \nu_{T}$ where ${(\nu_t)}_{\pi^*|\pi,o}\in \inte(\Delta(\Pi))$ for all rounds, we have
\begin{align*}
   \reg_T\leq \frac{\textup{diam}(\Psi)}{\eta}+ \E\lb \sum_{t=1}^T \lp\texttt{\textup{AIR}}^{\Psi}_{q_t,\eta}(p_t,\nu_t)+\sup_{\nu^*}\lla \left. \frac{\partial \air^\Psi_{q_t,\eta}(p_t,\nu) }{\partial \nu}\right\vert_{\nu=\nu_t} , \nu^*-{\nu_t} \rra\rp\rb.
\end{align*}
\end{theorem}
    
Note that an analogous result to Theorem \ref{thm model adaptive minimax sampling} can also be derived using the general Bregman divergence version of MAIR in the stochastic setting.

\paragraph{Extension to high probability bound}

We conjecture that the results in Section \ref{sec algorithm} may be able to be extended to high probability bounds, with some modification in our algorithms and complexity measures. We propose this direction as an open question and refer to \cite{foster2022complexity} for a possible approach to achieve this goal.

\subsection{Proof of Theorem \ref{thm regret}} \label{appendix thm regret}

{\bf Theorem 3.1} (Generic regret bound for arbitrary learning algorithm). {\it Given a finite decision space $\Pi$, a compact model class $\M$, the regret of an arbitrary learning algorithm $\alg$ is bounded as follows, for all $T\in\N$,
\begin{align*}
   \reg_T\leq \frac{\log |\Pi|}{\eta}+ \E\lb\sum_{t=1}^T \texttt{\textup{AIR}}_{q_t,\eta}(p_t,\nu_t)\rb.
\end{align*}
}

By the discussion in Section \ref{subsec regret AIR} and \ref{subsec minimax theory}, we only need to prove Lemma \ref{lemma identity Nash} in order to prove Theorem \ref{thm regret}.
\vspace{0.1in}

\noindent{\bf Lemma 5.2} (Identity by Nash equilibrium). {\it
Given $q\in \inte(\Delta(\Pi))$, $\eta>0$ and $p\in \Delta(\Pi)$, denote $\bar{\nu}\in \Delta(\M\times\Pi)$. Then we have 
\begin{align*}
    \sup_{M, \bar{\pi}}\E_{\pi\sim p,o\sim M(\pi)}\left[f_M(\bar{\pi})-f_M(\pi)-\frac{1}{\eta}\log\frac{{\bar{\nu}}_{\pi^*|\pi,o}(\bar{\pi})}{q(\bar{\pi})}\right]=\texttt{\textup{AIR}}_{q, \eta}(p, \bar{\nu}).
\end{align*}
}
\paragraph{Proof of Lemma \ref{lemma identity Nash}:}
Let $\mathcal{Q}$ be the space of all mappings from $\Pi\times \OC$ to $\Delta(\Pi)$. For every mapping $Q\in \mathcal{Q}$, denote $Q[\pi,o](\cdot)\in \Delta(\Pi)$ as the image of $(\pi, o)$, which corresponds to a distribution of $\pi^*$. Given the fixed decision probability $\pi\sim p$, define $B: \Delta(\M\times \Pi)\times \mathcal{Q}\rightarrow \R$ by
\begin{align}\label{eq: Q formulation}
    B(\nu, Q)= \E_{(M,\pi^*)\sim \nu, \pi\sim p}\left[f_{M}(\pi^*)-f_{M}(\pi)-\frac{1}{\eta}\log\frac{Q[\pi,o](\pi^*)}{q(\pi^*)}\right].
\end{align}
This is a strongly convex function with respect to $Q$. For every belief $\nu\in \Delta(M\times\Pi)$, denote  $Q_{\nu}=\arg\min_{q\in\mathcal{Q}} B(\nu,Q)$ as the unique minimizer of $B(\nu,Q)$ given $\nu$. By the first-order optimality condition, we have that $Q_{\nu}$ corresponds to the  posterior functional that always maps the observation pair $(\pi,o)$ to the marginal Bayesian posterior ${\nu}_{\pi^*|\pi,o}$ given prior $\nu$.
Plugging in such minimizer $Q_{\nu}$ into $B(\nu,Q)$, we have
\begin{align*}
   &\inf_{Q\in \mathcal{Q}}B(\nu, Q) \nonumber\\
   =& B(\nu, Q_{\nu})\nonumber\\
    = & \E_{(M,\pi^*)\sim \nu, \pi\sim p}\left[f_M(\pi^*)-f_M(\pi)-\frac{1}{\eta}\KL(\nu_{\pi^*|\pi,o},q)\right]\\
    =&\air_{q,\eta}(p,\nu).
\end{align*}

Applying our proposed constructive minimax theorem for minimax solutions (Lemma \ref{lemma minimax decision}), by choosing the worst-case prior belief 
\begin{align*}
    \bar{\nu}\in\arg\max\air_{q,\eta}(p,\nu), 
\end{align*} we can establish that $(\bar{\nu}, Q_{\bar{\nu}})$ will be a Nash equilibrium and $Q_{\bar{\nu}}$ is a construction of the minimax solution of the minimax optimization problem $\min_{Q}\max_{\nu}B(\nu,Q)$. As a result, we have
\begin{align*}
    &\texttt{\textup{AIR}}_{q, \eta}(p, \bar{\nu})\\
    = & B(\bar{\nu}, Q_{\bar{\nu}})\\
    =& \sup_{\nu}B(\nu,Q_{\bar{\nu}})
    \\=&\sup_{M, \bar{\pi}}\E_{(M,\pi^*)\sim \bar{\nu}, \pi\sim p}\left[f_M(\bar{\pi})-f_M(\pi)-\frac{1}{\eta}\log\frac{\bar{\nu}_{\pi^*|\pi,o}(\bar{\pi})}{q(\bar{\pi})}\right].
\end{align*}
Finally, note that in order to apply our proposed minimax theorem for minimax solutions (Lemma \ref{lemma minimax decision}) to the concave-convex objective function $B$, we need to verify that the sets $\mathcal{Q}$ and $\Delta(\M \times \Pi)$ are convex and compact sets, and $B$ is continuous with respect to both $Q \in \mathcal{Q}$ and $\nu\in \Delta(\M \times \Pi)$. This verification step assumes a basic understanding of general topology, as it involves infinite sets (compactness and continuity for finite sets are trivial). We demonstrate the verification step below, which is optional for readers with a general background.
\paragraph{Verification of the conditions of our constructive minimax theorem (Lemma \ref{lemma minimax decision}):} 
It is straightforward to see convexity of the sets $\mathcal{Q}$ and $\Delta(\M\times \Pi)$.  As a collection of mappings, $\mathcal{Q}$ is compact with respect to the product topology by Tychonoff's theorem, and  $B$ is continuous  with respect to $Q$ by the definition of product topology.  Because the probability measure on the compact set is compact with respect to the weak*-topology, $\Delta(\M\times \Pi)$ is a compact set. Finally, $B$ is continuous with respect to $\nu$ because $B$ is linear in $\nu$. For foundational knowledge in general topology, we direct readers to \cite{bogachev2007measure}. A similar verification process concerning compactness and continuity can be found in \cite{lattimore2021mirror}, where we borrow the technique.

\hfill$\square$

\subsection{Proof of Lemma \ref{lemma minimax decision}} 
{\bf Lemma 5.1} (A constructive minimax theorem for minimax solutions). {\it
Let $\X$ and $\Y$ be convex and compact sets, and $\psi: \X\times \Y\rightarrow \R$ a function which for all $y$ is strongly convex and continuous in $x$ and for all $x$ is concave and continuous in $y$. For each $y\in\Y$, let $x_y=\min_{x\in\X} \psi(x,y)$ be the corresponding unique minimizer.
Then, by maximizing the concave objective 
\begin{align*}\bar{y} \in \arg\max_{y\in\Y}\psi(x_y,y),\end{align*} we can establish that $x_{\bar{y}}$ is a minimax solution of $\min_{x\in\X}\max_{y\in\Y}\psi(x,y)$ and $(x_{\bar{y}},\bar{y})$ is a Nash equilibrium.
}

\paragraph{Proof of Lemma \ref{lemma minimax decision}:}
Since $\X$ and $\Y$ are convex sets, and $\psi:\X\times\Y\rightarrow \R$ is a function such that, for all $y\in\Y$, it is convex and continuous in $x$, and for all $x\in\X$, it is concave and continuous in $y$, all the conditions of Sion's minimax theorem (Lemma \ref{lemma sion}) are satisfied. By applying Sion's minimax theorem, we have the following equality:

\begin{align}\label{eq: minimax equality}
\min_{x\in\X}\max_{y\in\Y} \psi(x,y) = \max_{y\in\Y}\min_{x\in\X}\psi(x,y).
\end{align}

By the definition of $\bar{y}$, we know that $\bar{y}\in\arg\max_{y\in\Y}\psi(x,y)$ is a maximin solution of the convex-concave game $\psi$. Similarly, denote $\bar{x}\in\arg\min_{x\in\X}\psi(x,y)$ to be a minimax solution of the convex-concave game $\psi$. We claim that $(\bar{x},\bar{y})$ is a Nash equilibrium of the game, meaning $\bar{x}\in \arg\min_{x\in\X}\psi(x,\bar{y})$ and $\bar{y}\in\arg\max_{y\in\Y}\psi(\bar{x},y)$. This claim can be proved as follows:

Since $\bar{x}$ is a minimax solution, we have:
\begin{align}\label{eq: minimax consequence}
\min_{x\in\X} \max_{y\in\Y} \psi(x,y) = \max_{y\in\Y} \psi(\bar{x},y) \geq \psi(\bar{x},\bar{y}),
\end{align}
where the last inequality will be an equality if and only if $\bar{y}\in \max_{y\in\Y} \psi(\bar{x},y)$. Similarly, because $\bar{y}$ is a maximin solution, we have:
\begin{align}\label{eq: maximin consequence}
\max_{y\in\Y}\min_{x\in\X} \psi(x,y) = \min_{x\in\X}\psi(x,\bar{y}) \leq \psi(\bar{x},\bar{y}),
\end{align}
where the last inequality will be an equality if and only if $\bar{x}\in \min_{x\in\X} \psi(x,\bar{y})$.

Combining the equality \eqref{eq: minimax equality} from Sion's minimax theorem with \eqref{eq: minimax consequence} and \eqref{eq: maximin consequence}, we find that all five values in \eqref{eq: minimax consequence} and \eqref{eq: maximin consequence} are equal. This implies that $(\bar{x},\bar{y})$ is a Nash equilibrium of the game, satisfying $\bar{x}\in \arg\min_{x\in\X}\psi(x,\bar{y})$ and $\bar{y}\in\arg\max_{y\in\Y}\psi(\bar{x},y)$.

The traditional minimax theorem only provides the identity between values. To give a concrete construction of the minimax solution, we can impose strong convexity of $\psi$ with respect to $x$. Note that we have defined $\bar{y}$ as a worst-case choice $\arg\max_{y\in\Y}\psi(x_{y},y)$, and $x_{\bar{y}}$ as the unique minimizer of $\psi(x,\bar{y})$. From the fact that $(\bar{x},\bar{y})$ is a Nash equilibrium of the game, and $\bar{x}\in \arg\min_{x\in\X}\psi(x,\bar{y})$ (proved in the last paragraphs), along with the uniqueness of $x_{\bar{y}}$ due to strong convexity, we can conclude that $\bar{x}=x_{\bar{y}}$. Therefore, $x_{\bar{y}}$ is a minimax solution of $\min_{x\in\X}\max_{y\in\Y}\psi(x,y)$ and $(x_{\bar{y}},\bar{y})$ is a Nash equilibrium.

\hfill$\square$

\subsection{Proof of Theorem \ref{thm adaptive posterior sampling}}
{\bf Theorem 3.2} (Regret of APS). {\it Assume that  $f_M(\pi)\in[0,1]$ for all $M\in\M$ and $\pi\in\Pi$. The regret of Algorithm \ref{alg: adaptive posterior sampling} with $\eta= \sqrt{2\log|\Pi|/(( \ir_{\textup{H}}(\textup{TS})+4)\cdot T)}$ and $T\geq 5\log |\Pi|$ is bounded by
\begin{align*}
   \reg_T\leq  \sqrt{2\log |\Pi| \lp \texttt{\textup{IR}}_{\textup{H}}(\textup{TS})+4\rp T},
\end{align*}
where $\ir_{\textup{H}}(\textup{TS}):=\sup_{\nu}\ir_{\textup{H}}(\nu,\nu_{\pi^*})$ is the maximal value of information ratio for Thompson Sampling. Moreover, the regret of Algorithm \ref{alg: adaptive posterior sampling} with any $\eta\in(0,1/3]$ is bounded as follows,  for all $T\in\N$,
\begin{align*}
    \reg_T\leq \frac{\log|\Pi|}{\eta}+T\cdot \lp \dec_{2\eta}^{\textup{TS}}(\textup{conv}(\M))+2\eta\rp,
\end{align*}  
where $\dec_{2\eta}^{\textup{TS}}(\textup{conv}(\M)):=\sup_{\bar{M}\in\textup{conv}(\M)}\sup_{\mu\in\Delta(\textup{conv}(\M))}\E_{\mu, p^\textup{TS}}\lb f_M(\pi_M)-f_M(\pi)-\frac{1}{2\eta}D_{\textup{H}}^2(M(\pi),\bar{M}(\pi))\rb$ is DEC of $\textup{conv}(\M)$ for the Thompson Sampling strategy $p^{\textup{TS}}(\pi)=\mu(\{M:\pi_M=\pi\})$.
}

As stated in the footnote of Theorem \ref{thm adaptive posterior sampling}, we define the squared-Hellinger-distance version of information ratio by
\begin{align*}
    \ir_{\text{H}}(\nu, p)=\frac{\E_{\nu,p}\lb(f_M(\pi^*)-f_M(\pi))^2\rb}{\E_{\nu,p}\lb D_{\textup{H}}^2\lp\nu_{\pi^*|\pi,o},\nu_{\pi^*}\rp\rb}.
\end{align*}
  We prove the following lemma that upper bounds $\air_{q_t,\eta}(q_t,\nu_t)$ by  $\ir_{\text{H}}$ and DEC of Thompson Sampling. The main goal of Lemma \ref{lemma air ref alg} is to replace the decision probability $\pi\sim q$ in $\air_{q,\eta}(q,\nu)$ (which is the strategy of APS) with the decision probability $\pi\sim \nu_{\pi^*}$ (which is the actual strategy of Bayesian Thompson Sampling).

\begin{lemma}[Bounding AIR by DEC and IR for TS]\label{lemma air ref alg}Assume that $f_M(\pi)$ is bounded in $[0,1]$ for all $M,\pi$. Then for $\eta\in(0, 1/3]$ and all $q\in\inte(\Delta(\Pi))$, we have
\begin{align*}
    \air_{q,\eta}(q,\nu)\leq \dec_{2\eta}^{\textup{TS}}(\textup{conv}(\M))+2\eta\leq\frac{\eta}{2}\cdot \texttt{\textup{IR}}_{\textup{H}}({\textup{TS}})+2\eta.
\end{align*}
\end{lemma}

Theorem \ref{thm adaptive posterior sampling} will be a straightforward consequence of  Theorem \ref{thm regret} and Lemma \ref{lemma air ref alg}. By the regret bound \eqref{eq: thm regret} in Theorem \ref{thm regret}, for the APS algorithm where $p_t=q_t$ for all rounds, we have
\begin{align*}
    \reg_T\leq &\frac{\log|\Pi|}{\eta}+\E\lb\sum_{t=1}^T \air_{q_t,\eta}(q_t,\nu_t)\rb\\
    \leq &\frac{\log|\Pi|}{\eta}+\frac{\eta T}{2}\lp \ir_{\text{H}}(\text{TS})+4\rp
\end{align*}
In particular, for $\eta= \sqrt{2\log|\Pi|/(( \ir_{\textup{H}}(\textup{TS})+4)\cdot T)}$ and $T\geq 5\log|\Pi|$ (the condition on $T$ is to ensure the condition $\eta\leq 1/3$ in Lemma \ref{lemma air ref alg}), we have
\begin{align*}
   \reg_T\leq  \sqrt{2\log |\Pi| \lp \texttt{\textup{IR}}_{\textup{H}}(\textup{TS})+4\rp T}.
\end{align*}

\hfill$\square$

\noindent{\bf Proof of Lemma \ref{lemma air ref alg}: } 
 Given a probability measure $\nu$, Denote 
 $\nu_{o|\pi^*,\pi}=\E_{M\sim \nu_{M|\pi^*}}\lb M(\pi) \rb$ to be the posterior belief of observation $o$ conditioned on $\pi^*$ and $\pi$, and $\nu_{o|\pi}=\E_{(M,\pi^*)\sim \nu}\lb M(\pi)\rb$ to be posterior belief of $o$ conditioned solely on $\pi$. 

Denote  the $|\Pi|-$dimensional vector $X,Y$ by
\begin{align*}
      X(\pi) &= \E_{(M,\pi^*)\sim \nu}\lb f_M(\pi^*)-f_M(\pi)\rb,\\
     Y(\pi) &= \E_{(M,\pi^*)\sim \nu}\lb D_{\textup{H}}^2({\nu}_{o|\pi^*,\pi}, {\nu}_{o|\pi})\rb .
\end{align*} 

Note that the Algorithmic Information Ratio  can always be written as
\begin{align}\label{eq: air mutual information}
    \air_{q,\eta}(q,\nu)=& \E_{\nu, q}\left[f_M(\pi^*)-f_M(\pi)-\frac{1}{\eta}\KL(\nu_{\pi^*|\pi,o}, q)\right]\nonumber\\
    =&\E_{\nu, q}\left[f_M(\pi^*)-f_M(\pi)-\frac{1}{\eta}\KL(\nu_{\pi^*|\pi,o}, \nu_\pi^*)-\frac{1}{\eta}\KL(\nu_{\pi^*}, q)\right]\nonumber\\
    = &\E_{\nu, q}\left[f_M(\pi^*)-f_M(\pi)-\frac{1}{\eta}\KL(\nu_{o|\pi^*,\pi}, {\nu}_{o|\pi})-\frac{1}{\eta}\KL(\nu_{\pi^*}, q)\right],
\end{align}
where the first equality is the definition of AIR; the second equality is because the expectation of posterior is equal to prior; and the third equality is due to the symmetry property of mutual information.

By \eqref{eq: air mutual information} we have that
\begin{align}\label{eq: air dec ts}
&\air_{q,\eta}(q,\nu)\nonumber\\
\leq &\E_{ \nu, q}\lb f_M(\pi_M)-f_M(\pi)-\frac{1}{\eta}D_{\textup{H}}^2({\nu}_{o|\pi^*,\pi}, {\nu}_{o|\pi})\rb-\frac{1}{\eta}\KL({\nu}_\pi^*, \rho_t)\nonumber\\
    = &\lla q, X\rra-\frac{1}{2\eta}\KL({\nu}_{\pi^*}, q)-\frac{1}{\eta}\lla q, Y\rra-\frac{1}{2\eta}\KL({\nu}_{\pi^*}, q)\nonumber\\
    \leq &\lla {\nu}_{\pi^*}, X\rra +2\eta-\frac{1}{\eta}\lla q, Y\rra-\frac{1}{2\eta}\KL({\nu}_{\pi^*}, q)\nonumber\\
    \leq &\lla {\nu}_{\pi^*}, X\rra +2\eta-\frac{1}{\eta}\lla q, Y\rra-\frac{1}{2\eta}D_{\textup{H}}^2({\nu}_{\pi^*}, q)\nonumber\\
    \leq &\lla {\nu}_{\pi^*}, X\rra -\frac{(1-\eta)}{(1+\eta)\eta}\lla {\nu}_{\pi^*}, Y\rra+2\eta\nonumber\\
    \leq &\E_{\nu,  {\nu}_{\pi^*}}\lb f_M(\pi_M)-f_M(\pi)-\frac{1}{2\eta}D_{\textup{H}}^2(\nu_{o|\pi^*,\pi}, \nu_{o|\pi})\rb+2\eta,
\end{align}
where the first inequality is by Lemma \ref{lemma Hellinger KL}; the second inequality is by Lemma \ref{lemma drifted error bound} and the fact $f_M(\pi)\in[0,1]$ for all $M\in\M$ and $\pi\in\Pi$; the third inequality is by Lemma \ref{lemma Hellinger KL};  the fourth inequality is a consequence of Lemma \ref{lemma Pinsker Hellinger}, the AM-GM inequality, and $\eta\leq 1$; and the last  inequality uses the condition $\eta\leq \frac{1}{3}$. 

Therefore, we have 
\begin{align*}
        &\air_{q,\eta}(q,\nu)
        \\\leq &\E_{\nu,  {\nu}_{\pi^*}}\lb f_M(\pi_M)-f_M(\pi)-\frac{1}{2\eta}D_{\textup{H}}^2(\nu_{o|\pi^*,\pi}, \nu_{o|\pi})\rb+2\eta\\
        \leq &\dec_{2\eta}^{\textup{TS}}(\textup{conv}(\M))+2\eta
        \\\leq &\frac{\eta}{2}\cdot \texttt{\textup{IR}}_{\text{H}}(\textup{TS})+2\eta,
\end{align*}where the first inequality is by \eqref{eq: air dec ts}; the second inequality follows the same proof as in  Lemma \ref{lemma air dec} (see below Lemma \ref{lemma air dec}); and the last inequality is by the definition of $\texttt{\textup{IR}}_{\text{H}}(\textup{TS})$ defined in Theorem \ref{thm adaptive posterior sampling} and the AM-GM inequality.

\hfill$\square$

\subsection{Proof of Theorem \ref{thm adaptive minimax sampling}}

{\bf Theorem 3.3} (Regret of AMS). {\it For a finite decision space $\Pi$ and a compact model class $\M$, the regret of Algorithm \ref{alg: adaptive minimax sampling} with any $\eta>0$ is always bounded as follows, for all $T\in\N$, 
\begin{align*}
  \reg_T\leq  \frac{\log |\Pi|}{\eta}+\dec_{\eta}^{\KL}(\textup{conv}(\M))\cdot T.
\end{align*} In particular, the regret of Algorithm \ref{alg: adaptive minimax sampling} with  $\eta= 2\sqrt{\log|\Pi|/(\texttt{\textup{IR}}(\textup{IDS})\cdot T)}$ and all $T\in\N$ is bounded by 
\begin{align*}
  \reg_T\leq  \sqrt{\log |\Pi|\cdot \texttt{\textup{IR}}(\textup{IDS})\cdot T},
\end{align*}
where $\texttt{\textup{IR}}(\textup{IDS}):=\sup_{\nu}\inf_p \ir(\nu,p)$ is the  maximal information ratio of Information-Directed Sampling (IDS). 
}

\paragraph{Proof of Theorem \ref{thm adaptive minimax sampling}:}
The proof of Theorem \ref{thm adaptive minimax sampling} is straightforward.  For the DEC upper bound, We have
\begin{align*}
    \reg_T\leq &\frac{\log|\Pi|}{\eta}+\E\lb\sum_{t=1}^T \air_{q_t,\eta}(p_t,\nu_t)\rb\\
    \leq &\frac{\log|\Pi|}{\eta}+\dec_{\eta}^{\KL}(\textup{conv}(\M))\cdot T,
\end{align*}
where the first inequality is by Theorem \ref{thm regret}, and the second inequality is by Lemma \ref{lemma air dec}.
Moreover, for the IR upper bound, we have
\begin{align*}
    \reg_T\leq &\frac{\log|\Pi|}{\eta}+\E\lb\sum_{t=1}^T \air_{q_t,\eta}(p_t,\nu_t)\rb\\
    \leq &\frac{\log|\Pi|}{\eta}+\frac{\eta}{4}\ir(\textup{IDS})\cdot T,
\end{align*}
where the first inequality is by Theorem \ref{thm regret}, and the second inequality is by Lemma \ref{lemma air ir} and the definition of $\ir(\textup{IDS})$ in Theorem \ref{thm adaptive minimax sampling}. In particular, taking $\eta=2\sqrt{\log|\Pi|/(\ir(\textup{IDS})\cdot T)}$ we have
\begin{align*}
    \reg_T\leq \sqrt{\log|\Pi|\cdot\ir(\textup{IDS})\cdot T}.
\end{align*}

\hfill$\square$

\subsection{Proof of Theorem \ref{thm regret approx}}\label{appendix arbitrary}
{\bf Theorem 3.4} (Generic regret bound using approximate maximizers). {\it Given a finite  $\Pi$, a compact $\M$, an arbitrary algorithm $\alg$ that produces decision probability $p_1,\dots,p_T$, and a sequence of beliefs $\nu_1, \dots, \nu_{T}$ where $q_{t}={(\nu_{t-1})}_{\pi^*|\pi,o}\in \inte(\Delta(\Pi))$ for all rounds, we have
\begin{align*}
   \reg_T\leq \frac{\log |\Pi|}{\eta}+ \E\lb \sum_{t=1}^T 
   \lp\texttt{\textup{AIR}}_{q_t,\eta}(p_t,\nu_t)+\sup_{\nu^*}\lla  \left.\frac{\partial \air_{q_t,\eta}(p_t,\nu) }{\partial \nu} \right\vert_{\nu=\nu_t}, \nu^*-{\nu_t} \rra\rp\rb.
\end{align*}
}

The proof of Theorems \ref{thm regret approx} has two steps. The first step is regret decomposition, which is almost the same with the proof of Theorem \ref{thm regret},  and we have illustrate this step in Section \ref{subsec regret AIR}. The second step is to use the identity for arbitrary algorithmic beliefs (Lemma \ref{lemma identity approx}). 

\paragraph{Proof of Theorem \ref{thm regret approx}:}
By \eqref{eq: telescope}, \eqref{eq: expectation telescope}, and the first equality and the inequality in \eqref{eq: identity Nash} (without the last equality), we have that for every $\bar{\pi}\in \Pi$, 
\begin{align*}
\E\left[\sum_{t=1}^T f_{M_t}(\bar{\pi})-\sum_{t=1}^T f_{M_t}(\pi_t)\right]- \frac{\log|\Pi|}{\eta} \leq  \E\lb\sum_{t=1}^T \sup_{M, \bar{\pi}}\E_{t-1}\left[f_{M}(\bar{\pi})-f_{M}(\pi_t)-\frac{1}{\eta}\log\frac{{(\nu_t)}_{\pi^*|\pi_t,o_t}(\bar{\pi})}{q_t(\pi^*)}\right]\rb, 
\end{align*}
which implies
\begin{align*}
    \reg_T\leq \frac{\log|\Pi|}{\eta}+\E\lb\sum_{t=1}^T \sup_{M, \bar{\pi}}\E_{t-1}\left[f_{M}(\bar{\pi})-f_{M}(\pi_t)-\frac{1}{\eta}\log\frac{{(\nu_t)}_{\pi^*|\pi_t,o_t}(\bar{\pi})}{q_t(\pi^*)}\right]\rb.
\end{align*}
By Lemma \ref{lemma identity approx}, we have 
\begin{align*}
    \sup_{M, \bar{\pi}}\E_{t-1}\left[f_{M}(\bar{\pi})-f_{M}(\pi_t)-\frac{1}{\eta}\log\frac{{(\nu_t)}_{\pi^*|\pi_t,o_t}(\bar{\pi})}{q_t(\pi^*)}\right]= \texttt{\textup{AIR}}_{q_t, \eta}(p_t, \nu_t)+\lla\left.\frac{\partial \air_{q_t,\eta}(p_t,\nu)}{\partial \nu}\right\vert_{\nu=\nu_t}, \mathds{1}(M, \bar{\pi})-\nu_t\rra.
\end{align*}
Combining the above two inequalities proves Theorem \ref{thm regret approx}.

\hfill$\square$

\noindent{\bf Lemma 5.3} (Identity for arbitrary algorithmic belief and any environment). {\it
Given $q\in \inte(\Delta(\Pi))$, $\eta>0$  $p\in \Delta(\Pi)$, and an arbitrary algorithmic belief $\hat{\nu}\in \Delta(\M\times\Pi)$ selected by the agent and an arbitrary environment $(M, \bar{\pi})$ specified by the nature. Then we have 
\begin{align*}
    \E_{\pi\sim p,o\sim M(\pi) }\left[f_M(\bar{\pi})-f_M(\pi)-\frac{1}{\eta}\log\frac{{\hat{\nu}}_{\pi^*|\pi,o}(\bar{\pi})}{q(\bar{\pi})}\right]=\texttt{\textup{AIR}}_{q, \eta}(p, \hat{\nu})+\lla\left.\frac{\partial \air_{q,\eta}(p,{\nu})}{\partial {\nu}}\right\vert_{\nu=\hat{\nu}}, \mathds{1}(M, \bar{\pi})-\hat{\nu}\rra,
\end{align*}
where $\mathds{1}(M,\bar{\pi})$ is a vector in simplex $\Delta(\M\times \Pi)$ where all the weights are given to $(M,\bar{\pi})$.
}

\paragraph{Proof of Lemma \ref{lemma identity approx}:} Given a fixed $p$, we define 
\begin{align*}
    B(\nu, Q)= \E_{(M,\pi^*)\sim \nu,\pi\sim  p}\lb f_M(\pi^*)-f_M(\pi)-\frac{1}{\eta}\log\frac{Q[\pi,o](\pi^*)}{q(\pi^*)}\rb
\end{align*}
as in Appendix \ref{appendix thm regret}. 
We have that $B(\nu,Q)$ is a strongly convex function in $Q$ and is a linear function in $\nu$. For every $\nu$, the optimal choice of the mapping $Q:\Pi\times\OC\rightarrow\Delta(\Pi)$ will be the marginal posterior mapping $Q=\nu_{\pi^*|\cdot, \cdot}$, which is specified by $$Q[\pi,o]= \nu_{\pi^*|\pi,o},\quad \forall \pi\in\Pi,o\in\OC.$$ And the value of the $B$ function at this $(\nu,Q)=(\nu, \nu_{\pi^*|\cdot,\cdot})$ will be
\begin{align*}
    B(\nu, \nu_{\pi^*|\cdot,\cdot})=\air_{q,\eta}(p,\nu).
\end{align*}
Since $B$ is a linear function with respect to $\nu$, we have that for any other $\nu^*$,
\begin{align}\label{eq: lemma approx 1}
    B(\nu^*, \nu_{\pi^*|\cdot,\cdot})=&B(\nu, \nu_{\pi^*|\cdot,\cdot})+\lla \left.\frac{\partial B(\nu, Q)}{\partial\nu}\right|_{Q=\nu_{\pi^*|\cdot,\cdot}}, \nu^*-\nu\rra\nonumber\\
    = &\air_{q,\eta}(p,\nu)+ \lla \left.\frac{\partial B(\nu, Q)}{\partial\nu}\right|_{Q=\nu_{\pi^*|\cdot,\cdot}}, \nu^*-\nu\rra.
\end{align}
Now we use Danskin's theorem (Lemma \ref{lemma duality gradients}), which establishes equivalence between the gradient of the optimized objective and the partial derivative at the optimizer. We refer to Appendix \ref{appendix danskin} for the background of Danskin's theorem. By Danskin's theorem (Lemma \ref{lemma duality gradients}), we have that
\begin{align}\label{eq: lemma approx 2}
  \frac{\partial \air_{q,\eta}(p,\nu)}{\partial \nu} = \left.\frac{\partial B(\nu, Q)}{\partial\nu}\right|_{Q=\nu_{\pi^*|\cdot,\cdot}}.
\end{align}
Taking \eqref{eq: lemma approx 2} to \eqref{eq: lemma approx 1} proves Lemma \ref{lemma identity approx}. 

\hfill$\square$
\subsection{Proof of  Theorem \ref{thm regret approx Bregman}}\label{appendix thm approx bregman}
In this section we prove Theorem \ref{thm regret approx Bregman}, which is a more general extension to Theorem \ref{thm regret approx}: Theorem \ref{thm regret approx Bregman} applies to general Bregman divergence while  Theorem \ref{thm regret approx} is stated with the KL divergence. 
\vspace{0.1in}

\noindent{\bf Theorem A.1} (Using general Bregman divergence). {\it Assume $\Psi: \Delta(\Pi)\rightarrow \R\cup\infty$ is Legendre and has bounded diameter. Given a compact $\M$, an arbitrary algorithm $\alg$ that produces decision probability $p_1,\dots,p_T$, and a sequence of beliefs $\nu_1, \dots, \nu_{T}$ where ${(\nu_t)}_{\pi^*|\pi,o}\in \inte(\Delta(\Pi))$ for all rounds, we have
\begin{align*}
   \reg_T\leq \frac{\textup{diam}(\Psi)}{\eta}+ \E\lb \sum_{t=1}^T \lp\texttt{\textup{AIR}}^{\Psi}_{q_t,\eta}(p_t,\nu_t)+\sup_{\nu^*}\lla \left. \frac{\partial \air^\Psi_{q_t,\eta}(p_t,\nu) }{\partial \nu} \right\vert_{\nu=\nu_t}, \nu^*-{\nu_t} \rra\rp\rb.
\end{align*}
}

\paragraph{Proof of Theorem \ref{thm regret approx Bregman}:} Recall the definition of generalized AIR in Appendix \ref{appendix air bregman}, where $\Psi$ is a convex Legendre function:
\begin{align*}
     \texttt{AIR}^\Psi_{q,\eta}(p,\nu)= \E_{\nu,p}\lb f_M( \pi^*)-f_M(\pi)-\frac{1}{\eta}D_{\Psi}(\nu_{\pi^*|\pi,o}, {\nu}_{\pi^*})-\frac{1}{\eta}D_{\Psi}({\nu}_{\pi^*},q)\rb.
\end{align*}
Similar to the $B$ function \eqref{eq: Q formulation} we use in Appendix \ref{appendix thm regret} and Appendix \ref{appendix arbitrary}, given the decision probability $\pi\sim p$, for all $\nu\in \Delta(\M\times \Pi)$ such that $\nu_{\pi^*}\in \inte(\Delta(\Pi))$ and $R: \Pi\times \OC\rightarrow \R^{|\Pi|}$, we define
\begin{align*}
     B(\nu, R)= \E_{\nu,p}\left[f_{M}(\pi^*)-f_{M}(\pi)+\frac{1}{\eta}\langle \nabla \Psi(q)-R[\pi,o], \mathds{1}(\pi^*)-q\rangle+\frac{1}{\eta}D_{\Psi^*}(R[\pi,o], \nabla \Psi(q))\right],
\end{align*}
where $\mathds{1}(\pi^*)$ is the vector whose $\pi^*-$coordinate is 1 but all other coordinates are 0.
Note that $B(\nu, R)$ is a strongly convex function with respect to $R$.
We have the following formula:
\begin{align*}
    \texttt{AIR}^\Psi_{q,\eta}(p,\nu)= B(\nu, \nabla \Psi({\nu}_{\pi^*|\cdot, \cdot}))= \inf_{R}B(\nu,R),
\end{align*}
where the first equality is by Lemma \ref{lemma pythegorean} and the property (b) in Lemma \ref{lemma property legendre}, and the second equality is because that the first-order optimal condition implies that the minimizer of $\inf_{R}B(\nu,R)$ will be $R=\nabla\Psi(\nu_{\pi^*|\cdot,\cdot})$, which is specified by $$R[\pi,o]=\nabla\Psi(\nu_{\pi^*|\pi,o}),\quad\forall \pi\in\Pi, o\in\OC.$$ 
By Danskin's theorem (Lemma \ref{lemma duality gradients}), we further have
\begin{align*}
    \frac{\partial \texttt{AIR}^\Psi_{q,\eta}(p,\nu) }{\partial \nu} = \left.\frac{\partial B(\nu, R)}{\partial \nu}\right \vert_{R=\nabla\Psi({\nu}_{\pi^*|\cdot, \cdot})}.
\end{align*}
By the above identity and the linearity of $B(\nu,Q)$ with respect to $\nu$, we have
\begin{align}\label{eq: identity B}
   B(\nu^*, \nabla \Psi({\nu}_{\pi^*|\cdot, \cdot}))= \air^\Psi_{q,\eta}(p,\nu)+\lla  \frac{\partial \air^\Psi_{q,\eta}(p,\nu) }{\partial \nu} , \nu^*-{\nu} \rra, \quad\forall \nu^*.
\end{align}

Following the very similar steps in proving Theorem \ref{thm regret approx}, we are able to prove Theorem \ref{thm regret approx Bregman}. By Lemma \ref{lemma pythegorean} and the property (b) in Lemma \ref{lemma property legendre}, for every $\bar{\pi}\in\Pi$ we have
\begin{align*}
    &\frac{1}{\eta}\lla \nabla \Psi(q_{t+1})-\nabla \Psi(q_t), \mathds{1}(\bar{\pi})-q_t\rra -\frac{1}{\eta}D_{\Psi^*}(\nabla \Psi(q_{t+1}), \nabla \Psi(q_t))\\
    = &\frac{1}{\eta}\lla \nabla \Psi(q_{t+1})-\nabla \Psi(q_t), \mathds{1}(\bar{\pi})-q_t\rra -\frac{1}{\eta}D_{\Psi}(q_t, q_{t+1})\\
    = &D_\Psi(\mathds{1}(\bar{\pi}), q_{t})-D_{\Psi}(\mathds{1}(\bar{\pi}), q_{t+1}).
\end{align*}
Given an arbitrary algorithm and an arbitrary algorithmic belief sequence, we set the reference probability $q_{t+1}={(\nu_t)}_{\pi^*|\pi_t,o_t}$ to aid our analysis. By the above equation we have
\begin{align}\label{eq: telescope bregman}
    \E\lb\sum_{t=1}^T\E_{t-1}\lb \frac{1}{\eta}\lla \nabla \Psi({(\nu_t)}_{\pi^*|\pi,o})-\Psi(q_t), \mathds{1}(\pi^*)-q_t\rra-\frac{1}{\eta}D_{\psi^*}(\nabla \Psi({(\nu_t)}_{\pi^*|\pi,o}), \nabla\Psi(q_t))\rb\rb\leq \frac{\text{diam}(\Psi)}{\eta}.
\end{align}
Subtracting the telescope sum \ref{eq: telescope bregman} from regret, we have
\begin{align}\label{eq: regret bregman long}
&\reg_T-\frac{\textup{diam}(\Psi)}{\eta}\nonumber\\\leq
&\E\lb \sum_{t=1}^T f_{M_t}(\pi^*)-\sum_{t=1}^T f_{M_t}(\pi_t)\rb-\frac{\textup{diam}(\Psi)}{\eta}\nonumber\\
\leq &\E\lb\sum_{t=1}^T\E_{t-1}\lb f_{M_t}(\pi^*)-f_{M_t}(\pi_t)+\frac{1}{\eta}\lla \nabla \Psi(q_t)-\nabla \Psi({(\nu_t)}_{\pi^*|\pi,o}), \mathds{1}(\pi^*)-q_t\rra+\frac{1}{\eta}D_{\psi^*}(\nabla \Psi({(\nu_t)}_{\pi^*|\pi,o}), \nabla\Psi(q_t))\rb\rb\nonumber\\
\leq &\E\lb \sum_{t=1}\sup_{\nu^*}B_t(\nu^*, \nabla \Psi({(\nu_t)}_{\pi^*|\pi,o}))\rb,
\end{align}
where the first inequality is by the definition of regret; the second inequality is by \eqref{eq: telescope}; and the third inequality is by taking the worst-case environment at every round and the definition of the $B$ function. Finally, taking the equivalent characterization  \eqref{eq: identity B} for the $B$ function into \eqref{eq: regret bregman long} proves Theorem \ref{thm regret approx Bregman}.

\hfill$\square$

\subsection{Information ratio bounds using the squared Hellinger distance}\label{appendix information ratio H}
We use the squared Hellinger distance to define $\ir_{\textup{H}}$ (instead of KL as in the definition \eqref{eq: information ratio} of $\ir$) in Theorem \ref{thm adaptive posterior sampling} (regret of APS), because technically a bounded divergence is more convenient to apply change of measure arguments. Note that there is no essensial difference between the definitions of $\ir$ and $\ir_{\textup{H}}$ in practical applications, since all currently known bounds 
on the information ratio hold for the stronger definition $\ir_{\textup{H}}$ with added absolute constants. This fact is illustrated in the following lemma.
\begin{lemma}[Information ratio bounds using the squared Hellinger distance]\label{lemma ir H}
    Define information ratio for the squared Hellinger distance by
    \begin{align*}
        \ir_{\textup{H}}(\nu,p)=\frac{\lp\E_{\nu,p}\lb f_M(\pi^*)-f_M(\pi)\rb\rp^2}{\E_{\nu,p}\lb D^2_{\textup{H}}(\nu_{\pi^*|\pi,o},\nu_{\pi^*})\rb},
    \end{align*}
    and define information ratio for Thompson Sampling by $\ir_{\textup{H}}(\textup{TS})=\sup_{\nu}\ir_{\textup{H}}(\nu,\nu_{\pi^*})$. Then for $K-$armed bandits, we have
   $ \ir_{\textup{H}}(\textup{TS})\leq 4K$;
    for problems with full information, we have 
    $ \ir_{\textup{H}}(\textup{TS})\leq 4$;
    and for $d-$dimensional linear bandits, we have 
    $\ir_{\textup{H}}(\textup{TS})\leq 4d$.
\end{lemma}

\paragraph{Proof of Lemma \ref{lemma ir H}:} We refer to \cite{russo2016information} for the standard information ratio bounds using the KL divergence for $K-$armed bandits (Proposition 3 in \cite{russo2016information}), problems with full information (Proposition 4 in \cite{russo2016information}), and $d-$dimensional linear bandits (Proposition 5 in \cite{russo2016information}). A common step in these proof is applying Pinsker's inequality for the KL divergence to convert the KL divergence in the information ratio into the square distance. For any $h: \X\rightarrow[0,1]$, Pinsker's inequality is expressed as follows: 
\begin{align}\label{eq: Pinsker KL}
(\E_{\P}[h(X)]-\E_{\mathbb{Q}}[h(X)])^2\leq \frac{1}{2}\KL(\P,\mathbb{Q}).
\end{align} Now by Lemma \ref{lemma Pinsker Hellinger}, we establish a Pinsker-type inequality using the squared Hellinger distance: for any $h: \X\rightarrow[0,1]$,
\begin{align}\label{eq: Pinsker H}
(\E_{\P}[h(X)]-\E_{\mathbb{Q}}[h(X)])^2\leq 4 D_{\textup{H}}^2(\P,\mathbb{Q}).
\end{align}
We can establish Lemma \ref{lemma ir H} by closely following Propositions 3-5 in \cite{russo2016information} line by line. The only modification in the new proof is replacing the KL divergence with the squared Hellinger distance and substituting Pinsker's inequality \eqref{eq: Pinsker KL} with the new Pinsker-type inequality \eqref{eq: Pinsker H}. The final results are nearly identical, differing only in absolute constants.

\hfill$\square$
\section{Details and proofs for bandit problems}\label{appendix MAB}
The section is organized as follows. In Appendix \ref{subsec concave parameterization}, we demonstrate the parameterization of AIR for structured bandit problems with Bernoulli rewards. In Appendix \ref{appendix APS BMAB}, we derive and analyze Simplified APS for Bernoulli Multi-Armed Bandits (Algorithm \ref{alg: BMAB}). In Appendix \ref{subsec Gaussian parameterization}, we showcase the parameterization of AIR for structured bandit problems with Gaussian rewards. In Appendix \ref{appendix Gaussian linear APS}, we present Simplified APS for Gaussian linear bandits (Algorithm \ref{alg: GLB}). It is important to note that all the calculations in Appendices \ref{subsec concave parameterization} to \ref{appendix Gaussian linear APS} can be placed within the broader framework of the exponential family, as explained in Appendix \ref{subsec exponential family}. Within this unified framework, the calculations are simplified, and the intuition becomes more immediate. In particular, minimizing the derivatives has the immediate interpretation of making the natural parameters and the log-partition functions nearly "mean-zero" across different conditions. While Appendices \ref{subsec concave parameterization} to \ref{appendix Gaussian linear APS} provide concrete examples, readers who are comfortable with generic derivations can also start by reading Appendix \ref{subsec exponential family} first. Lastly, we prove a supporting lemma to analyze Algorithm \ref{alg: BMAB} in Appendix \ref{subsec proof boundedness}.

\subsection{Concave parameterization with Bernoulli reward}\label{subsec concave parameterization}
We consider Bernoulli structured bandit  with an action set $\Pi\subset\R^d$ and a mean reward function class $\F\subset(\Pi: \mapsto [0,1])$ that is convex. (As discussed in the beginning of Section \ref{sec Bernoulli MAB}, every bounded-reward bandit problem can  equivalently be reduced to a Bernoulli bandit problem.) For simplicity we make the standard assumption that $\Pi$ is finite, which can be removed using standard discretization and covering argument. The goal here is to make the computation complexity to be independent of the size of model class $\M$, but only depends on $|\Pi|$.  The general principle to achieve this goal is as follows. For each possible value of $\pi^*$, we assign an ``effective model'' $M_{\pi^*}$ to $\pi^*$ so that the optimization problem \eqref{eq: computation problem} reduces to selecting those $|\Pi|$ ``effective models,'' as well as the probability distribution over them. 

We introduce the following parameterization: $\forall a,a^*\in\Pi$, 
\begin{align}\label{eq: parameterization Bernoulli}
    \theta_{a^*}(a)=\E\lb r(a)|\pi^*=a^* \rb,
    \nonumber\\\alpha(a^*)={\nu}_{\pi^*}(a^*),\nonumber\\
    \beta_{a^*}(a)= \alpha(a^*)\cdot \theta_{a^*}(a).
\end{align}
Then we have represent AIR by $(\alpha, \{\beta_{\pi^*}\}_{\pi^*\in\Pi})$: 
\begin{align}\label{eq: AIR parameterization Bernoulli}
      \air_{q,\eta}(p,\nu)= \sum_{\pi^*\in \Pi}\beta_{\pi^*}(\pi^*)-\sum_{\pi^*,\pi\in[K]}p(\pi)\beta_{\pi^*}(\pi)\nonumber\\-\frac{1}{\eta}\sum_{\pi^*,\pi\in [K]}p(\pi)\alpha(\pi^*)\kl\lp\frac{\beta_{\pi^*}(\pi)}{\alpha(\pi^*)}, \sum_{\pi^*\in\Pi}{\beta_{\pi^*}}(\pi)\rp-\frac{1}{\eta}\KL(\alpha,q),
\end{align}
and the constraint of $(\alpha, {\beta_{\pi^*}}_{\pi^*\in\Pi})$ is that  the  functions parameterized by $\theta_{a^*}$ belong to the mean reward function class $\F$. We know the constraint set of $(\alpha, {\beta_{\pi^*}}_{\pi^*\in\Pi})$ to be convex because the convexity of perspective function.

Now we want to prove that the AIR objective in the maximization problem \eqref{eq: computation problem} is concave. Using the definition \eqref{eq: Q formulation} of $B$ function in Appendix \ref{appendix thm regret} (this $B$ function will also be  the key to simplify our calculation of derivatives in the next section), we have
\begin{align}\label{eq: B function Bernoulli}
   B(\nu, Q)= \E_{(M,\pi^*)\sim\nu, \pi\sim p, o\sim M(\pi)}\lb f_M(\pi^*)-f_M(\pi)+\frac{1}{\eta}\log \frac{q(\pi^*)}{ Q[\pi,o](\pi^*)} \rb\nonumber\\
   = \sum_{\pi^*} \beta_{\pi^*}(\pi^*)-\sum_{\pi,\pi^*}p(\pi)\beta_{\pi^*}(\pi)+\frac{1}{\eta}\sum_{\pi,\pi^*}p(\pi)\beta_{\pi^*}(\pi)\log \frac{q(\pi^*)}{Q[\pi,1](\pi^*)}\nonumber\\+\frac{1}{\eta}\sum_{\pi,\pi^*}p(\pi)(\alpha(\pi^*)-\beta_{\pi^*}(\pi^*))\log \frac{q(\pi^*)}{Q[\pi,0](\pi^*)}.
\end{align}
This means that after parameterizing $\nu$ with $\alpha$ and $\{\beta_{\pi^*}\}_{\pi^*\in\Pi}$, $B(\nu,Q)$ will be a linear function of $(\alpha, \{\beta_{\pi^*}\}_{\pi^*\in\Pi})$. By the relationship between the $B$ function and AIR showed in Appendix \ref{subsec concave parameterization}, we have 
\begin{align*}
    \air_{q,\eta}(p,\nu)=\inf_{Q}B(\nu, Q),
\end{align*}
which implies that AIR will be a concave function of $(\alpha, \{\beta_{\pi^*}\}_{\pi^*\in\Pi})$. So the optimization problem to maximize AIR is a convex optimization problem, whose computational complexity will be poly-logarithmic to the cardinality of $(\alpha, \{\beta_{\pi^*}\}_{\pi^*\in\Pi})$. As a result, the computational complexity to maximize AIR is polynomial in $|\Pi|$ and does not depends on cardinality of the model class. This discussion shows that we give finite-running-time algorithm with computational complexity $\text{poly}(e^d)$ even when the cardinality of model class is double-exponential. Still, the computation is 
only efficient for simple problems such as $K-$armed bandits, but we also give efficient algorithm for linear bandits in Appendix \ref{subsec linear bandits}.

\subsection{Simplified APS for Bernoulli MAB}\label{appendix APS BMAB}
For Bernoulli $K-$armed bandits discussed in in Section \ref{subsec APS BMAB}, we give the details about how to use first-order optimality conditions to derive Algorithm \ref{alg: BMAB}. 
For Bernoulli $K-$armed bandits, the decision space $\Pi=[K]=\{1,\cdots, K\}$ is a finite set of $K$ actions. We take the parameterization \eqref{eq: parameterization Bernoulli} introduced in Appendix \ref{subsec concave parameterization}, and utilize the expression \eqref{eq: AIR parameterization Bernoulli} of AIR and the expression \eqref{eq: AIR parameterization Bernoulli} of the B function. By \eqref{eq: AIR parameterization Bernoulli}, we have the following concave parameterization of AIR by the $K(K+1)-$dimensional vector $(\alpha, {{\beta}})=(\alpha, \beta_1,\cdots, \beta_K)$: 
\begin{align}\label{eq: AIR parameterization BMAB}
      \air_{q,\eta}(p,\nu)= \sum_{i\in [K]}\beta_{i}(i)-\sum_{i,j\in[K]}p(j)\beta_i(j)\nonumber\\-\frac{1}{\eta}\sum_{i,j\in [K]}p(j)\alpha(i)\kl\lp\frac{\beta_i(j)}{\alpha(i)}, \sum_{i\in[K]}{\beta_i}(j)\rp-\frac{1}{\eta}\KL(\alpha,q),
\end{align}
 By \eqref{eq: B function Bernoulli} we have the following expression of the $B$ function, \begin{align}\label{eq: B function BMAB}
    B(\nu, Q)
   = \sum_{i} \beta_{i}(i)-\sum_{j,i}p(j)\beta_{i}(j)+\frac{1}{\eta}\sum_{j,i}p(j)\beta_{i}(j)\log \frac{q(i)}{Q[j,1](i)}\nonumber\\+\frac{1}{\eta}\sum_{j,i}p(j)(\alpha(i)-\beta_{i}(i))\log \frac{q(i)}{Q[j,0](i)},
\end{align}which satisfies 
\begin{align*}
    \air_{q,\eta}(p,\nu)=\inf_{Q}B(\nu, Q),
\end{align*}
and will serve the role to simplify our following calculation of the derivatives. We denote ${\nu}_{\pi^*|j,1}(i)$ as the marginal posterior $\P(\pi^*=i|\pi=j,o=1)$ when the underlying probability measure is $\nu$.

\subsubsection{Derivation of Algorithm \ref{alg: BMAB}}
By \eqref{eq: B function BMAB} and Lemma \ref{lemma duality gradients} (i.e., using Danskin's theorem and the bivariate function \eqref{eq: B function BMAB} to calculate the derivatives is easier than directly calculating the derivatives of the AIR parameterization \eqref{eq: AIR parameterization BMAB}), we have for each $i\in[K]$, 
\begin{align}\label{eq: theta i i bernoulli}
    \frac{\partial \air_{q,\eta}(p,\nu)}{\partial \beta_{i}(i)}
    = \nonumber&\left.\frac{\partial B(\nu, Q)}{\partial \beta_i(i)}\right\vert_{Q=\nu_{\pi^*|\cdot, \cdot}}\\=&(1-p(i))- \frac{1}{\eta}p(i)\lp\log{\nu}_{\pi^*|i,1}(i)-\log{\nu}_{\pi^*|i,0}(i)\rp.
\end{align}
And for every $i\neq j\in [K]$, 
\begin{align}\label{eq: theta i j bernoulli}
    \frac{\partial \air_{q,\eta}(p,\nu)}{\partial \beta_{i}(j)}
    =&\left.\frac{\partial B(\nu, Q)}{\partial \beta_i(j)}\right\vert_{Q=\nu_{\pi^*|\cdot, \cdot}}\nonumber\\=&-p(j)-\frac{1}{\eta}p(j)\lp\log{\nu}_{\pi^*|j,1}(i)-\log{\nu}_{\pi^*|j,0}(i)\rp.
\end{align}
Lastly, for each $i\in [K]$, 
\begin{align}\label{eq: alpha i bernoulli}
    \frac{\partial \air_{q,\eta}(p,\nu)}{\partial\alpha(i)}=
  &\left.\frac{\partial B(\nu, Q)}{\partial \alpha(i)}\right\vert_{Q=\nu_{\pi^*|\cdot, \cdot}}\nonumber\\=&\frac{1}{\eta}\sum_{j\in[K]}p(j)\lp\log q(i)-\log {\nu}_{\pi^*|j,0}(i)\rp.
\end{align}
We let the derivatives in \eqref{eq: theta i i bernoulli} and \eqref{eq: theta i j bernoulli} be zero, which means that the derivatives with respect to all coordinates of $\beta$ are zero. We have for all $j\in[K]$
\begin{align}\label{eq: bernoulli equations}
    \log \frac{{\nu}_{\pi^*|j,1}(j)}{{\nu}_{\pi^*|j,0}(j)}&=\frac{\eta}{p(j)}-\eta,\nonumber\\
    \log \frac{\nu_{\pi^*|i,1}(j)}{\nu_{\pi^*|i,0}(j)}&=\log \frac{1-{\nu}_{\pi^*|j,1}(j)}{1-{\nu}_{\pi^*|j,0}(j)}=-\eta, \quad \forall i\neq j.
\end{align}
Solving the above two equations in \eqref{eq: bernoulli equations} we obtain closed form solutions for $\nu_{\pi^*}(j|j,1)$ and $\nu_{\pi^*}(j|j,0)$. Therefore we have
\begin{align}\label{eq: posterior bernoulli}
    {\nu}_{\pi^*|j,1}(j)&=\frac{1-\exp(-\eta)}{1-\exp(-\eta/p(j))}, \quad \forall j\in [K],\nonumber\\
    {\nu}_{\pi^*|j,1}(i)&=\frac{\exp(-\eta)-\exp(-\eta/p(j))}{1-\exp(-\eta/p(j))}\cdot \frac{\beta_i(j)}{\sum_{i}\beta_i(j)-\beta_j(j)}, \quad \forall i\neq j\in [K],\nonumber\\
    {\nu}_{\pi^*|j,0}(j)&=\frac{\exp(\eta)-1}{\exp(\eta/p(j))-1}, \quad \forall j\in [K],\nonumber\\
    {\nu}_{\pi^*|j,0}(i)&=\frac{\exp(\eta/p(j))-\exp(\eta)}{\exp(\eta/p(j))-1}\cdot \frac{\beta_i(j)}{\sum_{i}\beta_i(j)-\beta_j(j)}, \quad \forall i\neq j\in [K].
\end{align}
 And from \eqref{eq: alpha i bernoulli} and \eqref{eq: bernoulli equations} we have for every $i\in [K]$,
\begin{align}\label{eq: bernoulli alpha 1}
    \frac{\partial \air_{q,\eta}(p,\nu)}{\partial\alpha(i)}
    =&\frac{1}{\eta}\sum_{j\in[K]}p(j)\log \frac{q(i)}{{\nu}_{\pi^*|j,0}(i)}\nonumber\\
    =&\frac{1}{\eta}\sum_{j\in[K]}p(j)\log\frac{q(i)}{\nu_{\pi^*|j,1}(i)},
\end{align}
where the first equality is because 
of \eqref{eq: alpha i bernoulli} and the last equality is because of \eqref{eq: bernoulli equations}.

Now we set $\alpha=p=q$ so that the posterior updates \eqref{eq: posterior bernoulli} all have closed forms. This selection results in Algorithm \ref{alg: BMAB}. In the following part  we present the regret analysis of Algorithm \ref{alg: BMAB} and the proof to Theorem \ref{thm BMAB}.

\subsubsection{Regret analysis of Algorithm \ref{alg: BMAB}} We have shown that the derivatives of AIR with respect to all $\{\beta_{i}(j)\}_{i,j\in[K]}$ are zeros, so the optimization error in Theorem \ref{thm regret approx} will happens in the coordinates in $\alpha$ rather than $\beta$. We then need to control the optimization error incurred because of the derivatives of $\alpha$.
For this purpose we prove the following lemma (proof deferred to Appendix \ref{subsec proof boundedness}).
\begin{lemma}[Boundedness of derivatives]\label{lemma boundedness}
    In Bernoulli MAB, under the equations \eqref{eq: bernoulli equations} (which is the result of setting the derivatives of AIR with respect to all $\{\beta_i(j)\}_{i,j\in[K]}$ to be zeros), if we set $p=q=\alpha\in\textup{int}(\Delta([K]))$, then we have
    \begin{align*}
       -\eta\leq \frac{\partial \air_{q,\eta}(p,\nu)}{\partial \alpha(i)}\leq \eta, \quad \forall i\in [K].
    \end{align*}
\end{lemma}

Now we have shown that the derivatives of AIR with respect to all $\{\beta_{i}(j)\}_{i,j\in[K]}$ are zeros, and the derivatives of AIR with respect to all $\{\alpha(i)\}_{i\in [K]}$ are suitably bounded. Combining Theorem \ref{thm regret approx}, Lemma \ref{lemma boundedness}, and Lemma \ref{lemma air ref alg}, we prove Theorem \ref{thm BMAB} in a straightforward manner.
\vspace{0.1in}

\noindent{\bf Theorem 4.1} (Regret of Simplified APS for Bernoulli MAB). {\it The regret of Algorithm \ref{alg: BMAB} with $\eta= \sqrt{\log K/(2KT+4T)}$ and $T\geq 3$ is bounded by
\begin{align*}
    \reg_T\leq 2\sqrt{2(K+2)T\log K}.
\end{align*}
}

\paragraph{Proof of Theorem \ref{thm BMAB}:} By Theorem \ref{thm regret approx} (regret bound using approximate belief maximizers)  and  Lemma \ref{lemma boundedness} (boundedness of derivatives), we have
\begin{align}\label{eq: thm bmab 1}
    \reg_T\leq \frac{\log K}{\eta} +\E\lb \sum_{t=1}^T\lp \air_{p_t, \eta}(p_t, \nu_t)+2\eta\rp\rb.
\end{align}
By Lemma \ref{lemma air ref alg} (bounding AIR by IR for TS) and the fact that information ratio of TS is bounded by $4K$ for MAB (Lemma \ref{lemma ir H}), for any $\eta\in (0,1/3]$ we have
\begin{align}\label{eq: thm bmab 2}
    \air_{p_t, \eta}(p_t, \nu_t)\leq 2{\eta}(K+1).
\end{align}
Combining \eqref{eq: thm bmab 1} and \eqref{eq: thm bmab 2} and optimizing for $\eta$, we prove Theorem \ref{thm BMAB}.

\hfill$\square$

We would like to comment that, by using the language of the exponential family and the log-partition function introduced in Appendix \ref{subsec exponential family}, Lemma \ref{lemma boundedness} is equivalent to demonstrating that a ``mean-zero'' property of the natural parameter implies an approximate ``mean-zero'' property of the log-partition function. This conclusion appears quite natural, given that the log-partition function for the Bernoulli distribution has a flat growth and exhibits the so-called ``local norm'' property. Such analysis is generic and can be applied broadly, rather than solely serving for the single lemma here. The formal proof of Lemma \ref{lemma boundedness} can be found in Appendix \ref{subsec proof boundedness}.

We believe that our approach is no more complicated than the standard proofs in the bandit and RL literature. Once one understands the generic machinery for maximizing AIR and why the derivatives are bounded (e.g., for simple bandit problems, this can often be explained through the natural parameter and the log-partition function as in Appendix \ref{subsec exponential family}), it becomes convenient to principly derive optimal algorithms.

\subsection{Proof of Lemma \ref{lemma boundedness}}\label{subsec proof boundedness}
{\bf Lemma B.1} (Boundedness of derivatives). {\it
    In Bernoulli MAB, under the equations \eqref{eq: bernoulli equations} (which is the result of setting the derivatives of AIR with respect to all $\{\beta_i(j)\}_{i,j\in[K]}$ to be zeros), if we set $p=q=\alpha\in\textup{int}(\Delta([K]))$, then we have
    \begin{align*}
       -\eta\leq \frac{\partial \air_{q,\eta}(p,\nu)}{\partial \alpha(i)}\leq \eta, \quad \forall i\in [K].
    \end{align*}
}

\paragraph{Equivalent expression of Lemma \ref{lemma boundedness} in the language of log-partition function:} 
We firstly rewrite Lemma \ref{lemma boundedness} in the language of exponential family. Intuition introduced in Appendix \ref{subsec exponential family} are quite helpful for understanding the naturalness of our approach, while such knowledge is not required and the proof here will be self-contained.

It is easy to check that Algorithm \ref{alg: BMAB} satisfies the optimality condition stated in \eqref{eq: bernoulli equations}: 
\begin{align}\label{eq: bernoulli optimal condition}
    \log \frac{\nu_{\pi^*|j,1}(i)}{\nu_{\pi^*|j,0}(i)}=\eta\frac{\mathds{1}(i=j)}{p(j)}-\eta, \quad \forall i,j\in[K]
\end{align} and $\alpha=p=q$ at every round (we omit the subscript $t$ for $\alpha, p, q, \nu$). Furthermore, it is  easy to check that \eqref{eq: bernoulli optimal condition} is equivalent with the following ``mean-zero'' property (with respect to $j\sim p$) of the functions we refer to as ``natural parameters:''
\begin{align}\label{eq: bernoulli g}
    g_{i}(j)-g_{\avg}(j)=\eta\frac{\mathds{1}(i=j)}{p(j)}-\eta, \quad \forall i,j\in [K],
\end{align}
where we define
\begin{align*}
    g_{i}(j)&=g(\theta_{i}(j))=\log\frac{\theta_{i}(j)}{1-\theta_{i}(j)}, \\
    g_{\avg}(j)&=g(\theta_{\avg}(j))= \log\frac{\theta_{\avg}(j)}{1-\theta_{\avg}(j)}
\end{align*}
for the conditional mean $\theta_{i}(j)=\E[r(j)|\pi^*=i]$ and mean $\theta_{\avg}(j)=\E[r(j)]$. One can  straightforwardly check the equivalence between \eqref{eq: bernoulli optimal condition} and \eqref{eq: bernoulli g} from the posterior expression
\begin{align}\label{eq: posterior expression}
    \nu_{\pi^*|j,1}(i)&= \frac{\alpha(i)\theta_{i}(j)}{\sum_{i\in[K]}\alpha(i)\theta_{i}(j)}= \frac{\alpha(i)\theta_{i}(j)}{\theta_{\avg}(j)}, \nonumber\\
    \nu_{\pi^*|j,0}(i)&= \frac{\alpha(i)(1-\theta_{i}(j))}{\sum_{i\in[K]}\alpha(i)(1-\theta_{i}(j))}= \frac{\alpha(i)(1-\theta_{i}(j))}{1-\theta_{\avg}(j)}.
\end{align}
Alternatively, if the reader has been familiar with our derivation in Appendix \ref{appendix APS BMAB} and the exponential family interpretation in Appendix \ref{subsec exponential family}, then it is immediate to understand that both  \eqref{eq: bernoulli g} is the consequence of setting all the derivatives w.r.t. $\beta$ to be $0$.   

In addition, we can verify that when $p=q$, 
\begin{align}\label{eq: bernoulli lemma alpha}
    \frac{\partial \air_{q,\eta}(p,\nu)}{\partial \alpha(i)}=\frac{1}{\eta}\E_{j\sim p}\lb \bar{A}(g_{i}(j))-\bar{A}(g_{\avg}(j))\rb,
\end{align}
where $\bar{A}$ is the log-partition function with respect to the natural parameter $g$: 
\begin{align*}
    \bar{A}(g)=\log(1+e^g). 
\end{align*}
\eqref{eq: bernoulli lemma alpha} can be verified by using the derivative formula \eqref{eq: bernoulli alpha 1} and the posterior expression \eqref{eq: posterior expression} straightforwardly.  Again, if the reader has been familiar with our general derivation about the exponential family in Appendix \ref{subsec exponential family}, the derivative expression \eqref{eq: bernoulli lemma alpha} using the log-partition function  will be an immediate result.  

Given \eqref{eq: bernoulli g} and \eqref{eq: bernoulli lemma alpha}, we are now ready to equivalently rewrite Lemma \ref{lemma boundedness} as the following:
the ``mean-zero'' property \eqref{eq: bernoulli g} of the natural parameter with respect to $j\sim p$ implies an approximate ``mean-zero'' property of the log-partition function with respect to $j\sim p$.
\begin{lemma}[Equivalent expression of Lemma \ref{lemma boundedness}]\label{lemma equivalent}
  {\it Given the equations \eqref{eq: bernoulli g} and $\alpha=p=q$, for every fixed $i\in[K]$, we have that}
\begin{align*}
 -\eta\leq  \E_{j\sim p}\lb \bar{A}(g_{i}(j))-\bar{A}(g_{\avg}(j))\rb\leq \eta, 
\end{align*}  
where $\bar{A}$ is the log-partition function defined by \begin{align*}
    \bar{A}(g)=\log(1+e^g). 
\end{align*}
\end{lemma}

\paragraph{Proof of Lemma \ref{lemma equivalent} and Lemma \ref{lemma boundedness}:}

Now we prove Lemma \ref{lemma equivalent} in order to prove Lemma \ref{lemma boundedness}.

Because $0\leq\log((1+e^x)/(1+e^y))\leq x-y$ for all $x\geq y\in\R$, and $x-y\leq\log((1+e^x)/(1+e^y))\leq 0$ for all $x\leq y\in\R$, we have that
\begin{align}\label{eq: log partition inequality}
    \min\{0,x-y\}\leq \bar{A}(x)-\bar{A}(y)\leq \max\{x-y,0\}
\end{align}
Therefore, by \eqref{eq: log partition inequality} and \eqref{eq: bernoulli g} we have the upper bound
\begin{align*}
     &\E_{j\sim p}\lb \bar{A}(g_i(j))-\bar{A}(g_{\avg}(j))\rb\\\leq &\E_{j\sim p}\lb \max\{g_{i}(j)-g_{\avg}(j),0\}\rb\\\leq &\E_{j\sim p}\lb \eta\frac{\mathds{1}(i=j)}{p(j)}\rb\\
     =&\eta, 
 \end{align*} 
 and the lower bound
 \begin{align*}
     &\E_{j\sim p}\lb \bar{A}(g_i(j))-\bar{A}(g_{\avg}(j))\rb\\\geq &\E_{j\sim p}\lb \min\{0, g_{i}(j)-g_{\avg}(j)\}\rb\\\geq &-\eta. 
 \end{align*} 
 As a result,
 we finish the proof of Lemma \ref{lemma equivalent} and consequently prove the original Lemma \ref{lemma boundedness}.

\hfill$\square$


\subsection{Concave parameterization with Gaussian reward}\label{subsec Gaussian parameterization}
For structured bandit with Gaussian reward structure, we can  formulate the optimization problem as follows. To facilitate the handling of mixture distributions, we introduce the ``homogeneous noise'' assumption as follows: for all actions $\pi\in\Pi$, the reward of $\pi$ is denoted as $r(\pi) = \E[r(\pi)] + \epsilon$, where $\epsilon \sim N(0,1)$ is Gaussian noise that is identical across all actions (meaning all actions share the same randomness within the same rounds). This ``homogeneous noise'' simplifies our expression of AIR, and the resulting algorithms remain applicable to independent noise models and the broader sub-Gaussian setting.

We introduce the following parameterization of belief $\nu$: $\forall a,a^*\in\Pi$, 
\begin{align*}
    \theta_{a^*}(a)=\E_{\nu}\lb r(a)|\pi^*=a^* \rb,
    \nonumber\\\alpha(a^*)={\nu}_{\pi^*}(a^*),\nonumber\\
    \beta_{a^*}(a)= \alpha(a^*)\cdot \theta_{a^*}(a).
\end{align*} Note that AIR in this ``homogeneous noise'' setting can be expressed as 
\begin{align}\label{eq: oair}
    \oair_{q,\eta}(p,\nu)=\E\lb \theta_{\pi^*}(\pi^*)-\theta_{\pi^*}(\pi)-\frac{1}{\eta} \KL(N(\theta_{\pi^*}(\pi), \sigma^2), N(\theta_{\avg}(\pi), \sigma^2))-\frac{1}{\eta}\KL(\alpha, q)\rb,
\end{align}
where we denote
\begin{align*}
\theta_{\avg}=\sum_{\pi^*\in\Pi}\alpha(\pi^*)\theta_{\pi^*}.
\end{align*}
Similar to the definition of $B$ function in Appendix \ref{appendix thm regret} and Appendix \ref{subsec concave parameterization}, we define the following $\overline{B}$ function, which can simplify our calculation of derivatives in the next section. Note that in the following definition of the $\overline{B}$ function,  its second argument is a belief $\omega\in\Delta(\M\times\Pi)$, and we replace the mapping $Q: \Pi\times \OC\rightarrow \Delta(\Pi)$ in the original definition \eqref{eq: Q formulation} of the $B$ function in Appendix \ref{appendix thm regret} to the marginal posterior mapping $\omega_{\pi^*|\cdot,\cdot}: \Pi\times\OC\rightarrow\Delta(\Pi)$. To be specific, we define 
\begin{align*}
    \overline{B}(\nu,\omega)
=B(\nu, \omega_{\pi^*|\cdot,\cdot})
= \E_{(M,a^*)\sim\nu, a\sim p, o\sim M(a)}\lb \theta_{a^*}(a^*)-\theta_{a^*}(a)+\frac{1}{\eta}\log \frac{q(a^*)}{\omega_{\pi^*|a,o}(a^*)}\rb,
\end{align*}
which can be expressed as
\begin{align}\label{eq: B function Gaussian}
    &\overline{B}(\nu, \omega)\nonumber\\
    =&\E_{(M,a^*)\sim\nu, a\sim p}\lb \theta_{a^*}(a^*)-\theta_{a^*}(a)+\frac{1}{2\eta}\lp(\theta^\omega_{a^*}(\pi)-r(a))^2-(\theta^\omega_{\avg}(a)-r(a))^2\rp-\frac{1}{\eta}\log\frac{\alpha^{\omega}(a^*)}{q(a^*)}\rb\nonumber\\
    =& \sum_{a^*}\beta_{a^*}(a^*)-\sum_{a,a^*}p(a)\beta_{a^*}(a)+\frac{1}{2\eta}\sum_{a,a^*}p(a)\alpha(a^*)\lp\theta_{a^*}^\omega(a)^2-\theta_{\text{avg}}^\omega(a)^2\rp\nonumber\\&+\frac{1}{\eta}\sum_{a,a^*}p(a)(\theta_{\text{avg}}^\omega(a)-\theta_{a^*}^\omega(a))\beta_{a^*}(a)-\frac{1}{\eta}\sum_{a^*}\alpha(a^*)\log\frac{\alpha^\omega(a^*)}{q(a^*)},
\end{align}
where $\alpha^\omega(a^*)=\omega_{\pi^*}(a^*)$ and $\theta^\omega_{a^*}(a)=\E_{\omega}[r(a)|\pi^*=a^*]$ following the parameterization of $\omega$ (which is independent with the parameterization $(\alpha, \beta)$ of $\nu$ without the superscript).
By the relationship between the $B$ function and AIR showed in Appendix \ref{subsec concave parameterization}, we have 
\begin{align*}
    &\oair_{q,\eta}(p,\nu)\\=&\inf_{\omega} \overline{B}(\nu, \omega)\\
    =&\int_{\A}\beta_{a^*}^\top a^*da^*-\int_{\A}\int_{\A}p(a)\beta_{a^*}^\top a da^*da\nonumber\\&-\frac{1}{2\eta}\int_{\A}\int_{\A}p(a)\alpha(a^*)\lp\frac{\beta_{a^*}^\top a}{\alpha(a^*)}-\int_{\A}\beta_{a^*}^\top a da^*\rp^2da-\frac{1}{\eta}\KL(\alpha,q).
\end{align*}
This means that $\oair_{q,\eta}(p,\nu)$ will be a concave function of $(\alpha, \{\beta_{\pi^*}\}_{\pi^*\in \Pi})$. 

\subsection{Simplified APS for Gaussian linear bandits}\label{appendix Gaussian linear APS}
 In this subsection we derive Algorithm \ref{alg: GLB} for adversarial linear bandits with Gaussian reward. We work in the setting discussed in Section \ref{subsec linear bandits}, where the mean reward is bounded in $[-1,1]$ and all the rewards follow Gaussian distribution with variance 1. As the decision space is an $d-$dimensional action set $\Pi=\A\subseteq\R^d$, we will use the notations $\A$ (as action set), $a$ (as action) and $a^*$ (as optimal action) to follow the tradition of literature about linear bandits.

By \eqref{eq: B function Gaussian} and
and Lemma \ref{lemma duality gradients}, we have for each $a^*\in \A$,
\begin{align}\label{eq: theta i i}
    \frac{\partial \oair_{q,\eta}(p,\nu)}{\partial \beta_{a^*}}
    =&\left.\frac{\partial \overline{B}(\nu,\omega)}{\partial \beta_{a^*}}\right\vert_{\omega=\nu}
    \nonumber\\=&a^*-\E_{a\sim p}[a]- \frac{1}{\eta}\E_{a\sim p}[aa^\top]\lp\theta_{a^*}-{\theta}_{\text{avg}}\rp .
\end{align}
And for each $a^*\in \A$, 
\begin{align}\label{eq: alpha i}
    \frac{\partial \oair_{q,\eta}(p,\nu)}{\partial\alpha(a^*)}=&\left.\frac{\partial \overline{B}(\nu,\omega)}{\partial \alpha(a^*)}\right\vert_{\omega=\nu}\nonumber
   \\ =&\frac{1}{2\eta}\int_{\A}p(a)\lp\theta_{a^*}^\top a\rp^2da-\frac{1}{\eta}\log\frac{ \alpha(a^*)}{q(a^*)}.
\end{align}

Let the derivatives in \eqref{eq: theta i i} be close to zero. If the matrix $\E_{a\sim p}[aa^\top]$ have full rank, then we have
\begin{align*}
\theta_{a^*}-\theta_{\avg}&\approx\eta(\E_{a\sim p}[aa^\top])^{-1}(a^*-\E_{a\sim p}[a]), \quad\forall a^*\in \A,\\
\alpha&\approx p.
\end{align*}
We ask $\E_{a\sim p}[aa^\top]$ to be positive definite, and let
\begin{align}\label{eq: gaussian oair solution}
 \theta_{a^*}=\frac{\beta_{a^*}}{\alpha(a^*)}=\eta (\E_{a\sim p}[aa^\top])^{-1}a^*, \quad \forall a^*\in \A. 
\end{align}
Taking the relationship \eqref{eq: gaussian oair solution} into \eqref{eq: alpha i}, we have 
\begin{align}\label{eq: derivative alpha gaussian}
     \frac{\partial \oair_{q,\eta}(p,\nu)}{\partial \alpha(a^*)}=\frac{\eta}{2}(a^*)^\top(\E_{a\sim p}[aa^\top])^{-1}a^*-\frac{1}{\eta}\log\frac{\alpha(a^*)}{q(a^*)}.
\end{align}
Aiming to make the derivatives \eqref{eq: derivative alpha gaussian} to be zero, we propose the following approximate solutions that makes the derivatives in \eqref{eq: theta i i} to be nearly zero:
\begin{align}\label{eq: alpha zero}
    p&=q,\nonumber\\
    \theta_{a^*}&=\frac{\beta_{a^*}}{\alpha(a^*)}=\eta (\E[aa^\top])^{-1}a^*, \quad \forall a^*\in \A,\nonumber\\
    \alpha(a)&\propto q(a)\exp \lp \frac{\eta}{2}a^\top (\E_{a\sim p}[aa^\top])^{-1}a\rp, \quad \forall a^*\in \A
\end{align}

By Bayes' rule and \eqref{eq: gaussian oair solution}, the posterior update $\nu_{a^*|a,r(a)}$ can be expressed as follows.
Given $\bar{a}\in\A$, we have
\begin{align*}
    \nu_{a^*|a,r(a)}(\bar{a})
    = \frac{\alpha(\bar{a})\exp(-\frac{1}{2}(r(a)-\theta_{\bar{a}}^\top a)^2)}{\int_{\A}\alpha(a^*)\exp(-\frac{1}{2}(r(a)-\theta_{a^*}(a))^2)da^*}\\
    = \frac{\alpha(\bar{a})\exp\lp r(a)\theta_{\bar{a}}^\top a-\frac{1}{2}(\theta_{\bar{a}}^\top a)^2\rp}{\int_{\A}\alpha(a^*)\exp\lp r(a)\theta_{a^*}^\top a-\frac{1}{2}(\theta_{a^*}^\top a)^2\rp da^*}.
\end{align*}
The resulting algorithm is  an exponential weight algorithm
\begin{align*}
    p_t(a)\propto \exp\lp\eta \sum_{i=1}^{t-1}\hat{r_i}(a)\rp,
\end{align*}with the modified inverse probability weighting (IPW) estimator
\begin{align*}
    \hat{r_t}(a)= \underbrace{a^\top(\E_{a\sim p}[aa^\top])^{-1}a_t r_t(a_t)}_{\text{IPW estimator}} +\underbrace{\frac{\eta}{2} \lp a( \E_{a\sim p_t} [aa^\top])^{-1} a- (a^\top (\E_{a\sim p_t}[aa^\top])^{-1}a_t)^2\rp}_{\text{mean zero regularizer}}.
\end{align*}
The requirement that $\E_{a\sim p}[aa^\top]$ is positive definite can be ensured with the help of the volumetric spanners constructed in \cite{hazan2016volumetric}. In this way we have derived Algorithm \ref{alg: GLB}.

\subsection{Solving AIR with general exponential family distributions}\label{subsec exponential family}
For structured bandits with reward distribution belonging to the exponential family, we formulate the optimization problem as follows. To facilitate the handling of mixture distributions, we consider the ``homogeneous noise'' assumption as follows: for all actions $a\in\Pi$, the reward $r(a)$ of $a$ follows the exponential family distribution $r(a)\sim P_{\theta(a)}$, and all actions share the same randomness within the same rounds. This ``homogeneous noise'' simplifies our expression of AIR, and the resulting algorithms remain applicable to independent noise models. We assume the exponential family distribution has the probability density function 
\begin{align*}
    P_{\theta(a)}(r(a)=o)\propto h(o) \exp(g(\theta(a))T(o)-A(\theta(a))), \quad \forall a\in \Pi.
\end{align*}

We introduce the following parameterization of belief $\nu$: $\forall a,a^*\in\Pi$, 
\begin{align*}
    \theta_{a^*}(a)=\E_{\nu}\lb r(a)|\pi^*=a^* \rb,
    \\\alpha(a^*)={\nu}_{\pi^*}(a^*),\\
    \beta_{a^*}(a)= \alpha(a^*)\cdot \theta_{a^*}(a).
\end{align*} Note that AIR in this ``homogeneous noise'' setting can be expressed as
\begin{align}
    \oair_{q,\eta}(p,\nu)=\E\lb \theta_{\pi^*}(\pi^*)-\theta_{\pi^*}(\pi)-\frac{1}{\eta} \KL(P_{\theta_{\pi^*}(\pi)}, P_{\theta_{\avg}(\pi)})-\frac{1}{\eta}\KL(\alpha, q)\rb,\nonumber
\end{align}
where we denote
\begin{align*}
\theta_{\avg}=\sum_{\pi^*\in\Pi}\alpha(\pi^*)\theta_{\pi^*}.
\end{align*}
Similar to Appendix \ref{subsec Gaussian parameterization}, we define the following $\overline{B}$ function, which can simplify our calculation of derivatives. To be specific, we define 
\begin{align*}
    \overline{B}(\nu,\omega)
=B(\nu, \omega_{\pi^*|\cdot,\cdot})
= \E_{(M,a^*)\sim \nu, a\sim p, o\sim M(a)}\lb \theta_{a^*}(a^*)-\theta_{a^*}(a)+\frac{1}{\eta}\log \frac{q(a^*)}{\omega_{\pi^*|a,o}(a^*)}\rb,
\end{align*}which can be expressed as
\begin{align}
    &\overline{B}(\nu, \omega)\nonumber\\
    =& \sum_{a^*}\beta_{a^*}(a^*)-\sum_{a,a^*}p(a)\beta_{a^*}(a)+\frac{1}{\eta}\log \frac{q(a^*)}{\alpha^\omega(a^*)}\nonumber\\&+\frac{1}{\eta}\sum_{a^*,a}p(a)\alpha(a^*)\int_{\OC}( g(\theta^\omega_{\text{avg}}(a))-g(\theta^\omega_{a^*}(a)))T(o)do+\frac{1}{\eta}\sum_{a^*,a}p(a)\alpha(a^*)\lb A(\theta^\omega_{a^*}(a))-A(\theta^\omega_{\avg}(a))\rb.\nonumber
\end{align}
where $\alpha^\omega(a^*)=\omega_{\pi^*}(a^*)$ and $\theta^\omega_{a^*}(a)=\E_{\omega}[r(a)|\pi^*=a^*]$ following the parameterization of $\omega$ (which is independent with the parameterization $(\alpha, \beta)$ of $\nu$  without the superscript).
By the relationship between the $B$ function and AIR showed in Appendix \ref{subsec concave parameterization}, we have 
\begin{align*}
    \oair_{q,\eta}(p,\nu)=\inf_{\omega} \overline{B}(\nu, \omega).
\end{align*}

Now we consider exponential family where the sufficient statistics $T(o)=o$. Common exponential families with sufficient statistics $T(o)=o$ include Bernoulli distribution, Poisson distribution, exponential distribution,  Gaussian distribution with known variance $1$, etc. Then we have
\begin{align}
    &\overline{B}(\nu, \omega)\nonumber\\
    =& \sum_{a^*}\beta_{a^*}(a^*)-\sum_{a,a^*}p(a)\beta_{a^*}(a)+\frac{1}{\eta}\sum_{a^*}\alpha(a^*)\log \frac{q(a^*)}{\alpha^\omega(a^*)}\nonumber\\&+\frac{1}{\eta}\sum_{a^*,a}p(a)( g(\theta^\omega_{\text{avg}}(a))-g(\theta^\omega_{a^*}(a)))\beta_{a^*}(a)+\frac{1}{\eta}\sum_{a^*,a}p(a)\alpha(a^*)\lb A(\theta^\omega_{a^*}(a))-A(\theta^\omega_{\avg}(a))\rb.\nonumber
\end{align}
Clearly, $\overline{B}$ is a linear function with respect to the parameterization $(\alpha, \beta)$ of $\nu$.

We assume that the action set $\Pi=[K]$ is a finite set of $K$ actions. By Danskin's theorem (Lemma \ref{lemma duality gradients}), we can calculate the gradient of AIR with respect to $\beta$ as
\begin{align*}
    \frac{\partial \air_{q,\eta}(p,\nu)}{\partial \beta_{a^*}(a)}=&\left.\frac{\partial \overline{B}(\nu,\omega)}{\partial \beta_{a^*}(a)}\right\vert_{\omega=\nu}
    \\= &\mathds{1}(a=a^*) -p(a)+\frac{1}{\eta} p(a) \lb g(\theta_{\avg}(a))-g(\theta_{a^*}(a))\rb,
\end{align*}
and we can calculate the gradient of AIR with respect to $\alpha$ as
\begin{align*}
    \frac{\partial \air_{q,\eta}(p,\nu)}{\partial \alpha(a^*)}=&\left.\frac{\partial \overline{B}(\nu,\omega)}{\partial \alpha(a^*)}\right\vert_{\omega=\nu}\\= &\frac{1}{\eta}\sum_{a}p(a)\lb A(\theta_{a^*}(a))-A(\theta_{\avg}(a))\rb+\frac{1}{\eta}\log\frac{q(a^*)}{\alpha(a^*)}.
\end{align*}
In statistics theory, the function $g$ is called the ``natural parameter.'' Denote
\begin{align*}
   g_{a^*}(a)= g(\theta_{a^*}(a)), \quad g_{\avg}(a)=g(\theta_{\avg}(a)),
\end{align*}
and write $A(\theta)$ as a function of $g$ defined by $$\bar{A}(g)= A(\theta).$$ In statistics theory the function $\bar{A}(g)$, which is a function of the natural parameter $g$, is called the log-partition function, because it is the logarithm of a normalization factor, without which $P_{\theta}(o)$ will not be a probability distribution. Mathematically, it can be defined as $\bar{A}(g)= \log\left(\int_{\mathcal{O}} h(o) \exp\left(g(\theta) \cdot T(o)\right) do\right)$. However, for the sake of simplicity, we will directly use the expression $\bar{A}(g) = A(\theta)$.

Then we can rewrite the derivatives with respect to $\beta$ in the form of
\begin{align}\label{eq: exponential family beta}
    \frac{\partial \air_{q,\eta}(p,\nu)}{\partial \beta_{a^*}(a)}&= \mathds{1}(a=a^*) -p(a)+\frac{1}{\eta} p(a) \lb g_{\avg}(a)-g_{a^*}(a)\rb,
    \end{align} and rewrite the derivatives with respect to $\alpha$ in the form of
    \begin{align}\label{eq: exponential family alpha}
    \frac{\partial \air_{q,\eta}(p,\nu)}{\partial \alpha(a^*)}&= \frac{1}{\eta}\sum_{a}p(a)\lb \bar{A}(g_{a^*}(a))-\bar{A}(g_{\avg}(a))\rb+\frac{1}{\eta}\log\frac{q(a^*)}{\alpha(a^*)}.
\end{align}

 Now we discuss whether we should set the equations \eqref{eq: exponential family alpha} to be zeros (i.e., making natural parameters ``mean-zero''), or we should set the equations \eqref{eq: exponential family beta} to be zeros (i.e., making the log-partition functions ``mean-zeros''). Note that we only have $\E_{a^*\sim\alpha}[\theta_{a^*}]=\theta_{\avg}$ by definition, and both the equations \eqref{eq: exponential family alpha} and the equations \eqref{eq: exponential family beta} are our targets to make nearly zero by setting the correct $(\alpha,\beta)$. As we will see, there will be trade-off between natural parameters and log-partition functions, and one need to look at the geometry of the exponential family to determine the formula to choose. We discuss the following two attempts:

(1) {\bf Setting derivatives w.r.t. $\beta$ to be $0$:} we set the following equations
\begin{align}\label{eq: exponential family set}
    g_{a^*}(a)-g_{\avg}(a)=\eta\frac{\mathds{1}(a=a^*)}{p(a)}-\eta, \quad \forall a^*,a\in \A
\end{align}
together with $\alpha=p$. In this way so that all the derivatives with respect to $\beta$ are all zeros, and we have the following ``mean-zero'' property for natural parameters: for all $a^*\in[K]$.
\begin{align*}
    \E_{a\sim p}[g_{a^*}(a)-g_{\avg}(a)=0].
\end{align*}In order to the make sure that the derivatives with respect to $\alpha$ are also nearly zero, we consider the posterior sampling strategy $p=q$ and want to show that 
\begin{align*}
    \left|\E_{a\sim p}\lb\bar{A}(g_{a^*}(a))-\bar{A}(g_{a^*}(a))\rb\right|=O(\eta^2).
\end{align*}To achieve this, we need the log-partition function $\bar{A}$ to be very ``flat,'' in the sense that its Hessian at $z$ grows no faster than $1/z$ so that the error  $\bar{A}(g_{a^*}(a))-\bar{A}(g_{a^*}(a))$ can be bounded by the ``local norm'' 
\begin{align*}
    \left|\bar{A}(g_{a^*}(a))-\bar{A}(g_{a^*}(a))\right|\leq O\lp \sum_{a}p(a)\lb g_{a^*}(a)-g_{a^*}(a) \rb^2\rp.
\end{align*}This is true for Bernoulli distribution, where the log-partition function is  in the form 
$ \bar{A}(g)=\log(1+e^g),$
 as well as the exponential distribution, where the log-partition function is in the form $\bar{A}(g)=-\log(-g)$.

(2) {\bf Setting derivatives w.r.t. $\alpha$ to be $0$:} If the log-partition function is not flat but increases ``sharply'', then we instead set the derivatives with $\alpha$ to be zeros. Then we need to show that the derivatives with respect to $\beta$ are nearly zeros. This is the case for Gaussian distribution, where the log-partition function $\bar{A}(g)=g^2/2$. Taking Gaussian distribution for example, we need to make $\alpha$ slightly different than $p$ to make their derivatives to be zero as in \eqref{eq: alpha zero}. But then it is convenient to show that the derivatives respect to $\beta$ are nearly zero and the equations \eqref{eq: exponential family set} are approximately satisfied up to the difference between $\alpha$ and $p$.

So we conclude that in order to get closed-form solutions and avoid tedious boundary conditions, there may be trade off between making \eqref{eq: exponential family alpha} zeros (i.e., making natural parameters ``mean-zero''), and making \eqref{eq: exponential family beta} zeros (i.e., making the log-partition functions ``mean-zero''). But nice calculations can often been done with care. We believe it is valuable to further explore this trade-off between the natural parameter and log-partition function and its role in automatically seeking the optimal bias-variance trade-off.

Finally, we would like to comment on the IPW-type formulation for the natural parameter $\eta$ in \eqref{eq: exponential family set}. Following the discussion concerning the connection between APS and mirror descent (see the end of Section \ref{subsec APS BMAB}), we conclude that the vanilla IPW estimator and traditional EXP3-type algorithms correspond to using exponential or Gaussian distributions (or any exponential family distribution with a linear natural parameter $g(\theta)=a\theta+b$ where $a$ and $b$ are constants). In contrast, Algorithm \ref{alg: BMAB} employs a more advanced estimator, which can be interpreted from both Bayesian and frequentist perspectives (see \eqref{eq: estimator bernoulli}).

\section{Proofs for MAIR and RL}\label{appendix mair}

\subsection{Proof of Lemma \ref{lemma mdir relationship}}
{\bf Lemma 2.8} (Bounding MAIR by DEC). {\it Given model class $\M$ and $\eta>0$, we have 
\begin{align*}
\sup_{\rho\in  \inte(\Delta(\M))}\sup_{\mu}\inf_{p} \ {{\texttt{{\textup{MAIR}}}}}_{\rho,\eta}(p,\nu) = \dec_{\eta}^\KL(\M)\leq   \texttt{\textup{DEC}}_{\eta}(\M).
\end{align*}
}

\paragraph{Proof of Lemma \ref{lemma mdir relationship}:}
By the definitions of MAIR and DEC we have the following:
\begin{align*}
  &\sup_{\rho\in\inte(\Delta(\M)))} \sup_{\mu\in\Delta(\M)}\inf_{p\in\Delta(\Pi)}\mair_{\rho,\eta}(p,\mu)\\
    = &\sup_{\rho\in\inte(\Delta(\M)))} \sup_{\mu\in\Delta(\M)}\inf_{p\in\Delta(\Pi)}\E_{\pi\sim p, M\sim \mu}\lb f_M(\pi_M)-f_M(\pi)-\frac{1}{\eta}\KL(M(\pi), {\mu}_{o|\pi})-\frac{1}{\eta}\KL(\mu,\rho)\rb\\
    =&\inf_{p\in\Delta(\Pi)}\sup_{\mu\in \Delta(\M)}\sup_{ \rho\in\textup{int}(\Delta(\M))}\E_{\pi\sim p, M\sim \mu}\lb f_M(\pi_M)-f_M(\pi)-\frac{1}{\eta}\KL(M(\pi), {\mu}_{o|\pi})-\frac{1}{\eta}\KL(\mu,\rho)\rb\\
    =&\inf_{p\in \Delta(\Pi)}\sup_{\mu\in \Delta(\M)}\E_{\pi\sim p, M\sim \mu}\lb f_M(\pi_M)-f_M(\pi)-\frac{1}
    {\eta}\KL(M(\pi), {\mu}_{o|\pi})\rb\\
=&\inf_{p\in\Delta(\Pi)}\sup_{\mu\in\Delta(\M)}\sup_{\bar{M}\in \textup{conv}(\M)}\E_{\pi\sim p, M\sim \mu}\lb f_M(\pi_M)-f_M(\pi)-\frac{1}
    {\eta}\KL(M(\pi), \bar{M}(\pi))\rb\\
=&\sup_{\bar{M}\in\textup{conv}(\M)}\sup_{\mu\in\Delta(\M)}\inf_{p\in\Delta(\Pi)}\E_{\pi\sim p, M\sim \mu}\lb f_M(\pi_M)-f_M(\pi)-\frac{1}
    {\eta}\KL(M(\pi), \bar{M}(\pi))\rb\\
   = &\sup_{\bar{M}\in\textup{conv}(\M)}\inf_{p\in\Delta(\Pi)}\sup_{M\in\M}\E_{\pi\sim p,M\sim\mu}\lb f_M(\pi_M)-f_M(\pi)-\frac{1}
    {\eta}\KL(M(\pi), \bar{M}(\pi))\rb
   \\= &\dec_{\eta}^\KL(\M),\\
   \leq &\dec_{\eta}(\M),
\end{align*}
where the first equality is by \eqref{eq: mair mutual information}; the second inequality is an application of the minimax theorem (Lemma \ref{lemma sion}); the third equality is by taking the supremum over $\rho$ and performing limit analysis; the fourth inequality is because the supremum over $\bar{M}$ will be achieved at $\bar{M}=\E_{\mu}[M]$, as per the property of the KL divergence; the fifth and the sixth inequalities are applications of the minimax theorem (Lemma \ref{lemma sion}); the seventh equality is by the definition of the KL version of DEC; and the last inequality is because the squared Hellinger distance is bounded by the KL divergence (Lemma \ref{lemma Hellinger KL}).

\hfill$\square$

\subsection{Proof of Theorem \ref{thm regret approx stochastic}}
{\bf Theorem 7.1} (Generic regret bound for arbitrary learning algorithm leveraging MAIR). {\it In the stochastic setting, given a finite model class $\M$, the ground truth model $M^*\in\M$, an arbitrary algorithm $\alg$ that produces decision probability $p_1,\dots,p_T$, and a sequence of beliefs $\mu_1, \dots, \mu_{T}$ where $\rho_{t}={(\mu_{t-1})}_{\pi^*|\pi,o}\in \inte(\Delta(\M))$ for all rounds, we have
\begin{align*}
  \reg_T\leq \frac{\log |\M|}{\eta} +\E\lb\sum_{t=1}^T \lp\mair_{\rho_t,\eta}(p_t,\mu_t)+ \lla\left. \frac{\partial \mair_{\rho_t,\eta}(p_t,\mu)}{\partial \mu}\right|_{\mu=\mu_t}, \mathds{1}(M^*)-\mu_t\rra\rp\rb,
\end{align*}
where $\mathds{1}(M^*)$ is the vector whose $M^*-$coordinate is $1$ but all other coordinates are $0$. 
}

\paragraph{Proof of Theorem \ref{thm regret approx stochastic}:}
In this subsection we prove Theorem \ref{thm regret approx stochastic}, a generic regret bound using MAIR. Recall the definition of MAIR:
\begin{align*}
     \texttt{MAIR}_{\rho,\eta}(p,\mu)= \E_{\mu,p}\lb f_M( \pi_M)-f_M(\pi)-\frac{1}{\eta}\KL(\nu_{\pi^*|\pi,o}, {\nu}_{\pi^*})-\frac{1}{\eta}\KL({\nu}_{\pi^*},q)\rb.
\end{align*}
Similar to the $B$ function \eqref{eq: Q formulation} we use in Appendix \ref{appendix thm regret} and Appendix \ref{appendix arbitrary}, given the decision probability $\pi\sim p$, we define
\begin{align*}
     B(\mu, Q)= \E_{\mu,p}\left[f_{M}(\pi_M)-f_{M}(\pi)+\frac{1}{\eta}\frac{\log \rho(M)}{\log Q[\pi,o](M)}\right].
\end{align*}
Note that $B(\nu, Q)$ is a strongly convex function with respect to $Q$.
We have the following formula:
\begin{align*}
    \texttt{MAIR}_{\rho,\eta}(p,\mu)= B(\mu, {\mu}(\cdot|\cdot, \cdot))= \inf_{Q}B(\mu,Q),
\end{align*}
where $\mu(\cdot|\cdot,\cdot)$ denote the ``posterior functional'' that maps the observation $(\pi,o)$ to posterior $\mu(M|\pi,o)\in\Delta(\M)$. Here, the first equality is by the definition of MAIR, and the second equality is by the first-order optimal condition.
By Danskin's theorem (Lemma \ref{lemma duality gradients}), we have
\begin{align*}
    \frac{\partial \texttt{MAIR}_{\rho,\eta}(p,\mu) }{\partial \mu} = \left.\frac{\partial B(\mu, Q)}{\partial \mu}\right \vert_{Q={\mu}(\cdot|\cdot, \cdot)}.
\end{align*}
By the above identity and the linearity of $B(\mu,Q)$ with respect to $\mu$, we have
\begin{align}\label{eq: model identity B}
   B(\mathds{1}(M^*), {\mu}(\cdot|\cdot, \cdot))= \mair_{\rho,\eta}(p,\mu)+\lla  \frac{\partial \mair^\Psi_{\rho,\eta}(p,\mu) }{\partial \mu} , \mathds{1}(M^*)-{\mu} \rra,
\end{align}
where $M^*$ is the underlying true model in the stochastic setting.

Following the very similar steps in proving Theorem \ref{thm regret} and Theorem \ref{thm regret approx}, we are able to prove Theorem \ref{thm regret approx stochastic}. For every $M^*\in\M$ we have
\begin{align*}
    \sum_{t=1}^T \lb \log\frac{\rho_{t+1}(M^*)}{\rho_t(M^*)}\rb=\log\frac{\rho_T(M^*)}{\rho_1(M^*)}\leq \log |\M|.
\end{align*}
Given an arbitrary algorithm and an arbitrary algorithmic belief sequence, we set the reference probability $\rho_{t+1}=\mu_t(M|\pi_t,o_t)$ to aid our analysis. By the above equation we have
\begin{align}\label{eq: model telescope bregman}
    \E\lb\sum_{t=1}^T\E_{t-1}\lb \frac{1}{\eta} \log\frac{\mu_t(M^*|\pi_t,o_t)}{\rho_t(M^*)}\rb\rb\leq \frac{\log|\M|}{\eta}.
\end{align}
Subtracting the telescope sum \ref{eq: telescope bregman} from regret, we have
\begin{align}\label{eq: model regret bregman long}
&\reg_T-\frac{\log|\M|}{\eta}\nonumber\\\leq
&\E\lb \sum_{t=1}^T f_{M^*}(\pi_{M^*})-\sum_{t=1}^T f_{M^*}(\pi_t)\rb-\frac{\log|\M|}{\eta}\nonumber\\
\leq &\E\lb\sum_{t=1}^T\E_{t-1}\lb f_{M^*}(\pi_{M^*})-f_{M^*}(\pi_t)-\frac{1}{\eta}\log\frac{\mu_t(M^*|\pi_t,o_t)}{\rho_t(M^*)}\rb\rb\nonumber\\
= &\E\lb \sum_{t=1}B_t(\mathds{1}(M^*), \mu_t(\cdot|\cdot, \cdot))\rb,
\end{align}
where the first inequality is by the definition of regret; the second inequality is by \eqref{eq: telescope}; and the third inequality by the definition of the $B$ function and the stochastic environment. Finally, taking the equivalent characterization  \eqref{eq: model identity B} for the $B$ function into \eqref{eq: model regret bregman long} proves Theorem \ref{thm regret approx stochastic}.

\hfill$\square$

\subsection{Proof of Theorem \ref{thm model adaptive minimax sampling}}
{\bf Theorem 7.2} (Regret of MAMS). {\it For a finite and compact model class $\M$, the regret of Algorithm \ref{alg: model adaptive minimax sampling} with any $\eta>0$ is always bounded as follows, for all $T\in\N$, 
\begin{align*}
  \reg_T\leq  \frac{\log|\M|}{\eta}+\E\lb\sum_{t=1}^T \mair_{\rho_t,\eta}(p_t,\mu_t)\rb\leq \frac{\log|\M|}{\eta}+\dec_{\eta}^{\KL}(\M)\cdot T.
\end{align*}
}

\paragraph{Proof of Theorem \ref{thm model adaptive minimax sampling}:}
Combining Theorem \ref{thm regret approx stochastic} and  Lemma \ref{lemma mdir relationship}, we prove Theorem \ref{thm model adaptive minimax sampling} in a straightforward manner:
\begin{align*}
    \reg_T\leq &\frac{\log|\M|}{\eta}+\E\lb\sum_{t=1}^T \mair_{\rho_t,\eta}(p_t,\nu_t)\rb\\
    \leq &\frac{\log|\M|}{\eta}+\dec_{\eta}^{\KL}(\M)\cdot T,
\end{align*}
where the first inequality is by Theorem \ref{thm regret approx} and the fact that MAMS always pick 
\begin{align*}
    \mu_t =\arg\max_{\mu\in\Delta(\M)}\mair_{\rho_t,\eta}(p_t,\mu),
\end{align*}
and the second inequality is by Lemma \ref{lemma mdir relationship}.

\hfill$\square$

\subsection{Proof for Theorem \ref{thm regret model}}\label{appendix subsec mair regret}
{\bf Theorem 7.3} (Regret for arbitrary algorithm with closed-form beliefs). {\it Given a finite model class $\M$ where the underlying true model is $M^*\in\M$, and $f_M(\pi)\in [0,1]$ for every $M\in\M$ and $\pi\in \Pi$. For an arbitrary algorithm $\alg$ and any $\eta>0$,  the regret of algorithm $\alg$ is bounded as follows, for all $T\in\N$,
\begin{align*}
    \reg_T\leq \frac{\log|\M|}{\eta}+\E\lb\sum_{t=1}^T\E_{\mu_t, p_t}\lb  f_M(\pi_M)-f_M(\pi)- \frac{1}{3\eta} D_{\textup{H}}^2(M(\pi), (\rho_t)_{o|\pi})-\frac{1}{3\eta}\KL(\mu_t,\rho_t)\rb\rb.
\end{align*}
where $\mu_t$ and $\mu_t$ are closed-form beliefs generated according the Algorithm \ref{alg: model index close form}.
}

\paragraph{Proof of Theorem \ref{thm regret model}:}
From Theorem \ref{thm regret approx stochastic}
we have that
\begin{align}
  \reg_T\leq \frac{\log |\M|}{\eta} +\sum_{t=1}^T \lp\mair_{\rho_t,\eta}(p_t,\mu_t)+ \lla\left. \frac{\partial \mair_{\rho_t,\eta}(p_t,\mu)}{\partial \mu}\right|_{\mu=\mu_t}, \mathds{1}(M^*)-\mu_t\rra\rp,\label{eq: regret model proof}
\end{align}
By Lemma \ref{lemma duality gradients} we have
\begin{align*}
    \frac{\partial\mair_{\rho,\eta}(p,\mu)}{\partial \mu(M)}= 
    f_M(\pi_M)-\E_{\pi\sim p}\lb f_M(\pi)\rb+\frac{1}{\eta}\E_{\pi\sim p,o\sim M(\pi)}\lb\log\frac{\rho(M)}{\mu(M|\pi,o)}\rb\\
    =  f_M(\pi_M)-\E_{\pi\sim p}\lb f_M(\pi)\rb-\frac{1}{\eta}\log\frac{\mu(M)}{\rho(M)}-\frac{1}{\eta}\E_{\pi\sim p,{o}\sim M(\pi)}\lb\log\frac{[M(\pi)]({o})}{\mu_o({o}|\pi)}\rb\\
    = f_M(\pi_M)-\E_{\pi\sim p}\lb f_M(\pi)\rb-\frac{1}{\eta}\log\frac{\mu(M)}{\rho(M)}-\frac{1}{\eta}\E_{\pi\sim p,{o}\sim M(\pi)}\lb\KL(M(\pi), \mu_{o|\pi})\rb.
\end{align*}
By using the updating rule in \eqref{eq: algorithmic belief MAIR}, 
 we have
  \begin{align*}
    \frac{\partial\mair_{\rho,\eta}(p,\mu)}{\partial \mu_t(M)}= 
   -\frac{1}{\eta}\E_{\pi\sim p_t}\lb\KL(M(\pi), {(\mu_t)}_{o|\pi})\rb+\frac{1}{3\eta}\E_{\pi\sim p}\lb D_{\text{H}}^2(M(\pi),{(\rho_t)}_{o|\pi})\rb+C_1, \forall M\in\M,
\end{align*}
where $C_1$ is some value that is the same for all $M\in \M$. This implies that
\begin{align}
   &\lla\left. \frac{\partial \mair_{\rho_t,\eta}(p_t,\mu)}{\partial \mu}\right|_{\mu=\mu_t}, \mathds{1}(M^*)-\mu_t\rra \nonumber\\
   = &\frac{1}{\eta}\E_{\mu_t,p_t}\lb \KL(M(\pi),{(\mu_t)}_{o|\pi})\rb-\frac{1}{\eta}\E_{\pi\sim p_t}\lb \KL(M^*(\pi),{(\mu_t)}_{o|\pi})\rb\nonumber\\ &
   -\frac{1}{3\eta}\E_{\mu_t, p_t}\lb D^2_{\text{H}}(M(\pi),{(\rho_t)}_{o|\pi})\rb+ \frac{1}{3\eta}\E_{\mu_t, p_t}\lb D^2_{\text{H}}(M^*(\pi),{(\rho_t)}_{o|\pi})\rb.
   \label{eq: model opt error}
\end{align}

Taking \eqref{eq: model opt error} into \eqref{eq: regret model proof}, we have
\begin{align}\label{eq: application thm regret model}
    \reg_T\leq &\frac{\log |\M|}{\eta}+\sum_{t=1}^T \E_{\mu_t,p_t}\bigg[ f_M(\pi_M)-f_M(\pi)-\frac{1}{3\eta}D^2_{\textup{H}}(M(\pi),(\rho_t)_{o|\pi})-\frac{1}{\eta}\KL(\mu_{t},\rho_t)\nonumber\\&+\frac{1}{3\eta}D^2_{\text{H}}(M^*(\pi),{(\rho_t)}_{o|\pi})-\frac{1}{\eta}\E\lb\KL(M^*(\pi),{(\mu_t)}_{o|\pi})\rb\bigg]\nonumber\\
    \leq &\frac{\log |\M|}{\eta}+\sum_{t=1}^T \E_{\mu_t,p_t}\bigg[ f_M(\pi_M)-f_M(\pi)-\frac{1}{3\eta}D^2_{\textup{H}}(M(\pi),(\rho_t)_{o|\pi})-\frac{1}{3\eta}\KL(\mu_{t},\rho_t)\bigg],
\end{align}
where the last inequality is because by information-theoretical inequalities, we have
\begin{align*}
    \frac{1}{3}D^2_{\text{H}}(M^*(\pi),{(\rho_t)}_{o|\pi})\leq  &\frac{2}{3}D^2_{\text{H}}(M(\pi),{(\mu_t)}_{o|\pi})+\frac{2}{3}D^2_{\text{H}}({(\mu_t)}_{o|\pi},{(\rho_t)}_{o|\pi})\\
    \leq &\frac{2}{3}D^2_{\text{H}}(M(\pi),{(\mu_t)}_{o|\pi})+\frac{2}{3}D^2_{\text{H}}(\mu_t,\rho_t)\\
    \leq &\frac{2}{3}\KL(M(\pi),{(\mu_t)}_{o|\pi})+\frac{2}{3}\KL(\mu_t,\rho_t),
\end{align*}
where the first inequality is by Lemma \ref{lemma triangle squared Hellinger}; the second inequality is by Lemma \ref{lemma data processing}; and the last inequality is by  Lemma \ref{lemma Hellinger KL}. Finally, by \eqref{eq: application thm regret model} we have that
\begin{align*}
\reg_T\leq \frac{\log|\M|}{\eta}+\E\lb\sum_{t=1}^T\E_{\mu_t, p_t}\lb  f_M(\pi_M)-f_M(\pi)- \frac{1}{3\eta} D_{\textup{H}}^2(M(\pi), (\rho_t)_{o|\pi})-\frac{1}{3\eta}\KL(\mu_t,\rho_t)\rb\rb.
\end{align*}
Therefore, we prove Theorem \ref{thm regret model}.

\hfill$\square$

\subsection{Proof of Theorem \ref{thm regret model ts}}
{\bf Theorem 7.4} (Regret of Model-index Adaptive Posterior Sampling). {\it
Given a finite model class $\M$ where $f_M(\pi)\in[0,1]$ for every $M\in\M$ and $\pi\in\Pi$. The regret of Algorithm \ref{alg: model index APS} with $\eta\in(0,1/3]$ is bounded as follows, for all $T\in\N$,
\begin{align*}
   \reg_T\leq \frac{\log|\M|}{\eta}+ \texttt{\textup{DEC}}^{\textup{TS}}_{6\eta}(\M,\bar{M})\cdot T+6\eta T.
\end{align*}
}

\paragraph{Proof of Theorem \ref{thm regret model ts}:}
Consider the Bayesian posterior sampling strategy 
induced by $\mu\in \Delta(M)$, which samples $M\sim \mu$ and plays $\pi_{M}$.  Denote the induced decision probability as
\begin{align*}
    {\mu}_{\pi_M}(\pi)=\sum_{M\in\M, \pi_M=\pi}\mu(M). 
\end{align*}
For arbitrary $\mu\in\Delta(M)$, denote the $|\Pi|-$dimensional vectors $X,Y$ by
\begin{align*}
    X(\pi)&= \E_{\mu}\lb f_M(\pi_M)-f_M(\pi)\rb, \\
    Y(\pi)&=  \E_{\mu}\lb D_{\textup{H}}^2(M(\pi), {(\rho)}_{o|\pi}))\rb.
\end{align*}
Then 
\begin{align*}
&\E_{M\sim\mu, \pi\sim{\rho}_{\pi_M}}\lb  f_M(\pi_M)-f_M(\pi)-\frac{1}{3\eta}D_{\textup{H}}^2(M(\pi), {\rho}_{o|\pi})\rb-\frac{1}{3\eta}\KL(\mu, \rho)\\
    \leq &\lla \rho_{\pi_M}, X\rra-\frac{1}{3\eta}\lla \rho_{\pi_M}, Y\rra-\frac{1}{3\eta}\KL({\mu}_{\pi_M}, \rho_{\pi_M})\\
    \leq &\lla \mu_{\pi_M}, X\rra +6\eta-\frac{1}{3\eta}\lla \rho_{\pi_M}, Y\rra-\frac{1}{6\eta}\KL(\mu_{\pi_M}, \rho_{\pi_M})\\
    \leq &\lla \mu_{\pi_M}, X\rra +6\eta-\frac{1}{3\eta}\lla \rho_{\pi_M}, Y\rra-\frac{1}{6\eta}D_{\textup{H}}^2(\mu_{\pi_M}, \rho_{\pi_M})\\
    \leq &\lla \mu_{\pi_M}, X\rra -\frac{(1-\eta)}{3(1+\eta)\eta}\lla \mu_{\pi_M}, Y\rra+6\eta\\
    \leq &  \E_{\mu,\pi\sim \mu_{\pi_M}}\lb f_M(\pi_M)-f_M(\pi)-\frac{1}{6\eta}D_{\textup{H}}^2(M(\pi), {(\rho)}_{o|\pi})\rb+6\eta,
\end{align*}
where the first inequality is by the data processing inequality (Lemma \ref{lemma data processing}); the second inequality is by Lemma \ref{lemma drifted error bound} and the fact $f_M(\pi)\in[0,1]$ for all $M$ and $\pi$; the third inequality is by Lemma \ref{lemma Hellinger KL};  the fourth inequality is a consequence of Lemma \ref{lemma Pinsker Hellinger} and the AM-GM inequality; and the last inequality uses the condition $\eta\leq \frac{1}{3}$.

Combining the above inequality with Theorem \ref{thm regret model}, we prove Theorem \ref{thm regret model ts}.

\hfill$\square$

\subsection{Proof of Theorem \ref{thm rl}}
{\bf Theorem 7.6} (Regret of MAMS and MAPS for RL problems in the bilinear class). {\it
    In the on-policy case, Model-index AMS (Algorithm \ref{alg: adaptive minimax sampling}) and Model-index APS (Algorithm \ref{alg: adaptive posterior sampling}) with optimally tuned $\eta$ achieve regret
\begin{align*}
\reg_T\leq O\big(H^2 L{\text{bi}}^2d_{\text{bi}}(\M)\cdot T\cdot\log|\M|\big).
\end{align*}
In the general case, Model-index AMS (Algorithm \ref{alg: adaptive minimax sampling}) and Model-index APS (Algorithm \ref{alg: adaptive posterior sampling}) with forced exploration rate $\gamma=\big(8\eta H^3L_{\textup{bi}}^2(\M)d_{\textup{bi}}(\M,\bar{M})\big)^{1/2}$ and optimally tuned $\eta$ achieves regret
\begin{align*}
\reg_T\leq O\big(\big(H^3 L_{\text{bi}}^2 d_{\text{bi}}(\M)\log|\M|\big)^{1/3}\cdot T^{2/3}\big).
\end{align*}
}

\paragraph{Proof of Theorem \ref{thm rl}:}
By Theorem 7.1 in \cite{foster2021statistical}, we have upper bounds for $\dec_{\eta}^{\text{TS}}$ as follows.

\begin{lemma}[Upper bounds for bilinear class reinforcement learning]\label{lemma upper bound dec rl}
    Let $\M$ be a bilinear class and let $\bar{M}\in\textup{conv}(\M)$. Let $\mu\in\Delta(\M)$ be given, and consider the modified Bayesian posterior sampling strategy that samples $M\sim \mu$ and plays $\pi_{M}^{\alpha}$, where $\alpha\in[0,1]$ is a parameter.
    
    1. If $\pi_M^{\textup{est}}=\pi_M$ (i.e., estimation is on-policy), this strategy with $\alpha=0$ certifies that
    \begin{align*}
      \dec^{\textup{TS}}_{\eta}(\M,\bar{M})\leq 4\eta H^2 L_{\textup{bi}}^2(\M)d_{\textup{bi}}(\M;\bar{M})
    \end{align*}
    for all $\eta>0$.
    
    2. For general estimation policies, this strategy with $\gamma=\lp8\eta H^3L_{\textup{bi}}^2(\M)d_{\textup{bi}}(\M,\bar{M})\rp^{1/2}$ certifies that
    \begin{align*}
        \dec^{\textup{TS}_\gamma}_{\eta}(\M,\bar{M})\leq \lp32\eta H^3 L_{\textup{bi}}^2(\M)d_{\textup{bi}}(\M;\bar{M})\rp^{1/2}.
    \end{align*}
 whenever $\gamma \geq 32 H^3 L_{\textup{bi}}^2(\M) d_{\textup{bi}}(\M,\bar{M})$.   
\end{lemma}
By applying the upper bounds on $\dec^{\textup{TS}}{\eta}$ from Lemma \ref{lemma upper bound dec rl} to Theorem \ref{thm adaptive minimax sampling} and Theorem \ref{thm regret model ts}, we immediately prove the regret bounds in Theorem \ref{thm rl} for Model-index AMS and Model-index APS.

\hfill$\square$

\section{Technical background}
\subsection{Conditional entropy}
In the discussion after Definition \ref{def air}, we utilize the following result stating that conditional entropy is concave. The reference provides a succinct proof to this result.
\begin{lemma}[Conditional  entropy of a probability measure is concave, \cite{3080334}]\label{lemma conditional entropy}
Let $\P$ be a probability measure on locally compact space $\X$, and let $\mathfrak{E}, \mathfrak{F}$ be countable partitions of the space. Define the entropy with respect to the partition $\mathfrak{E}$ as 
\begin{align*}
    H(\P,\mathfrak{E})=-\sum_{E\in\mathfrak{E}}\P(E)\log \P(E),
\end{align*}
and the conditional entropy as
\begin{align*}
    H(\P, \mathfrak{E}|\mathfrak{F})=\sum_{F\in\mathfrak{F}} \P(F)H(\P(\cdot|F), \mathfrak{E}).
\end{align*}
Then the conditional entropy $H(\P, \mathfrak{E}|\mathfrak{F})$ is a concave function with respect to $\P$.
\end{lemma}
\subsection{Minimax theorem}
We introduce the classical minimax theorem for convex-concave game.
\begin{lemma}[Sion's minimax theorem for values, \cite{sion1958general}]\label{lemma sion}
Let $\X$ and $\Y$ be convex and compact sets, and $\psi: \X\times \Y\rightarrow \R$ a function which for all $y\in \Y$ is convex and continuous in $x$ and for all $x\in \X$ is concave and continuous in $y$. Then 
\begin{align*}
    \min_{x\in \X}\max_{y\in \Y}\psi(x,y)= \max_{y\in \Y}\min_{x\in \X}\psi(x,y).
\end{align*}
\end{lemma}

\subsection{Danskin's theorem}\label{appendix danskin}
We reference Danskin's theorem, and the version we apply in this context is a consequence of the general result in \cite{bernhard1995theorem}.

\begin{lemma}[Danskin's theorem]\label{lemma duality gradients}
Let $\X$ be a convex and compact subset of a topological space,   $\Y$ be a convex and compact subset of a Hilbert space, and $\phi: \X\times \Y\rightarrow \R$ a function such that for all $y\in \Y$, $\phi(\cdot,y)$ is a strongly convex in $x$; and for all $x\in\X$, $\phi(x,\cdot)$ is a  continuously differentiable function in $y$. For each $y\in\Y$, let $x_y$ be the unique minimizer of the optimization problem
\begin{align*}
    \min_{x\in\X} \phi(x,y).
\end{align*}  Then for all $y\in \Y$, we have
\begin{align*}
    \nabla_y \phi(x_y,y)=\left.\frac{\partial \phi(x,y)}{\partial y}\right\vert_{x=x_y}.
\end{align*}
\end{lemma}
The directional derivative version of this lemma follows from Theorem D1 in \cite{bernhard1995theorem}, which imposes the following requirements: $\X$ is a compact subset of a topological space, $\Y$ is a subset a Banach space, $\phi$ has minimizers in $x$, and $\phi$ satisfies various upper semi-continuity conditions in $y$. If $\phi$ is differentiable in $y$, then directional derivatives exist along every direction and are equal to the inner product between the direction and the derivative of $\phi$. This leads us to Lemma \ref{lemma duality gradients} by the property of inner product. In our application of the lemma, $\X$ is a functional space, $\Y$ is a probability simplex, and $\phi$ exhibits strong convexity in $x$ while being linear in $y$. In fact, slightly generalizing Danskin's theorem for the differentiable case in \cite{bertsekas1999nonlinear} to a general compact $\X$ suffices for our purpose. Beyond its applicability to general theorems, we frequently utilize this fact in our calculations as a tool for simplifying computations (without the need to check any conditions in practical calculations). Intuitively (although not fully rigorous due to differentiability considerations), Lemma \ref{lemma duality gradients} can be understood as follows:
\begin{align*}
    \nabla_y \phi(x_y,y)= \left.\frac{\partial \phi(x,y)}{\partial x}\right\vert_{x=x_y}\cdot \nabla_{y} x_y+ \left.\frac{\partial(x,y)}{\partial y}\right\vert_{x=x_y}= \left.\frac{\partial(x,y)}{\partial y}\right\vert_{x=x_y},
\end{align*}
where the first equality follows from the elementary rules of derivative and the second equality arises from the optimality condition $$\left.\frac{\partial \phi(x,y)}{\partial x}\right\vert_{x=x_y}=0.$$

\subsection{Convex analysis}\label{subsec convex analysis}
We review some background in convex analysis, and refer to Part V in \cite{rockafellar2015convex} for their proofs.
\begin{definition}[Legendre function, \cite{lattimore2020bandit}]\label{def legendre}
    A proper convex function $f: \mathbb{R}^d\rightarrow \mathbb{R}\cup \infty$ is Legendre if 
    
    1) $\mathcal{D}=\text{int}(\text{dom}(\Psi))$ is non-empty; 
    
    2) $\Psi$ is differentiable and strictly convex on $\mathcal{D}$; and
    
    3) $\lim_{n\rightarrow \infty}\|\nabla f(u_n)\|_2=\infty$ for any sequence $\{u_n\}$ with $u_n\in \mathcal{D}$ for all $n$ and $\lim_{n\rightarrow\infty}u_n=u$ and some $u\in \partial \mathcal{D}$.
\end{definition}

 It is known that the minimizer of a Legendre function is always in its interior:
\begin{lemma}[Minimizer of Legendre function is in its interior, \cite{ lattimore2020bandit}]\label{lemma minimizer Legendre} If $\Psi$ is Legendre and $u\in\argmin_{u\in\textup{dom}(\Psi)} \Psi(u)$, then $u^*\in\inte(\textup{dom}(\Psi))$.
\end{lemma}

We assume the $d-$dimensional set $\W\subset \textup{dom}(\Psi):= \{u\in \R^d: \Psi(u)<\infty\}$ has bounded diameter, i.e.,
\begin{align*}
    \diam(\W):= \sup_{u,v\in \W} \Psi(u)-\Psi(v)<\infty.
\end{align*}
We denote the Fenchel-Legendre dual of $\Psi$ as 
\begin{align*}
    \Psi^*(a)=\sup_{u\in \R^d}\langle a,u\rangle -\Psi(u), \quad \forall a\in \R^d,
\end{align*}
and denote the Bregman divergences with respect to $\Psi$ and $\Psi^*$ as
\begin{align*}
    D_\Psi(u,v)=\Psi(u)-\Psi(v)-\langle \nabla \Psi(v), u-v\rangle,\\
    D_{\Psi*}(a,b)= \Psi^*(a)- \Psi^*(b) -\langle \nabla \Psi^*(b), a-b\rangle.
\end{align*}
We have the following properties of Legendre functions.
\begin{lemma}[Properties of Legendre function, \cite{lattimore2020bandit}]\label{lemma property legendre} If $\Psi$ is a Legendre function, then 

(a) $\nabla \Psi$ is a bijection between $\inte(\dom(\Psi))$ and $\inte(\dom(\Psi^*))$ with the inverse $(\nabla \Psi)^{-1}= \nabla \Psi^*$. That is, for $u\in \inte(\dom(\Psi))$, if $a= \nabla \Psi(u)$, then $a\in \inte(\dom(\Psi^*))$ and $\nabla \Psi^*(a)=u$;

(b) $  D_\Psi(u,v)= D_{\Psi_*}(\nabla \Psi(v), \nabla \Psi(u))$ for all $u,v\in \inte(\dom(\Psi))$; 

(c) the Fenchel conjugate $\Psi^*$ is Legendre; and 

\end{lemma}

We also introduce the generalized Pythagorean theorem of Bregman divergence.
\begin{lemma}[Generalized Pythagorean theorem]\label{lemma pythegorean} For a convex function $\Psi: \W\rightarrow\R\cup\infty$ and $u,v,w\in \W$, we have
\begin{align*}
   D_{\Psi}(u,w)=D_\Psi(u,v)+D_{\Psi}(v,w)+\langle u-v,\nabla \Psi(v)-\nabla \Psi(w) \rangle.
\end{align*}
\end{lemma}

\subsection{Information theory}
We have the following result stating that the squared Hellinger distance between two probability measures are smaller than the KL divergence between those two probability measures.
\begin{lemma}[Squared Hellinger distance smaller than KL divergence]\label{lemma Hellinger KL} For probability measures $\P$ and $\mathbb{Q}$,  the following inequalities hold:
\begin{align*}
    D_{\textup{H}}^2(\P,\mathbb{Q})\leq \KL(\P, \mathbb{Q}).
\end{align*}
\end{lemma}

For general $f-$divergences (which include KL divergence and squared Hellinger distance), we have the following data processing inequality. 

\begin{lemma}[Data Processing Inequality]\label{lemma data processing} Consider a channel that produces $Y$ given $X$ based on the
conditional law $\P_{Y|X}$.
Let $\P_Y$ (resp. $\mathbb{Q}_Y$) denote the distribution of $Y$ when $X$ is distributed as $\P_X$ (resp. $\mathbb{Q}_X$). For any f-divergence $D_f(\cdot||\cdot)$,
\begin{align*}
   D_f(\P_Y||\mathbb{Q}_Y) \leq D_f(\P_X||\mathbb{Q}_X). 
\end{align*}

\end{lemma}
We introduce a  localized version of Pinsker-type inequality using squared Hellinger distance (which will be stronger than using the KL divergence).
\begin{lemma}[Multiplicative Pinsker-type inequality for Hellinger distance, \cite{foster2021statistical}]\label{lemma Pinsker Hellinger} Let $\P$ and $\mathbb{Q}$ be probability measures on compact space $\X$. For all $h:\X\rightarrow \R$ with $0\leq h(X)\leq R$ almost surely under $\P$ and $\mathbb{Q}$, we have
\begin{align*}
    |\E_{\P}[h(X)]-\E_{\mathbb{Q}}[h(X)]|\leq \sqrt{2R(\E_{\P}[h(X)]+\E_{\mathbb{Q}}[h(X)])\cdot D_{\textup{H}}^2(\P,\mathbb{Q})}.
\end{align*}
\end{lemma}

We introduce a standard one-sided bound using KL divergence. Compared with Lemma \ref{lemma Pinsker Hellinger}, the upper bound in Lemma \ref{lemma drifted error bound} only depends on the probability measure $q$, while the bound is one-sided and it does not take the square-root form as in Lemma \ref{lemma Pinsker Hellinger}.
\begin{lemma}[Drifted error bound using KL divergence]\label{lemma drifted error bound}For any $p,q\in\Delta(\Pi)$, $\eta>0$, and any vector $y\in\R^\Pi$ where $y(\pi)\leq 1/\eta$ for all $\pi\in\Pi$, we have
\begin{align*}
    \langle y, p-q\rangle -\frac{1}{\eta}\KL(p,q)\leq \eta\sum_{\pi\in\Pi}q(\pi)y(\pi)^2.
\end{align*}
\end{lemma}
\noindent{\bf Proof of Lemma \ref{lemma drifted error bound}: }
consider the KL divergence $\psi_{q,\eta}(p)=\frac{1}{\eta}\KL(p||q)$, it is known that the convex conjugate duality of $\psi_q$ is the log partition function
\begin{align}\label{eq: entropy barrier}
   \psi_{q,\eta}^*(y): = \sup_{p\in \Delta(\Pi)}\left\{\langle y, p\rangle -\frac{1}{\eta}\KL(p||q)\right\}\nonumber\\
   = \frac{1}{\eta}\log\lp\sum_{\pi\in\Pi} q(\pi) \exp(\eta {y(\pi)})\rp.
\end{align}
We have
\begin{align}\label{eq: key analysis duality}
    \langle y,p\rangle-\frac{1}{\eta}\KL(p||q)\nonumber\\
    \leq \frac{1}{\eta}\log \lp\sum_{\pi\in\Pi} q\exp(\eta y(\pi)) \rp\nonumber\\
    \leq \frac{1}{\eta}\log \lp \sum_{\pi} q(\pi)(1+\eta y(\pi)+\eta^2y(\pi)^2)\rp
    \nonumber\\
    =\frac{1}{\eta}\log\lp 1+ \eta\langle y, q\rangle+\eta^2\sum_{\pi\in\Pi}q(\pi)y(\pi)^2\rp 
   \nonumber \\ \leq \langle y, q \rangle+ \eta\sum_{\pi\in\Pi}q(\pi)y(\pi)^2,
\end{align} 
where the first equality is because of \eqref{eq: entropy barrier}; the second inequality is because
$ e^z\leq 1+z+z^2$ for all $z\leq 1$
and the last inequality is due to $\log (1+z)\leq z$ for all $z\in \R$. 
Therefore we have
\begin{align*}
    \langle y,p-q\rangle -\frac{1}{\eta}\KL(p||q)\leq \eta\sum_{\pi\in\Pi}q(\pi)y(\pi)^2
\end{align*} 
for all $y\in\R^{|\Pi|}$ where $y(\pi)\leq 1/\eta$ for all $\pi$.

\hfill$\square$

By the fact that the Hellinger distance satisfies the canonical triangle inequality, we have the following triangle-type inequality for the squared Hellinger distance.

\begin{lemma}[Triangle-type inequality for squared Hellinger distance]\label{lemma triangle squared Hellinger}For every distributions $p,\bar{p},q\in\Delta(\X)$, we have
\begin{align*}
    D_{\textup{H}}^2(p,\bar{p})\leq 2\lp D_{\textup{H}}^2(p,q)+D_{\textup{H}}^2(q,\bar{p})\rp.
\end{align*}As a result, given any mixture $\mu\in\Delta(\Delta(\X))$, for every  $q\in\Delta(\X)$, we have
\begin{align*}
  \E_{p\sim \mu, \bar{p}\sim \mu}\lb D_{\textup{H}}^2(p,\bar{p})\rb\leq 4 \E_{p\sim \mu}\lb D_{\textup{H}}^2(p,q)\rb.
\end{align*}
\end{lemma}

\end{document}